\documentclass{article}

\usepackage{arxiv}

\usepackage[utf8]{inputenc}
\usepackage[T1]{fontenc}
\usepackage{amsmath,amssymb}
\usepackage{array}
\usepackage{booktabs}
\usepackage{enumitem}
\usepackage{longtable}
\usepackage{microtype}
\usepackage{graphicx}
\usepackage{natbib}
\usepackage{placeins}
\usepackage{xcolor}
\usepackage{tikz}
\usetikzlibrary{arrows.meta,positioning,shapes.geometric,plotmarks}
\usepackage{url}
\usepackage{hyperref}

\newcounter{algorithmblock}
\newenvironment{algorithmblock}[1]
{%
  \refstepcounter{algorithmblock}%
  \begin{center}
  \begin{minipage}{0.96\linewidth}
  \small
  \hrule\vspace{0.5em}
  \noindent\parbox{\linewidth}{\textbf{Algorithm \thealgorithmblock: #1}}\par
  \vspace{0.35em}\noindent\ignorespaces
}
{%
  \vspace{0.4em}\hrule
  \end{minipage}
  \end{center}
}

\title{AI Consilium: \\ a Benchmark of LLM Clinical Multidisciplinary Second Opinions}

\author{%
  \begin{minipage}{0.98\textwidth}
    \centering
    \normalsize
    Alexey A. Bokov\textsuperscript{1,\(\dagger\)},
    Ivan I. Panin\textsuperscript{1,\(\dagger\)},
    Nikolay B. Nechaev\textsuperscript{1},\\[0.3em]
    Mikhail Yu. Valkov\textsuperscript{2,3},
    Valeria Yu. Chernina\textsuperscript{2,4},
    Natalia Gombolevskaya\textsuperscript{5},\\[0.3em]
    Ayyaz Hussain\textsuperscript{2,6},
    Vladimir M. Agafonov\textsuperscript{2},
    Victor A. Gombolevskiy\textsuperscript{1,7},\\[0.3em]
    Aleksandr I. Nesterov\textsuperscript{1},
    Iaroslav R. Bespalov\textsuperscript{1},\\[0.3em]
    Ivan V. Oseledets\textsuperscript{1,\(\ddagger\)},
    Dmitry V. Dylov\textsuperscript{8,1,*,\(\ddagger\)}
    \\[0.9em]
    \small
    \textsuperscript{1} AIRI, Moscow, Russian Federation\\
    \textsuperscript{2} Northern State Medical University,
    Arkhangelsk, Russian Federation\\
    \textsuperscript{3} Arkhangelsk Clinical Oncological Dispensary,
    Arkhangelsk, Russian Federation\\
    \textsuperscript{4} Intelligent Radiology Assistance Laboratories
    LLC (IRA LABS), Moscow, Russian Federation\\
    \textsuperscript{5} ``Professor Gorbakov Clinic'' LLC,
    Krasnogorsk, Moscow Region, Russian Federation\\
    \textsuperscript{6} Department of Anesthesiology and Intensive
    Care Medicine, First City Hospital named after E.~E.~Volosevich,
    Arkhangelsk, Russian Federation\\
    \textsuperscript{7} Sechenov University,
    Moscow, Russian Federation\\
    \textsuperscript{8} Skoltech, Moscow, Russian Federation
    \\[0.7em]
    \footnotesize
    \textsuperscript{\(\dagger\)} These authors contributed equally
    to this work and share first authorship.\\
    \textsuperscript{\(\ddagger\)} These authors jointly supervised
    this work and share senior authorship.
    \\[0.4em]
    \textsuperscript{*} Correspondence:
    Dmitry V. Dylov 
    (\href{mailto:dylov@airi.net}{\texttt{dylov@airi.net}}).
  \end{minipage}%
}

\date{}

\renewcommand{\shorttitle}{AI Consilium: a Benchmark of LLM Clinical Multidisciplinary Second Opinions}

\hypersetup{
  hidelinks,
  pdftitle={AI Consilium: a Benchmark of LLM Clinical Multidisciplinary Second Opinions},
    pdfauthor={Alexey A. Bokov, Ivan I. Panin, Nikolay B. Nechaev,
    Mikhail Yu. Valkov, Valeria Yu. Chernina, Natalia Gombolevskaya,
    Ayyaz Hussain, Vladimir M. Agafonov, Victor A. Gombolevskiy,
    Alexander I. Nesterov, Iaroslav R. Bespalov,
    Ivan V. Oseledets, Dmitry V. Dylov},
  pdfkeywords={large language models, clinical reasoning, second opinion, multidisciplinary consultation, LLM-as-a-judge}
}

\newcommand{\ConsModelCountWord}{eight}

\newcommand{\ConsModelList}{DeepSeek-V4-Pro, Gemma-4-31B, GLM-5.2, GPT-5.6-Sol, GPT-OSS-120B, Qwen3.5-27B, Qwen3-235B, and Qwen3-Next-80B}

\newcommand{\ConsPanels}{1,520}

\newcommand{\ConsModelFormatCombinations}{16}
\newcommand{\ConsRoleSlots}{7,271}
\newcommand{\ConsMeanRoles}{4.78}
\newcommand{\ConsRoleOutputsTotal}{14,542}
\newcommand{\ConsExactLabels}{696}
\newcommand{\ConsPrimaryExactLabels}{212}
\newcommand{\ConsTwoRolePanels}{8}
\newcommand{\ConsTwoRolePanelsPct}{0.5}
\newcommand{\ConsThreeRolePanels}{53}
\newcommand{\ConsThreeRolePanelsPct}{3.5}
\newcommand{\ConsFourRolePanels}{199}
\newcommand{\ConsFourRolePanelsPct}{13.1}
\newcommand{\ConsFiveRolePanels}{1,260}
\newcommand{\ConsFiveRolePanelsPct}{82.9}
\newcommand{\ConsModelsMedianFive}{7}
\newcommand{\ConsNeurologyPrimary}{245}
\newcommand{\ConsNeurologyPrimaryPct}{16.1}
\newcommand{\ConsInfectiousPrimary}{199}
\newcommand{\ConsInfectiousPrimaryPct}{13.1}
\newcommand{\ConsRadiologyAll}{729}
\newcommand{\ConsRadiologyPrimary}{3}
\newcommand{\ConsRadiologyPrimaryAmongAssignmentsPct}{0.4}
\newcommand{\ConsTotalArms}{3,040}
\newcommand{\ConsInitialUnanimous}{1,204}
\newcommand{\ConsInitialUnanimousPct}{39.6}
\newcommand{\ConsConsensus}{2,974}
\newcommand{\ConsConsensusPct}{97.8}

\newcommand{\ConsInitiallyDisputed}{1,836}
\newcommand{\ConsInitiallyDisputedConsensus}{1,770}
\newcommand{\ConsInitiallyDisputedConsensusPct}{96.4}
\newcommand{\ConsConsensusByRoundTwo}{2,576}
\newcommand{\ConsConsensusByRoundTwoPct}{84.7}

\begin{document}
\maketitle

\begin{abstract}
Large language models (LLMs) can generate differential diagnoses for complex cases, but a useful second opinion must also be factually grounded, clinically coherent, actionable, and safe. We developed a multidimensional benchmark using 300 diagnostically resolved NEJM--MedQA cases: 110 for physician reference ratings and Judge calibration and 190 for held-out evaluation. Eight single-model pipelines generated independent role-specific opinions and interactive-consilium reports under free-form and structured clinical-reasoning conditions. A physician-calibrated LLM Judge evaluated 3,040 final reports and the corresponding structured primary-role outputs. Model identity separated report quality more strongly than response format. Structured final reports had higher overall mean scores in seven of eight model pipelines, with their clearest advantage in management (2.740 vs.\ 2.436). Compared with the structured pre-consilium primary-role opinion, the complete structured synthesis had higher pooled scores across all five shared criteria, with the largest increases in problem prioritization, factual grounding, and reasoning coherence. GPT-5.6-Sol led every evaluated component and reached consensus in the fewest review rounds. These findings show that the benchmark can distinguish model capability, the contribution of multidisciplinary synthesis, criterion-specific effects of response structure, deliberation dynamics, and inference requirements within one common-case evaluation.
\end{abstract}


\section{Introduction}
Large language models have already demonstrated substantial progress on complex diagnostic tasks. GPT-4 generated clinically meaningful differential diagnoses for cases from the \emph{Case Records of the Massachusetts General Hospital}, a long-running clinicopathological-conference series published in the \emph{New England Journal of Medicine} (NEJM) \citep{brinkmann2024caserecords,kanjee2023gpt4}. We refer to cases from this series as ``NEJM cases.'' Subsequent studies reported strong performance on challenging clinical cases across multiple specialties \citep{rutledge2024diagnostic,hirosawa2024generative}. In the study by McDuff et al., the Articulate Medical Intelligence Explorer (AMIE) system was evaluated on 302 NEJM cases and demonstrated strong performance both when independently generating differential diagnoses and when used by physicians as a decision-support tool \citep{mcduff2025amie}. These findings suggest that contemporary LLMs are already capable of making a meaningful contribution to diagnostic reasoning in complex clinical cases.

However, a clinical second opinion is not reducible to generating a differential diagnosis or selecting the most likely diagnostic hypothesis. A second opinion may change the representation of the patient's condition, identify missing or contradictory information, reprioritize diagnostic hypotheses, alter the diagnostic workup or treatment strategy, and highlight previously underestimated risks. Studies by Payne et al. and Meyer et al. have shown that repeat expert review may lead to changes not only in diagnosis but also in treatment, with management changes in some settings occurring more frequently than diagnostic changes; moreover, it is not always possible to determine objectively which of the two opinions was more accurate \citep{payne2014secondopinion,meyer2015secondopinion}. Thus, for LLMs, high diagnostic accuracy alone does not imply a high-quality clinical second opinion: diagnostic agreement does not fully capture factual grounding, the coherence of reasoning, the quality of diagnostic and therapeutic planning, uncertainty, or clinical safety.

One approach to improving LLM performance is to explicitly structure the reasoning process. In radiology tasks, separating the structuring of clinical information from subsequent diagnosis generation improved accuracy compared with conventional zero-shot chain-of-thought prompting \citep{sonoda2025structured}. A two-stage diagnostic framework in which a distinct verification stage was introduced between the preliminary and final diagnoses also produced moderate, model-dependent improvements on MedQA and NEJM cases, while increasing answer consistency and reducing the frequency of several reasoning errors \citep{shao2025verification}. However, the evidence remains heterogeneous. Prompts designed to imitate intuitive, analytical, Bayesian, or differential-diagnostic styles of physician reasoning did not produce consistent accuracy gains, particularly for stronger models \citep{savage2024reasoning}. More generally, the presence of an elaborate textual rationale should not be treated as evidence that the internal process leading to the answer was itself correct: chain-of-thought explanations can retrospectively rationalize decisions that were in fact driven by information not acknowledged in the explanation \citep{turpin2023unfaithful}. Structured clinical reasoning should therefore be evaluated as a distinct empirical intervention rather than assumed to be superior merely because the response is longer or appears more ``physician-like.''

A second approach is to distribute analysis across multiple agents representing different medical specialties. Its natural clinical analogue is the multidisciplinary team (MDT), in which specialists from different disciplines contribute complementary information to diagnosis and treatment planning. Studies of real-world MDTs, particularly in oncology, suggest that team discussion can alter diagnosis and management, although the magnitude of the effect varies substantially across studies and depends on information completeness, team structure, workload, leadership, and case complexity \citep{lamb2011mdt,basta2017mdt,koco2021mdt,soukup2020mdt}. This logic has already been transferred to LLM-based systems. MDAgents dynamically selects between individual and collaborative modes depending on estimated medical complexity \citep{kim2024mdagents}. Multi-Agent Conversation (MAC) employs multiple physician agents and a supervising agent that iteratively discuss clinical cases \citep{chen2025mac}. Model Confrontation and Collaboration (MCC) implements structured debate among heterogeneous LLMs \citep{sun2026mcc}. EvoMDT combines specialized agents, retrieval, and a coordinating component \citep{liu2026evomdt}. These systems demonstrate the feasibility of multi-agent medical reasoning and motivate the use of a multidisciplinary LLM council, but they do not by themselves establish that inter-agent interaction is the causal source of performance improvements.

This distinction is particularly important because multi-agent systems generally consume more computational resources than a single generation. Repeated sampling and self-consistency can improve reasoning performance without any message exchange between agents \citep{wang2023selfconsistency}, while Self-Refine improves responses through repeated cycles of critique and revision performed by the same model \citep{madaan2023selfrefine}. In non-medical settings, multi-agent debate has improved reasoning quality and factual accuracy \citep{du2024debate}; however, compute-matched analyses suggest that its advantage over self-consistency may be substantially smaller than its advantage over a single model response produced without iterative critique or inter-agent exchange \citep{wunderlich2026compute}. Deliberation itself may also introduce new classes of errors: consensus pressure can suppress a correct minority position, motivating consensus-free and trajectory-aware approaches \citep{cui2026freemad}. Role assignment constitutes another independent variable. Specialized physician roles are intuitively appealing, yet studies of role and persona prompting indicate that assigning a role label alone does not guarantee improved objective performance and may make outputs sensitive to clinically irrelevant persona attributes \citep{zheng2024personas,araujo2025personas}. A rigorous medical evaluation should therefore attempt to disentangle the effects of additional sampling, role diversity, aggregation, and inter-agent interaction itself.

A separate challenge is how quality should be evaluated. Contemporary studies of medical LLMs increasingly rely on physician-developed rubrics rather than diagnostic agreement alone \citep{cabral2024clinical,goh2024diagnostic}. Scalable automated evaluation methods are also emerging: LLMEval-Med combines expert-designed checklists with LLM-as-a-judge evaluation \citep{zhang2025llmeval}, MedThink-Bench evaluates expert-curated clinical reasoning trajectories using reference-informed LLM judging \citep{zhou2026medthink}, and studies of clinical summarization have reported high agreement between some reasoning models and physician evaluators \citep{croxford2025judge}. These findings make automated evaluation a promising measurement tool, but they do not eliminate the need for physician reference ratings, criterion-specific validation, or analysis of judge-model bias and stability.

In this study, we develop a benchmark of LLM-generated clinical second opinions using 300 published NEJM cases from the NEJM-MedQA diagnostic-reasoning dataset \citep{savage2024reasoning}. Response format, inference method, role composition, and model choice are treated as separate comparison dimensions. The benchmark has two evaluation checkpoints. At the first checkpoint, every case-specific clinical role produces an independent, standalone second opinion. At the second, all role-specific opinions are provided to an interactive multidisciplinary consilium that synthesizes a case-level final opinion through iterative review and revision. The interactive consilium protocol provides a fixed synthesis stage across these comparisons. This design supports controlled comparisons without requiring benchmark users to reproduce one fixed generation architecture. A fixed set of 110 cases is used for physician annotation and judge calibration, the remaining 190 cases are reserved for held-out evaluation.

The main contributions of this work are:
\begin{enumerate}
    \item a benchmark that evaluates role-specific opinions and final consilium reports as distinct clinical products on the same cases
    \item a multidimensional clinical-quality framework that separates factual grounding, problem prioritization, data use, reasoning coherence, differential diagnosis, final-conclusion quality, management, and safety
    \item controlled comparison of model identity, response structure, and synthesis stage together with deliberation dynamics and inference requirements
    \item a physician-developed rubric and criterion-wise calibration of an LLM Judge for scalable evaluation of open-ended clinical second opinions
\end{enumerate}

\section{Related Work}
\label{sec:related-work}

\subsection{Second Opinions and Multidisciplinary Teams}

In clinical practice, second opinions are evaluated by more than whether they change a diagnosis. A systematic review of patient-initiated second opinions showed that repeat expert assessment can lead to changes in diagnostic formulation, treatment, and patient satisfaction \cite{payne2014secondopinion}. An analysis of a national second-opinion program likewise demonstrated that disagreement with the original assessment may involve both diagnosis and management \cite{meyer2015secondopinion}. These findings support evaluating a second opinion as a multidimensional clinical product rather than solely by agreement on the final diagnosis.

A similar logic emerges from the literature on multidisciplinary teams (MDTs). Systematic reviews in oncology show that MDT discussion frequently changes management, although its effect is heterogeneous and depends on workflow organization, the quality and completeness of the available information, and team composition \cite{lamb2011mdt,basta2017mdt,koco2021mdt}. Observational studies of team decision-making further suggest that collective deliberation should not be equated with a guarantee of correctness \cite{soukup2020mdt}. Accordingly, consensus itself cannot serve as a surrogate endpoint for quality in a multidisciplinary LLM council. Related medical-AI evaluation work formalizes the same concern by comparing algorithmic diagnoses against multiple expert opinions rather than a single majority label, thereby accounting explicitly for inter-expert variability \cite{kopanichuk2026beyond}. More informative questions concern which facts, hypotheses, investigations, risks, and treatment options are preserved, revised, or lost during the process of council deliberation and synthesis.

\subsection{LLMs on Complex Clinical Cases: From Diagnosis to the Full Decision Trajectory}

Early studies showed that contemporary LLMs can solve difficult diagnostic problems based on published clinical cases. Kanjee et al. evaluated GPT-4 on diagnostic challenges, while Cabral et al. compared model-generated clinical reasoning with physician reasoning \cite{kanjee2023gpt4,cabral2024clinical}. Similar findings have been reported in series of complex case reports and clinical scenarios \cite{hirosawa2024generative,rutledge2024diagnostic}. The AMIE study extended this line of work to 302 challenging published case reports and a physician-assistance setting, demonstrating the potential of LLMs both for differential diagnosis and for supporting physicians \cite{mcduff2025amie}. Complementary clinical AI systems have addressed narrower but clinically important tasks, including diagnosis prediction from longitudinal health records and automated mapping of free-text diagnoses to ICD codes \cite{umerenkov2025ai,nesterov2025ruccod}. These systems illustrate the broader clinical-decision-support context in which generated second opinions would operate.

High diagnostic accuracy, however, is not equivalent to comprehensive response quality. Hager et al. developed MIMIC-CDM using 2,400 real-world cases and evaluated not only diagnosis but also information gathering, interpretation of laboratory findings, adherence to diagnostic and treatment recommendations, robustness to the order and amount of available information, and instruction-following ability \cite{hager2024limitations}. Their results indicate that strong diagnostic accuracy does not guarantee adequate performance across the full clinical decision-making process. Similarly, a randomized trial of management reasoning found that GPT-4 assistance can improve physician performance on open-ended patient-care tasks, but that its effect needs to be evaluated separately from diagnostic accuracy and with consideration of interaction time \cite{goh2025patientcare}.

In 2026, Medieval introduced a broad multidisciplinary blinded expert evaluation framework for diagnosis and management across 685 real and simulated clinical questions, including explicit recording of fabricated medical facts \cite{chen2026medieval}. This work is particularly relevant as a precedent for multidimensional evaluation of free-form clinical responses. Reliability also requires recognizing when the available evidence is insufficient: uncertainty-aware selective-prediction methods for medical text can improve diagnostic reliability by allowing a system to abstain from uncertain decisions \cite{vazhentsev2026uncertainty}. Taken together, these studies support separately assessing case-fact utilization, quality of explanation, differential diagnosis, diagnostic workup, treatment, risk, uncertainty, and safety rather than relying on a single aggregate measure of diagnostic accuracy.

\subsection{Structured Clinical Reasoning, Verification, and Test-Time Inference}

Medical studies of structured prompting suggest that the format of reasoning can affect output quality even when the underlying model remains unchanged. Sonoda et al. compared direct and structured prompting on the same radiology cases and observed improved diagnostic performance with an explicit clinical-reasoning structure \cite{sonoda2025structured}. Savage et al. showed that diagnostic reasoning prompts can elicit more interpretable clinical rationales, although the generated chain of thought should not be interpreted as a faithful record of the model's latent internal reasoning process \cite{savage2024reasoning}. Shao and Zhang employed a two-stage framework with predefined verification steps on MedQA and NEJM cases and found that an explicit verification stage could improve diagnostic performance and uncertainty handling \cite{shao2025verification}.

Strong inference-time baselines are essential for interpreting such gains. Self-consistency demonstrates that simply increasing the number of independent generations and subsequently aggregating them can itself improve reasoning performance \cite{wang2023selfconsistency}. Self-Refine demonstrates the value of iterative self-critique and answer revision by a single model \cite{madaan2023selfrefine}. Therefore, when a structured or council-based condition uses more model calls, more tokens, or additional critique stages, superior performance cannot automatically be attributed to the structure or to inter-agent interaction. General-domain equal-compute studies of test-time scaling further highlight the importance of comparing debate, mixtures or ensembles, and self-consistency under comparable inference budgets \cite{wunderlich2026compute}. Recent medical work has further treated the decision pipeline itself as an inference-time optimization target, using LLM-guided evolution to improve accuracy--cost trade-offs in interactive consultation and other clinical decision tasks \cite{sviridov2026llm}. Although these findings are methodological rather than direct clinical evidence, they provide an important basis for designing appropriate baseline comparisons. 

\subsection{Medical Multi-Agent Systems and the Computational Confounding Problem}

Multi-agent architectures in medicine are developing rapidly. MedAgents organizes LLM roles as a set of discipline-specific experts that first provide independent assessments, then engage in multiple rounds of discussion, and finally converge on a shared decision \cite{tang2024medagents}. MDAgents adaptively selects a solo, group, or more complex collaboration mode based on estimated task complexity and achieves strong results across several medical benchmark settings \cite{kim2024mdagents}. MAC employs multiple doctor agents and a supervisor to diagnose rare diseases and subsequently recommend additional investigations \cite{chen2025mac}. More recent systems extend this paradigm through confrontation and debate or structured MDT-style orchestration \cite{sun2026mcc,liu2026evomdt}. Related deployed systems include CLARITY, which combines structured dialogue control with collaborative LLM agents for patient routing, clinical consultation, and triage \cite{shaposhnikov2025clarity}.

These studies establish the feasibility of multi-agent approaches but do not demonstrate that interaction itself is responsible for performance gains. A multi-agent arm typically introduces several changes simultaneously: more generations, larger context windows, a supervisor or aggregator, different prompting strategies, and in some cases retrieval or structured knowledge. MDAgents also modifies the workflow itself according to case complexity, meaning that its group mode does not cleanly isolate the effect of interaction \cite{kim2024mdagents}. In a large benchmark of agentic clinical decision systems, Liu et al. found only moderate accuracy improvements over standalone LLMs despite more than a tenfold increase in token usage and more than a twofold increase in latency \cite{liu2026agentbenchmark}. These findings make inference cost and compute-matched controls essential for interpreting the performance of a multidisciplinary LLM council.

Role labels should likewise not be equated with validated medical specialization. In most medical-agent studies, specialization is induced through prompting rather than through dedicated training and external validation of agent competence. General-domain studies of persona prompting have produced mixed results, further limiting strong claims that assigning a specialist role automatically creates stable domain-specific expertise \cite{zheng2024personas,araujo2025personas}.

\subsection{Human--AI Collaboration, Automation Bias, and Clinical Safety}

Even a strong standalone LLM does not guarantee improvement in a physician--AI team. In a randomized diagnostic trial, access to GPT-4 did not produce a statistically significant improvement in the primary diagnostic-reasoning outcome compared with conventional resources, despite strong exploratory performance of GPT-4 alone \cite{goh2024diagnostic}. In a subsequent randomized management trial, physicians using GPT-4 achieved higher scores on patient-care tasks but required more time, and the physician-plus-GPT condition did not automatically outperform GPT alone \cite{goh2025patientcare}. In a 2026 randomized controlled trial, both first-opinion and second-opinion AI workflows improved diagnostic performance relative to conventional resources, yet the collaborative workflow again did not demonstrate automatic superiority over the AI-alone condition \cite{everett2026teammate}.

Systematic reviews and meta-analyses published in 2026 further support the heterogeneity of human--LLM collaboration. For several diagnostic endpoints, LLM-assisted clinicians outperform clinicians alone on average, but effect sizes vary across models, study designs, and case types, while risks of bias and publication bias remain substantial \cite{wang2026humanllmcollab,chen2026diagnosistriagemeta}. These studies concern clinician outcomes and human--AI interaction, which are outside the scope of the present benchmark; here, physician ratings are used as an expert reference for the quality of the generated material itself.

A separate concern is overreliance on incorrect AI advice. In a randomized vignette study, Jabbour et al. found that systematically biased AI recommendations reduced clinicians' diagnostic accuracy, and visual explanations did not eliminate this effect \cite{jabbour2023impact}. Safety evaluation should therefore account not only for factual or therapeutic errors in an LLM-generated response, but also for the possibility that a plausibly formulated second opinion may shift a physician's decision in the wrong direction.

LLM-generated explanations can themselves influence clinical acceptance: Umerenkov et al. found that explanations increased physicians' agreement with presented diagnoses while also containing nontrivial errors \cite{umerenkov2023deciphering}. Persuasiveness and clinical correctness should therefore be evaluated separately.

\subsection{LLM-as-a-Judge and Physician Expert Evaluation}

Automated evaluation of free-form medical responses is attractive because physician review is expensive and variable, but the available evidence does not support treating LLM-based evaluation as an unconditional surrogate for ground truth. Croxford et al. reported high agreement between LLM-as-a-judge and human evaluation in the relatively constrained task of clinical summarization \cite{croxford2025judge}. Zhou et al. likewise investigated automated expert-level evaluation of medical reasoning \cite{zhou2026medthink}. More recent studies, however, suggest substantial task- and criterion-dependence. In a multicenter study of real de-identified cases, ratings from more than 400 physicians differed substantially from matched AI evaluators \cite{shi2026physicianjudge}. Williams et al. compared five LLM judges with clinicians and found that even the best-performing judge reproduced human ratings for only a subset of eleven criteria, while jury-style aggregation produced only limited improvements in agreement \cite{williams2026judge}.

These findings motivate a more conservative validation framework: physicians remain the primary expert reference, while an LLM judge should be validated against physician ratings separately for each evaluation criterion. Agreement on one dimension, such as factual completeness, should not be assumed to generalize to treatment appropriateness, safety, or uncertainty calibration.

\subsection{Positioning of the Present Study}

The closest prior studies address individual pairs of components in our setting, but we did not identify a published study that, on the same set of real published clinical cases, evaluates standalone role-specific second opinions and subsequent multidisciplinary synthesis while independently varying response structure, sampling and aggregation, and council interaction, and while validating an LLM judge against physicians criterion by criterion. Sonoda et al. provide a direct precedent for a same-case free-form versus structured comparison, whereas MAC, MDAgents, and other medical multi-agent systems compare single- and multi-agent variants without exposing all intermediate opinions as benchmark units or fully separating interaction from inference budget \cite{sonoda2025structured,chen2025mac,kim2024mdagents}. Accordingly, the contribution of the present study is not a presumption that multidisciplinary deliberation is intrinsically superior, but an empirical decomposition of the stages and design choices that produce a clinical second opinion.

\section{Methods}

\subsection{Study Design}

We conducted a benchmark-development and evaluation study using published, diagnostically resolved clinical cases. The generation pipeline comprised three stages: (i) case-specific assignment of clinical roles, (ii) independent role-specific analysis before consilium (the \emph{pre-consilium} stage), and (iii) multidisciplinary synthesis (the \emph{consilium} stage). The dataset's separate \texttt{Answer} field was not supplied to the generation models and was reserved for evaluation.

\begin{figure}[tb]
\centering
\begin{tikzpicture}[
  font=\small,
  stage/.style={
    rectangle,
    rounded corners=2.5pt,
    draw=blue!55!black,
    line width=0.65pt,
    align=center,
    text width=2.72cm,
    minimum height=18mm,
    inner sep=4pt,
    fill=blue!4
  },
  checkpoint/.style={
    stage,
    draw=teal!65!black,
    fill=teal!7
  },
  component/.style={
    rectangle,
    rounded corners=2pt,
    draw=teal!65!black,
    line width=0.6pt,
    align=center,
    text width=4.15cm,
    minimum height=12mm,
    inner sep=4pt,
    fill=teal!5
  },
  scorebox/.style={
    rectangle,
    rounded corners=2pt,
    draw=orange!70!black,
    line width=0.65pt,
    align=center,
    text width=14.55cm,
    minimum height=10mm,
    inner sep=4pt,
    fill=orange!7
  },
  label/.style={
    font=\scriptsize\bfseries,
    text=blue!55!black,
    anchor=west
  },
  flow/.style={-{Latex[length=2mm]}, line width=0.9pt, draw=blue!65!black},
  assess/.style={-{Latex[length=1.8mm]}, line width=0.65pt, draw=teal!65!black}
]
\node[label] at (-7.85,1.38) {GENERATION PIPELINE};

\node[stage, draw=black!55, fill=black!2] (case) at (-6.35,0) {
  \textbf{Clinical case corpus}\\[2pt]
  300 resolved NEJM--MedQA cases\\[-1pt]
  {\scriptsize 110 calibration \textbullet{} 190 held-out}
};

\node[stage] (roles) at (-3.18,0) {
  \textbf{Role assignment}\\[2pt]
  1--5 case-specific roles\\[-1pt]
  {\scriptsize one primary-role label}
};

\node[checkpoint] (pre) at (0,0) {
  \textbf{Independent opinions}\\[2pt]
  role-specific analysis\\[-1pt]
  {\scriptsize pre-consilium checkpoint}
};

\node[stage, fill=blue!8] (council) at (3.18,0) {
  \textbf{Interactive consilium}\\[2pt]
  multidisciplinary review,\\[-1pt]
  revision, and consensus
};

\node[checkpoint] (final) at (6.35,0) {
  \textbf{Final synthesis}\\[2pt]
  case-level consilium report\\[-1pt]
  {\scriptsize final checkpoint}
};

\draw[flow] (case) -- (roles);
\draw[flow] (roles) -- (pre);
\draw[flow] (pre) -- (council);
\draw[flow] (council) -- (final);

\node[label, text=teal!65!black] at (-7.85,-1.70) {PHYSICIAN EVALUATION};

\node[component] (part1) at (-4.55,-2.62) {
  \textbf{Part 1}\\
  role-specific clinical reasoning
};
\node[component] (part2) at (0,-2.62) {
  \textbf{Part 2}\\
  role-specific clinical opinion
};
\node[component] (part3) at (4.55,-2.62) {
  \textbf{Part 3}\\
  final consilium conclusion
};

\coordinate (prebranch) at (0,-1.38);
\coordinate (finalbranch) at (6.35,-1.55);
\draw[assess] (pre.south) -- (prebranch) -| (part1.north);
\draw[assess] (prebranch) -- (part2.north);
\draw[assess] (final.south) -- (finalbranch) -| (part3.north);

\node[scorebox] (scores) at (0,-4.15) {
  \textbf{Eight response-quality criteria (0--3)}\\[-1pt]
  {\scriptsize applied independently to each component}
};

\draw[assess] (part1.south) -- (part1.south |- scores.north);
\draw[assess] (part2.south) -- (scores.north);
\draw[assess] (part3.south) -- (part3.south |- scores.north);
\end{tikzpicture}

\caption{Generation and physician-evaluation workflow. Case-specific roles produced independent pre-consilium opinions before interactive multidisciplinary synthesis. Physicians scored role-specific clinical reasoning, role-specific clinical opinion, and the final consilium conclusion as three separate components using the same eight 0--3 criteria. Panel adequacy, case complexity, and overall quality across all stages were rated separately.}
\label{fig:study-design}
\end{figure}
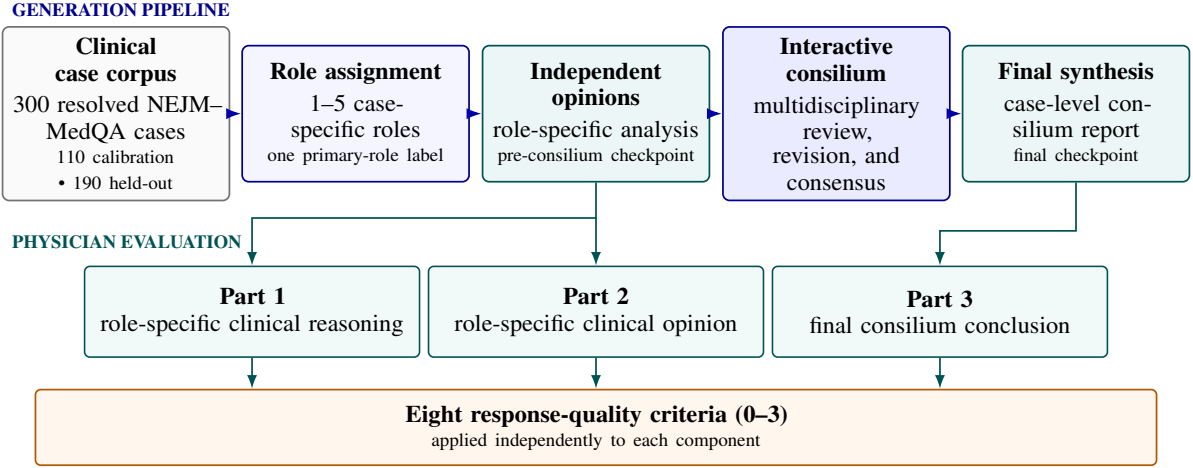

Physician evaluation comprised three independently scored components (Figure~\ref{fig:study-design}). First, physicians evaluated the clinical reasoning within the relevant role-specific pre-consilium output, excluding its clinical opinion. Second, they evaluated the clinical opinion of the same role-specific output. Third, they evaluated the final case-level consilium conclusion. The same eight quality criteria were applied independently to each component using a 0-3 scale, a score assigned to one component was not carried over to another. Three additional ratings-panel adequacy, case complexity, and overall quality across all stages - were assigned once per case assessment.

The reference pipeline supported two pre-consilium prompt conditions---free-form narrative responses and structured, schema-constrained responses with expanded clinical-reasoning guidance---and two inference topologies (single and $N+1$). All multidisciplinary synthesis used the interactive consilium protocol described below. The conditions included in each analysis were specified prospectively. Model identifiers, prompts, generation parameters, and prerequisite artifacts were recorded for reproducibility.

\subsection{NEJM Clinical Case Dataset, Selection Rationale, and Prespecified Split}
\label{sec:clinical-case-dataset}

We used the \texttt{nejm\_test} configuration of NEJM--MedQA, which mirrors the supplementary data of Savage et al. \citep{savage2024reasoning,link2024nejmmedqa}. The source dataset includes 300 extended clinicopathological cases selected from 310 recent Case Records after excluding ten without a definitive diagnosis or within the applicable context limit \citep{brinkmann2024caserecords,savage2024reasoning}. We used 300 DOI-unique cases published in 2015--2023. Case descriptions contained a median of 1,036 words (IQR, 852--1,197; range, 231--2,167).

We retained the clinical presentation and normalized metadata but removed source-study instructions, examples, prior model responses, and grades. The separate \texttt{Answer} field was withheld during generation and used only for evaluation. Role assignment and both response conditions received the same DOI, title, and case description; the consilium received an identifier-stripped case, the fixed role opinions, and the review history.

We chose this corpus because its long, diagnostically resolved narratives support open-ended assessment of reasoning, management, and safety, not only diagnostic agreement. The published diagnosis was used for diagnostic concordance but not as exhaustive ground truth for management. Because Case Records favor unusual and difficult cases, results should not be interpreted as performance in routine practice \citep{kanjee2023gpt4,savage2024reasoning,mcduff2025amie,shao2025verification}.

Before evaluation, we reused a prespecified fixed DOI manifest defining 110 cases for physician annotation and Judge calibration, the 190-case held-out evaluation cohort was its exact set complement, with source-dataset order preserved. Here, ``held-out'' means that the 190 evaluation cases were not used for physician annotation or Judge calibration in this study.

\subsection{Role Assignment And Independent Role-Specific Second Opinions}

For each case, the configured role-assignment procedure selected between one and five clinical roles, designated one as primary, and provided a brief case-specific rationale for each role and for the panel as a whole. Exact duplicate role names were rejected case-insensitively, and the primary role was required to match one of the selected roles. The primary-role designation was retained as metadata and did not alter subsequent generation: every selected role produced an opinion. Role-specific prompts received the role name but not its selection rationale or primary-role status.

Every selected role received the same case input and produced its opinion without seeing the opinions produced for other roles. Consequently, each output could be read and evaluated independently as a second opinion from that prompted clinical perspective. The role-specific outputs were fixed before the consilium and were not regenerated during discussion.

In the free-form condition, the model produced a self-contained clinical opinion and was instructed to state missing information and assumptions. In the structured condition, the required fields were: subjective and objective information, clinical syndromes and involved mechanisms, differential diagnosis with reasoning, diagnostic and treatment tactics and final clinical opinion. The structured condition combined a prespecified response schema with more detailed clinical-reasoning instructions. Comparisons with the free-form condition therefore reflected the overall difference between the two prompting approaches, rather than the isolated effect of response format.

\subsection{Interactive Multidisciplinary Consilium Protocol}

We use \emph{consilium} to denote structured multidisciplinary review of a clinical case. The reference workflow adapted the five-stage logic of MedAgents from multiple-choice medical question answering to generation of a self-contained clinical second opinion \citep{tang2024medagents}. It was an adaptation rather than a literal reproduction: the output was an open-ended clinical report, roles addressed both diagnosis and management, and all role-specific opinions were retained.

The interactive protocol first synthesized the case and fixed role-specific opinions into a shared report. In each round, every role independently reviewed the current report and returned either a \texttt{yes} vote or a \texttt{no} vote with a clinically relevant critique and proposed revision. From the second round onward, reviewers also received the reviews and moderator decision from the immediately preceding round. If every role voted \texttt{yes}, the controller recorded consensus. Otherwise, the moderator considered the objections, checked them against the case, and either revised the report or recorded that no substantive change was warranted.

The cycle ended when all roles accepted the report, the moderator reported no substantive change, a revised report exactly repeated a previously seen report state, or five review rounds were completed. When dissent persisted through the fifth review and the moderator made a substantive revision, that revision was retained at the maximum-round stop without a sixth role vote. The finalizer then received the case, all fixed role-specific opinions, the last retained report, all specialist reviews, and summaries of the moderator's revision decisions. It produced a self-contained final opinion while retaining unresolved disagreement and uncertainty. The controller recorded the number of completed rounds, consensus status, and stop reason. The full protocol is specified in Algorithm~\ref{alg:interactive}.

The final free-form output was a self-contained narrative opinion. The structured output contained a consilium summary, resolved and unresolved disagreements, final differential diagnosis, diagnostic and management plans, safety warnings, final second opinion, and confidence level. Structured voting and termination records were used in both response formats. Complete interactive-consilium reports for the same calibration case are provided for the free-form and structured conditions in Appendices~\ref{app:example-report-free} and~\ref{app:example-report-structured}, respectively.

\subsection{Construction of the Physician-Annotation Corpus}
The physician-annotation corpus comprised 110 cases. Each case was processed through both the free-form and structured pipelines using the same finalized role panel.

Before generating the corpus, we screened open-weight models offered through OpenRouter in July 2026 \citep{openrouter2026models}. Eligibility required public weights, support for structured outputs or reasoning, sufficient context for the longest prompts in our pipeline, reliable provider access, and a feasible projected cost. We selected the four highest-ranked eligible models when the OpenRouter catalogue was sorted by the Artificial Analysis Intelligence Index.

GLM-5.2 \citep{zai2026glm52} ranked highest and combined a long context window with broad provider availability, so it was assigned to combine the workers' outputs and conduct the interactive consilium. MiniMax M3 and DeepSeek-V4-Pro \citep{minimax2026m3,deepseek2026v4pro} were the next two eligible models and offered long context windows at comparatively low inference cost. Kimi K2.6 \citep{kimi2026k26} was the next-ranked eligible model and added a third independent developer and model family to the worker panel. Thus, the strongest eligible model performed the final synthesis, while the next three provided independent opinions without relying on a single model family. Role assignment and role-specific pre-consilium generation used a 3+1 procedure: the three worker models independently completed each task, and GLM-5.2 consolidated their outputs.
In both pipelines, the pre-consilium opinions were then passed to the interactive consilium. GLM-5.2 generated one response at each step, with a maximum of five review rounds. This produced 110 final consilium conclusions per pipeline.

\subsection{Single-Model Evaluation}

The 190-case held-out cohort was generated separately with \ConsModelCountWord{} candidate models: \ConsModelList{}. Within each model-specific configuration, the same candidate model performed role assignment, generated every role-specific opinion, and conducted every review, revision, and finalization call. The role assignment for a case was reused by its free-form and structured pipelines.

All models received the same ordered held-out case set and the same stage- and response-format-specific prompts and schemas. Generation used temperature~0, top-$p$~1, seed~42, and at most five review rounds. Model-specific maximum output allowances were retained: 1,024, 2,048, or 4,096 tokens for role assignment, 16,000 or 16,384 tokens for pre-consilium generation and 16,000 or 32,768 tokens for the interactive consilium. Endpoint routing and additional decoding controls were also fixed per model and recorded in the run manifests.

The descriptive process and Judge-scored analyses used these \ConsModelCountWord{} configurations. Role-level Judge scoring was performed only for the primary role in the structured pipeline, whereas final consilium conclusions were scored in both the free-form and structured pipelines; role-level Judge scores were not produced for free-form outputs.

\subsection{Evaluation Criteria}

We used a physician-developed measurement framework comprising eight response-quality criteria and three separately interpreted panel- or case-level ratings (Table~\ref{tab:criteria}). The framework was informed by literature on multidisciplinary clinical review~\citep{lamb2011mdt,basta2017mdt,koco2021mdt,soukup2020mdt}, clinical LLM assessment~\citep{tang2024medagents,kocaman2025clever,mccoy2025assessment,brodeur2026performance}, uncertainty~\citep{savage2025uncertainty}, and safety~\citep{asgari2025framework}. The eight response-quality criteria covered factual grounding, problem prioritization, data use, reasoning coherence, differential diagnosis, the final clinical conclusion, management, and safety. Their applicability was determined by the component being assessed.

The eight response-quality criteria used a 0--3 ordinal scale: 0 - indicated that the criterion was not met, 1 - a major deficiency, 2 - a mostly adequate response with a noticeable limitation and 3 - full and clinically convincing satisfaction. Panel adequacy used the same numeric range but evaluated expertise coverage rather than response quality. Case complexity used 0, 1, 2, and 3 to denote low, moderate, high, and very high complexity. Overall quality was a physician-reported global assessment of all consilium stages.

To account for characteristic LLM errors, an additional rule was introduced. If the final output contained a fabricated clinically significant fact - such as an examination result, numerical value, date, diagnosis, complication, or treatment recommendation absent from the source documents and not designated as a hypothesis - the ``Factual accuracy and sources'' criterion was assigned 0 points. If such a fact served as the basis for differential diagnosis, the final conclusion, or management strategy, the corresponding criteria were scored no higher than 1. This rule ensured that critical factual errors could not be offset by apparent logical coherence, structural completeness, or formatting quality.


\begin{longtable}{@{}>{\raggedright\arraybackslash}p{0.05\textwidth}>{\raggedright\arraybackslash}p{0.29\textwidth}>{\raggedright\arraybackslash}p{0.59\textwidth}@{}}
\caption{Prespecified evaluation criteria.}\label{tab:criteria}\\
\toprule
\textbf{No.} & \textbf{Criterion} & \textbf{Evaluation target} \\
\midrule
\endfirsthead
\toprule
\textbf{No.} & \textbf{Criterion} & \textbf{Evaluation target} \\
\midrule
\endhead
1 & Factual Correctness and Source Grounding & Agreement of clinically material claims with the supplied case, without fabricated patient facts or consequential distortion. \\
2 & Identification and Prioritization of Clinical Problems & Recognition of the principal and decision-relevant problems and their clinically justified ordering. \\
3 & Completeness and Relevance of Data Use & Integration of supplied information that affects conclusions or management, without penalizing information absent from the input. \\
4 & Coherence of Clinical Reasoning & Validity of links from confirmed facts to conclusions, including mechanisms, chronology, alternatives, and uncertainty. \\
5 & Quality of the Differential Diagnosis & Relevance and prioritization of hypotheses, evidence for and against them, and concrete discriminating checks. \\
6 & Quality of the Final Clinical Conclusion & Clarity, responsiveness, evidence grounding, agreement with the reference diagnosis, and calibration to uncertainty. \\
7 & Justification and Specificity of Management & Prioritization, specificity, timing, justification, monitoring, and revision conditions for diagnostic and therapeutic actions. \\
8 & Clinical Safety & Handling of urgent risks, contraindications, interactions, complications, monitoring, and escalation of care. \\
9 & Adequacy of Consilium Composition and Expertise Coverage & Complementary coverage of the main problem, key decisions, and risks without a material expertise gap. For panel outputs only. \\
10 & Clinical Case Complexity & Complexity of the source case based on competing hypotheses, interacting problems, incomplete or conflicting data, and multidisciplinary uncertainty. Analyzed separately. \\
11 & Overall Quality of All Consilium Stages & Global quality of the complete consilium process, including the coherence of its stages and the clarity and clinical justification of the final decision. Analyzed separately. \\
\bottomrule
\end{longtable}

\subsection{Physician Reference Evaluation}

Physician ratings provided the reference for Judge calibration and internal evaluation. The evaluation panel comprised 6 physicians, each possessing cross-disciplinary expertise in multiple adjacent medical specialties. This configuration ensured comprehensive coverage of specialty-specific evaluation perspectives, composition of the multidisciplinary panels across the dataset. Each clinical case was evaluated from the standpoint of three specialties represented in the original consensus. Experts were exclusively assigned cases aligning with their clinical domain.

Evaluation was conducted under single-blind conditions: model outputs were randomly ordered and presented without model identifiers, generation condition labels, or paired responses from the alternative condition. Case presentations were delivered without source attribution or retrieval metadata, strictly adhering to the input protocol used for model inference.

Raters used the fixed 0--3 rubric in Table~\ref{tab:criteria}. Eight criteria were applied separately to clinical reasoning, the role-specific clinical opinion, and the final consilium report (24 fields). Three further fields covered panel adequacy, case complexity, and overall quality. Raters also entered a case comment and could use \emph{non-evaluable} only where the rubric allowed.

\subsection{Automated LLM Judge}

We used GPT-OSS-120B as a physician-calibrated automated Judge \citep{openai2025gptoss}. Prompts combined a fixed protocol, the criterion definition and four score anchors, an applicability rule, and an optional criterion-specific instruction. The case and candidate response were delimited from the instructions to reduce prompt injection. During calibration, only the optional instructions for criteria~1--9 varied. Each call received the case and one complete target component, but no other role outputs or later reports. The reference diagnosis was visible only for criterion~6 and could not support patient facts absent from the presentation. Criterion~11 remained physician reported and was not scored by the Judge.

Each applicable call returned a rationale and either a 0--3 score, \emph{not evaluable}, or \emph{model failure}; infrastructure failures were never converted to zero. Calls used an OpenAI-compatible endpoint with temperature~0, top-$p$~1, and an 8,000-token output limit.

The reasoning component was assessed on criteria~1--5, 7, and~8; the role-specific clinical opinion on criteria~1, 2, 4, 6, and~8; and the final consilium report on criteria~1--8. Criteria~9 and~10 were analyzed separately. We reported unweighted means and numeric coverage within each component. Structural exclusions and \emph{not evaluable} outcomes were not scored, and means based on different criterion sets were not compared as estimates of checkpoint improvement.

\subsection{Exploratory Analysis of Completion Rounds and Judge Scores}
\label{methods:judge-rounds}

We performed a descriptive analysis of whether final-answer Judge scores varied with the number of completed consilium review rounds. The analysis included \ConsTotalArms{} outputs from 190 cases, \ConsModelCountWord{} models, and two response formats. Because rounds~4 and~5 were uncommon, outputs were grouped as one, two, or at least three completed rounds. For each group, we report the mean and population standard deviation (SD) of the output-level mean across numeric scores for criteria~1--8, the proportion of outputs with a mean below~2, and the criterion~10 case-complexity score. Criterion~10 was not included in the response-quality mean. The round count described review of the retained report, whereas the Judge assessed the subsequent final answer and did not receive the round count or stopping reason.

\subsection{Calibration of Judge Guidance}

We optimized one operational prompt, $g_k$, for each criterion $k\in\{1,\ldots,9\}$ using GigaEvo and MAP-Elites \citep{khrulkov2025gigaevo,mouret2015mapelites,lehman2022evolution,lim2024llmqd}. The general Judge instructions and all criterion definitions, scales, applicability rules, schemas, and reference policies were fixed. Candidate prompts were limited to 800 characters and could not migrate across criteria.

The prespecified 110-case sampling frame was partitioned by DOI into 80 evolutionary-training cases and 30 internal-validation cases. All completed targets and ratings from one DOI remained in the same subset. Evolution used only eligible structured targets and physician ratings from the training subset. Ratings from the internal-validation subset were withheld from mutation and parent selection.

Let $\Omega_{k,\mathrm{train}}$ denote the set of training case--target--physician-rating observations applicable to criterion $k$, and let $\mathcal{R}=\{42,43,44\}$ denote the prespecified Judge repeat seeds. For a numeric Judge score, the item-level loss was
\begin{equation}
e_{dk}^{(r)}(g_k)=\left|\widehat{y}_{dk}^{(r)}(g_k)-y_{dk}\right|.
\end{equation}
The criterion-specific objective was
\begin{equation}
\mathcal{L}_{k}(g_k)=
\frac{1}{|\mathcal{R}|}\sum_{r\in\mathcal{R}}
\frac{1}{|\Omega_{k,\mathrm{train}}|}
\sum_{d\in\Omega_{k,\mathrm{train}}}e_{dk}^{(r)}(g_k).
\label{eq:criterion-loss}
\end{equation}
Thus, no score or loss from another criterion entered the optimization of $g_k$.

MAE expresses Judge--physician disagreement directly in scale points and is less sensitive to large errors than squared loss \citep{gneiting2011pointforecasts,willmott2005mae}. It is commonly used for ordinal prediction \citep{baccianella2009ordinal}, but assumes meaningful distances between score categories and does not fully characterize agreement; we therefore treated it as an error summary, not a complete validity measure.

The search used nine isolated criterion archives (capacity 540), initialized with two candidates per criterion. GPT-OSS-120B generated mutations at temperature~0.3 with a 4,096-token limit. We retained 225 mutants (25 per criterion) from at most 320 attempts. For each criterion, the lowest-MAE prompt was selected, with ties resolved by repeat stability, length, and program identifier. Selected prompts were re-evaluated on training and internal-validation cases with fresh seeds and compared with no-guidance, initial-prompt, and training-median baselines before semantic review.

Judge calibration was summarized by criterion-specific MAE against individual physician ratings, repeat-to-repeat SD, numeric coverage, and signed error where reported. The held-out clinical-quality analyses were descriptive. We retained the original 0--3 criterion scores and reported numeric means, population SDs, outcome distributions, and numeric coverage. No weighted composite, 0--100 transformation, readiness threshold, or single confirmatory primary quality outcome was used. Because evaluation components had different structurally applicable criterion sets, their pooled means were not treated as directly exchangeable estimates of checkpoint improvement.

Response-format and model summaries used the fixed model roster described above. Numeric means excluded structurally inapplicable fields and response-dependent \emph{not evaluable} outcomes; both were reported separately and neither was assigned zero. Model failures were likewise retained as failures rather than converted to clinical scores. The process analyses used DOI-clustered bootstrap intervals as specified above. Held-out Judge-score tables were used for descriptive comparison rather than confirmatory model ranking or multiplicity-adjusted hypothesis testing.


\raggedbottom
\clearpage
\section{Results}


\subsection{Evaluation Cases Composition}

\begin{table}[!ht]
\caption{Characteristics of the physician-annotation/Judge-calibration and held-out evaluation cohorts.}\label{tab:cohort-comparison}
\centering
\footnotesize
\setlength{\tabcolsep}{3.5pt}
\begin{tabular}{@{}p{0.32\linewidth}p{0.25\linewidth}p{0.25\linewidth}r@{}}
\toprule
\textbf{Characteristic} & \textbf{Annotation/Judge calibration} & \textbf{Held-out evaluation} & \textbf{$|$SMD$|$} \\
\midrule
Case records / patients & 110 / 110 & 190 / 191 & - \\
Age, median [IQR]; range, years & 44.0 [26.3--60.8]; 0--89 & 48.0 [29.0--64.5]; 0--84 & 0.107 \\
\quad Younger than 18 years & 15/110 (13.6\%) & 23/191 (12.0\%) & 0.048 \\
\quad 18--64 years & 76/110 (69.1\%) & 120/191 (62.8\%) & 0.132 \\
\quad 65 years or older & 19/110 (17.3\%) & 48/191 (25.1\%) & 0.193 \\
Male & 59/110 (53.6\%) & 105/191 (55.0\%) & 0.027 \\
Female & 51/110 (46.4\%) & 86/191 (45.0\%) & 0.027 \\
Publication year, median [IQR]; range & 2018 [2016--2019]; 2015--2020 & 2020.5 [2018--2022]; 2015--2023 & 1.177 \\
Description length, median [IQR]; range, words & 1,036 [826--1,166]; 231--1,805 & 1,035 [866--1,202]; 295--2,167 & 0.082 \\
\bottomrule
\end{tabular}
\begin{flushleft}
\footnotesize Age and reported male/female descriptors use patients as the denominator, publication year and description length use case records. No age, descriptor, publication-year, or case-description values were missing. Continuous-variable SMDs use means and pooled standard deviations although medians and IQRs are displayed, binary-variable SMDs are standardized prevalence differences. SMDs are descriptive and no cohort-comparison hypothesis tests were performed.
\end{flushleft}
\end{table}

All models completed both response-format pipelines for all 190 held-out cases (100\% coverage). The \ConsPanels{} unique case--model panels contained \ConsRoleSlots{} role slots (mean, \ConsMeanRoles{} per panel); reuse of each panel across the free-form and structured conditions yielded \ConsRoleOutputsTotal{} role-specific opinions and \ConsTotalArms{} final consilium opinions. These artifact counts describe coverage, whereas the DOI remains the inferential unit.

Panel selection was strongly concentrated at the configured ceiling. Among \ConsPanels{} case--model panels, \ConsTwoRolePanels{} (\ConsTwoRolePanelsPct\%) contained two roles, \ConsThreeRolePanels{} (\ConsThreeRolePanelsPct\%) contained three, \ConsFourRolePanels{} (\ConsFourRolePanelsPct\%) contained four, and \ConsFiveRolePanels{} (\ConsFiveRolePanelsPct\%) contained five (Figure~\ref{fig:panel-architecture}b). \ConsModelsMedianFive{} models had a median of five roles, Gemma-4-31B had a median of four. Because the selector could assign at most five roles, this concentration shows that most configurations frequently exhausted the panel capacity, not that five was an empirically optimal panel size, it also provided little empirical variation for assessing whether panel cardinality adapted to the case.

The \ConsRoleSlots{} assignments used \ConsExactLabels{} exact model-generated role labels, including \ConsPrimaryExactLabels{} exact labels among the \ConsPanels{} primary designations. This lexical variety largely reflected differences in naming granularity rather than hundreds of clinically distinct disciplines. After mutually exclusive harmonization, infectious disease, radiology, and neurology were the most frequent categories, whereas neurology (\ConsNeurologyPrimary{}/\ConsPanels{}, \ConsNeurologyPrimaryPct\%) and infectious disease (\ConsInfectiousPrimary{}/\ConsPanels{}, \ConsInfectiousPrimaryPct\%) were the most frequent primary roles (Figure~\ref{fig:panel-architecture}c). Radiology-containing labels appeared in \ConsRadiologyAll{} assignments but received only \ConsRadiologyPrimary{} primary designations (\ConsRadiologyPrimaryAmongAssignmentsPct\% of radiology assignments), indicating that the selector usually prioritized another discipline as the principal case perspective.

\definecolor{consBlue}{HTML}{0072B2}
\definecolor{consOrange}{HTML}{D55E00}
\definecolor{consTeal}{HTML}{009E73}
\definecolor{consDark}{HTML}{3F4A4F}
\definecolor{consLight}{HTML}{F3F5F6}
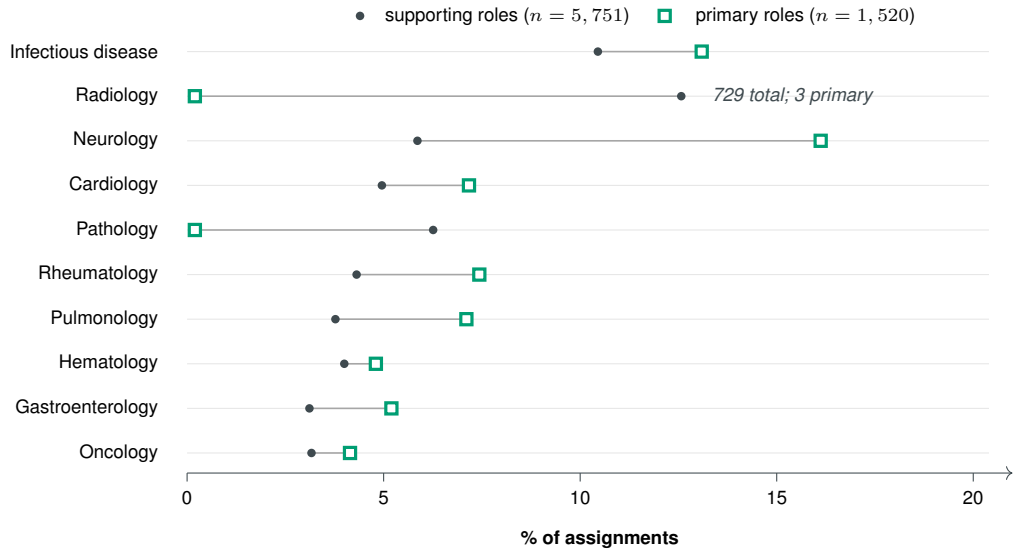
\begin{figure}[p]
\centering
\makebox[\linewidth][l]{\textbf{a}\hspace{0.6em}\sffamily\textbf{Cohort and output flow}}\par\smallskip
\resizebox{\linewidth}{!}{%
\begin{tikzpicture}[x=1cm,y=1cm,font=\sffamily\scriptsize,box/.style={rectangle,rounded corners=2pt,draw=consDark,align=center,minimum height=1.0cm,fill=consLight},arr/.style={-{Latex[length=1.8mm]},semithick,draw=consDark}]
\node[box,text width=2.05cm] (cases) at (0,0) {300 DOI-unique\\case records};
\node[box,text width=2.25cm,draw=black!45,fill=black!4] (calibration) at (3.0,0.72) {110 annotation/\\Judge calibration};
\node[box,text width=2.25cm,draw=consBlue,fill=consBlue!6] (heldout) at (3.0,-0.72) {190 held-out\\case records};
\node[box,text width=2.55cm,draw=consBlue,fill=consBlue!6] (panels) at (6.55,-0.72) {8 models\\1,520 shared panels\\7,271 role slots (mean 4.78)};
\node[box,text width=2.65cm,draw=consBlue,fill=consBlue!6] (free) at (10.35,0.18) {Free-form pipeline\\7,271 role outputs\\1,520 final opinions};
\node[box,text width=2.65cm,draw=consOrange,fill=consOrange!7] (structured) at (10.35,-1.62) {Structured pipeline\\7,271 role outputs\\1,520 final opinions};
\node[box,text width=2.35cm,draw=consTeal,fill=consTeal!7] (final) at (14.25,-0.72) {3,040 final opinions\\100\% complete};
\draw[arr] (cases.east) -- ++(0.45,0) |- (calibration.west);
\draw[arr] (cases.east) -- ++(0.45,0) |- (heldout.west);
\draw[arr] (heldout) -- (panels);
\draw[arr] (panels.east) -- ++(0.45,0) |- (free.west);
\draw[arr] (panels.east) -- ++(0.45,0) |- (structured.west);
\draw[arr] (free.east) -- ++(0.42,0) |- (final.west);
\draw[arr] (structured.east) -- ++(0.42,0) |- (final.west);
\end{tikzpicture}%
}\par\medskip
\makebox[\linewidth][l]{\textbf{b}\hspace{0.6em}\sffamily\textbf{Panel-size distribution by model}}\par\smallskip
\makebox[\linewidth][c]{%
\begin{tikzpicture}[x=1.20cm,y=0.60cm,font=\sffamily\scriptsize]
\node[anchor=east,font=\sffamily\scriptsize\bfseries] at (3.75,9.35) {Model};
\node[font=\sffamily\scriptsize\bfseries] at (4.925,9.35) {2 roles};
\node[font=\sffamily\scriptsize\bfseries] at (6.675,9.35) {3 roles};
\node[font=\sffamily\scriptsize\bfseries] at (8.425,9.35) {4 roles};
\node[font=\sffamily\scriptsize\bfseries] at (10.175,9.35) {5 roles};
\node[anchor=east] at (3.75,8.00) {DeepSeek-V4-Pro};
\path[fill=consBlue!8,draw=white,line width=0.4pt,rounded corners=1pt] (4.050,7.620) rectangle (5.760,8.380);
\node[text=consDark,font=\sffamily\scriptsize\bfseries] at (4.905,8.000) {0.0\%};
\path[fill=consBlue!8,draw=white,line width=0.4pt,rounded corners=1pt] (5.800,7.620) rectangle (7.510,8.380);
\node[text=consDark,font=\sffamily\scriptsize\bfseries] at (6.655,8.000) {0.5\%};
\path[fill=consBlue!9,draw=white,line width=0.4pt,rounded corners=1pt] (7.550,7.620) rectangle (9.260,8.380);
\node[text=consDark,font=\sffamily\scriptsize\bfseries] at (8.405,8.000) {1.1\%};
\path[fill=consBlue!89,draw=white,line width=0.4pt,rounded corners=1pt] (9.300,7.620) rectangle (11.010,8.380);
\node[text=white,font=\sffamily\scriptsize\bfseries] at (10.155,8.000) {98.4\%};
\node[anchor=east] at (3.75,7.00) {Gemma-4-31B};
\path[fill=consBlue!11,draw=white,line width=0.4pt,rounded corners=1pt] (4.050,6.620) rectangle (5.760,7.380);
\node[text=consDark,font=\sffamily\scriptsize\bfseries] at (4.905,7.000) {4.2\%};
\path[fill=consBlue!27,draw=white,line width=0.4pt,rounded corners=1pt] (5.800,6.620) rectangle (7.510,7.380);
\node[text=consDark,font=\sffamily\scriptsize\bfseries] at (6.655,7.000) {22.6\%};
\path[fill=consBlue!52,draw=white,line width=0.4pt,rounded corners=1pt] (7.550,6.620) rectangle (9.260,7.380);
\node[text=consDark,font=\sffamily\scriptsize\bfseries] at (8.405,7.000) {53.7\%};
\path[fill=consBlue!24,draw=white,line width=0.4pt,rounded corners=1pt] (9.300,6.620) rectangle (11.010,7.380);
\node[text=consDark,font=\sffamily\scriptsize\bfseries] at (10.155,7.000) {19.5\%};
\node[anchor=east] at (3.75,6.00) {GLM-5.2};
\path[fill=consBlue!8,draw=white,line width=0.4pt,rounded corners=1pt] (4.050,5.620) rectangle (5.760,6.380);
\node[text=consDark,font=\sffamily\scriptsize\bfseries] at (4.905,6.000) {0.0\%};
\path[fill=consBlue!10,draw=white,line width=0.4pt,rounded corners=1pt] (5.800,5.620) rectangle (7.510,6.380);
\node[text=consDark,font=\sffamily\scriptsize\bfseries] at (6.655,6.000) {2.1\%};
\path[fill=consBlue!26,draw=white,line width=0.4pt,rounded corners=1pt] (7.550,5.620) rectangle (9.260,6.380);
\node[text=consDark,font=\sffamily\scriptsize\bfseries] at (8.405,6.000) {21.6\%};
\path[fill=consBlue!71,draw=white,line width=0.4pt,rounded corners=1pt] (9.300,5.620) rectangle (11.010,6.380);
\node[text=white,font=\sffamily\scriptsize\bfseries] at (10.155,6.000) {76.3\%};
\node[anchor=east] at (3.75,5.00) {GPT-5.6-Sol};
\path[fill=consBlue!8,draw=white,line width=0.4pt,rounded corners=1pt] (4.050,4.620) rectangle (5.760,5.380);
\node[text=consDark,font=\sffamily\scriptsize\bfseries] at (4.905,5.000) {0.0\%};
\path[fill=consBlue!10,draw=white,line width=0.4pt,rounded corners=1pt] (5.800,4.620) rectangle (7.510,5.380);
\node[text=consDark,font=\sffamily\scriptsize\bfseries] at (6.655,5.000) {2.1\%};
\path[fill=consBlue!28,draw=white,line width=0.4pt,rounded corners=1pt] (7.550,4.620) rectangle (9.260,5.380);
\node[text=consDark,font=\sffamily\scriptsize\bfseries] at (8.405,5.000) {24.7\%};
\path[fill=consBlue!68,draw=white,line width=0.4pt,rounded corners=1pt] (9.300,4.620) rectangle (11.010,5.380);
\node[text=white,font=\sffamily\scriptsize\bfseries] at (10.155,5.000) {73.2\%};
\node[anchor=east] at (3.75,4.00) {GPT-OSS-120B};
\path[fill=consBlue!8,draw=white,line width=0.4pt,rounded corners=1pt] (4.050,3.620) rectangle (5.760,4.380);
\node[text=consDark,font=\sffamily\scriptsize\bfseries] at (4.905,4.000) {0.0\%};
\path[fill=consBlue!8,draw=white,line width=0.4pt,rounded corners=1pt] (5.800,3.620) rectangle (7.510,4.380);
\node[text=consDark,font=\sffamily\scriptsize\bfseries] at (6.655,4.000) {0.0\%};
\path[fill=consBlue!8,draw=white,line width=0.4pt,rounded corners=1pt] (7.550,3.620) rectangle (9.260,4.380);
\node[text=consDark,font=\sffamily\scriptsize\bfseries] at (8.405,4.000) {0.5\%};
\path[fill=consBlue!90,draw=white,line width=0.4pt,rounded corners=1pt] (9.300,3.620) rectangle (11.010,4.380);
\node[text=white,font=\sffamily\scriptsize\bfseries] at (10.155,4.000) {99.5\%};
\node[anchor=east] at (3.75,3.00) {Qwen3.5-27B};
\path[fill=consBlue!8,draw=white,line width=0.4pt,rounded corners=1pt] (4.050,2.620) rectangle (5.760,3.380);
\node[text=consDark,font=\sffamily\scriptsize\bfseries] at (4.905,3.000) {0.0\%};
\path[fill=consBlue!8,draw=white,line width=0.4pt,rounded corners=1pt] (5.800,2.620) rectangle (7.510,3.380);
\node[text=consDark,font=\sffamily\scriptsize\bfseries] at (6.655,3.000) {0.5\%};
\path[fill=consBlue!8,draw=white,line width=0.4pt,rounded corners=1pt] (7.550,2.620) rectangle (9.260,3.380);
\node[text=consDark,font=\sffamily\scriptsize\bfseries] at (8.405,3.000) {0.5\%};
\path[fill=consBlue!89,draw=white,line width=0.4pt,rounded corners=1pt] (9.300,2.620) rectangle (11.010,3.380);
\node[text=white,font=\sffamily\scriptsize\bfseries] at (10.155,3.000) {98.9\%};
\node[anchor=east] at (3.75,2.00) {Qwen3-235B};
\path[fill=consBlue!8,draw=white,line width=0.4pt,rounded corners=1pt] (4.050,1.620) rectangle (5.760,2.380);
\node[text=consDark,font=\sffamily\scriptsize\bfseries] at (4.905,2.000) {0.0\%};
\path[fill=consBlue!8,draw=white,line width=0.4pt,rounded corners=1pt] (5.800,1.620) rectangle (7.510,2.380);
\node[text=consDark,font=\sffamily\scriptsize\bfseries] at (6.655,2.000) {0.0\%};
\path[fill=consBlue!8,draw=white,line width=0.4pt,rounded corners=1pt] (7.550,1.620) rectangle (9.260,2.380);
\node[text=consDark,font=\sffamily\scriptsize\bfseries] at (8.405,2.000) {0.0\%};
\path[fill=consBlue!90,draw=white,line width=0.4pt,rounded corners=1pt] (9.300,1.620) rectangle (11.010,2.380);
\node[text=white,font=\sffamily\scriptsize\bfseries] at (10.155,2.000) {100.0\%};
\node[anchor=east] at (3.75,1.00) {Qwen3-Next-80B};
\path[fill=consBlue!8,draw=white,line width=0.4pt,rounded corners=1pt] (4.050,0.620) rectangle (5.760,1.380);
\node[text=consDark,font=\sffamily\scriptsize\bfseries] at (4.905,1.000) {0.0\%};
\path[fill=consBlue!8,draw=white,line width=0.4pt,rounded corners=1pt] (5.800,0.620) rectangle (7.510,1.380);
\node[text=consDark,font=\sffamily\scriptsize\bfseries] at (6.655,1.000) {0.0\%};
\path[fill=consBlue!10,draw=white,line width=0.4pt,rounded corners=1pt] (7.550,0.620) rectangle (9.260,1.380);
\node[text=consDark,font=\sffamily\scriptsize\bfseries] at (8.405,1.000) {2.6\%};
\path[fill=consBlue!88,draw=white,line width=0.4pt,rounded corners=1pt] (9.300,0.620) rectangle (11.010,1.380);
\node[text=white,font=\sffamily\scriptsize\bfseries] at (10.155,1.000) {97.4\%};
\draw[black!35] (0.4,0.48) -- (11.2,0.48);
\node[anchor=east,font=\sffamily\scriptsize\bfseries] at (3.75,-0.25) {All eight models};
\path[fill=consBlue!8,draw=white,line width=0.4pt,rounded corners=1pt] (4.050,-0.630) rectangle (5.760,0.130);
\node[text=consDark,font=\sffamily\scriptsize\bfseries] at (4.905,-0.250) {0.5\%};
\path[fill=consBlue!11,draw=white,line width=0.4pt,rounded corners=1pt] (5.800,-0.630) rectangle (7.510,0.130);
\node[text=consDark,font=\sffamily\scriptsize\bfseries] at (6.655,-0.250) {3.5\%};
\path[fill=consBlue!19,draw=white,line width=0.4pt,rounded corners=1pt] (7.550,-0.630) rectangle (9.260,0.130);
\node[text=consDark,font=\sffamily\scriptsize\bfseries] at (8.405,-0.250) {13.1\%};
\path[fill=consBlue!76,draw=white,line width=0.4pt,rounded corners=1pt] (9.300,-0.630) rectangle (11.010,0.130);
\node[text=white,font=\sffamily\scriptsize\bfseries] at (10.155,-0.250) {82.9\%};
\node[anchor=west,text=consDark!75] at (4.05,-1.0) {Cell values are percentages of 190 panels per model.};
\end{tikzpicture}%
}\par\smallskip
\makebox[\linewidth][l]{\textbf{c}\hspace{0.6em}\sffamily\textbf{Primary and supporting specialty assignments}}\par\smallskip
\makebox[\linewidth][c]{%
\begin{tikzpicture}[x=0.52cm,y=0.59cm,font=\sffamily\scriptsize]
\draw[->,draw=consDark] (0,-0.15) -- (21,-0.15);
\draw[draw=consDark] (0,-0.15) -- (0,-0.34) node[below] {0};
\draw[draw=consDark] (5,-0.15) -- (5,-0.34) node[below] {5};
\draw[draw=consDark] (10,-0.15) -- (10,-0.34) node[below] {10};
\draw[draw=consDark] (15,-0.15) -- (15,-0.34) node[below] {15};
\draw[draw=consDark] (20,-0.15) -- (20,-0.34) node[below] {20};
\node[anchor=north,yshift=-18pt,font=\sffamily\scriptsize\bfseries] at (10.5,-0.15) {\% of assignments};
\node[anchor=east] at (-0.5,9.30) {Infectious disease};
\draw[black!10] (0,9.30) -- (20.4,9.30);
\draw[black!35,semithick] (10.4504,9.30) -- (13.0921,9.30);
\node[circle,fill=consDark,inner sep=0pt,minimum size=3.2pt] at (10.4504,9.30) {};
\node[rectangle,draw=consTeal,very thick,fill=white,inner sep=0pt,minimum size=4.2pt] at (13.0921,9.30) {};
\node[anchor=east] at (-0.5,8.30) {Radiology};
\draw[black!10] (0,8.30) -- (20.4,8.30);
\draw[black!35,semithick] (0.1974,8.30) -- (12.5717,8.30);
\node[circle,fill=consDark,inner sep=0pt,minimum size=3.2pt] at (12.5717,8.30) {};
\node[rectangle,draw=consTeal,very thick,fill=white,inner sep=0pt,minimum size=4.2pt] at (0.1974,8.30) {};
\node[anchor=west,text=consDark,font=\sffamily\scriptsize\itshape] at (13.1217,8.30) {729 total; 3 primary};
\node[anchor=east] at (-0.5,7.30) {Neurology};
\draw[black!10] (0,7.30) -- (20.4,7.30);
\draw[black!35,semithick] (5.8599,7.30) -- (16.1184,7.30);
\node[circle,fill=consDark,inner sep=0pt,minimum size=3.2pt] at (5.8599,7.30) {};
\node[rectangle,draw=consTeal,very thick,fill=white,inner sep=0pt,minimum size=4.2pt] at (16.1184,7.30) {};
\node[anchor=east] at (-0.5,6.30) {Cardiology};
\draw[black!10] (0,6.30) -- (20.4,6.30);
\draw[black!35,semithick] (4.9557,6.30) -- (7.1711,6.30);
\node[circle,fill=consDark,inner sep=0pt,minimum size=3.2pt] at (4.9557,6.30) {};
\node[rectangle,draw=consTeal,very thick,fill=white,inner sep=0pt,minimum size=4.2pt] at (7.1711,6.30) {};
\node[anchor=east] at (-0.5,5.30) {Pathology};
\draw[black!10] (0,5.30) -- (20.4,5.30);
\draw[black!35,semithick] (0.1974,5.30) -- (6.2598,5.30);
\node[circle,fill=consDark,inner sep=0pt,minimum size=3.2pt] at (6.2598,5.30) {};
\node[rectangle,draw=consTeal,very thick,fill=white,inner sep=0pt,minimum size=4.2pt] at (0.1974,5.30) {};
\node[anchor=east] at (-0.5,4.30) {Rheumatology};
\draw[black!10] (0,4.30) -- (20.4,4.30);
\draw[black!35,semithick] (4.3123,4.30) -- (7.4342,4.30);
\node[circle,fill=consDark,inner sep=0pt,minimum size=3.2pt] at (4.3123,4.30) {};
\node[rectangle,draw=consTeal,very thick,fill=white,inner sep=0pt,minimum size=4.2pt] at (7.4342,4.30) {};
\node[anchor=east] at (-0.5,3.30) {Pulmonology};
\draw[black!10] (0,3.30) -- (20.4,3.30);
\draw[black!35,semithick] (3.7733,3.30) -- (7.1053,3.30);
\node[circle,fill=consDark,inner sep=0pt,minimum size=3.2pt] at (3.7733,3.30) {};
\node[rectangle,draw=consTeal,very thick,fill=white,inner sep=0pt,minimum size=4.2pt] at (7.1053,3.30) {};
\node[anchor=east] at (-0.5,2.30) {Hematology};
\draw[black!10] (0,2.30) -- (20.4,2.30);
\draw[black!35,semithick] (3.9993,2.30) -- (4.8026,2.30);
\node[circle,fill=consDark,inner sep=0pt,minimum size=3.2pt] at (3.9993,2.30) {};
\node[rectangle,draw=consTeal,very thick,fill=white,inner sep=0pt,minimum size=4.2pt] at (4.8026,2.30) {};
\node[anchor=east] at (-0.5,1.30) {Gastroenterology};
\draw[black!10] (0,1.30) -- (20.4,1.30);
\draw[black!35,semithick] (3.1125,1.30) -- (5.1974,1.30);
\node[circle,fill=consDark,inner sep=0pt,minimum size=3.2pt] at (3.1125,1.30) {};
\node[rectangle,draw=consTeal,very thick,fill=white,inner sep=0pt,minimum size=4.2pt] at (5.1974,1.30) {};
\node[anchor=east] at (-0.5,0.30) {Oncology};
\draw[black!10] (0,0.30) -- (20.4,0.30);
\draw[black!35,semithick] (3.1647,0.30) -- (4.1447,0.30);
\node[circle,fill=consDark,inner sep=0pt,minimum size=3.2pt] at (3.1647,0.30) {};
\node[rectangle,draw=consTeal,very thick,fill=white,inner sep=0pt,minimum size=4.2pt] at (4.1447,0.30) {};
\node[circle,fill=consDark,inner sep=0pt,minimum size=3.2pt] at (4.4,10.1) {}; \node[anchor=west] at (4.8,10.1) {supporting roles ($n=5,751$)};
\node[rectangle,draw=consTeal,very thick,fill=white,inner sep=0pt,minimum size=4.2pt] at (12.15,10.1) {}; \node[anchor=west] at (12.70,10.1) {primary roles ($n=1,520$)};
\end{tikzpicture}%
}
\caption{Held-out cohort flow and model-generated panel architecture. (a) Only the 190 held-out cases entered the eight-model evaluation. Shared role assignments fed separate free-form and structured outputs; counts denote cases and artifacts, not enrolled participants. (b) Cell values are percentages within each model; the pooled row summarizes all 1,520 panels. (c) Categories are ranked by all 7,271 assignments and use separate primary and supporting denominators. The complete exact-label mapping is retained with the generated artifacts.}
\label{fig:panel-architecture}
\end{figure}

\FloatBarrier

\subsection{Consilium Dynamics}

Across models and response formats, the interactive review showed a clear pattern of early convergence. Report consensus-defined as unanimous role acceptance of the reviewed shared report before finalization-was achieved at the first review in \ConsInitialUnanimous{} of \ConsTotalArms{} case--model--format instances (\ConsInitialUnanimousPct\%). By the end of round~2, \ConsConsensusByRoundTwo{} instances (\ConsConsensusByRoundTwoPct\%) had reached consensus, and \ConsConsensus{} (\ConsConsensusPct\%) ultimately did so (Figures~\ref{fig:consilium-dynamics}a and~\ref{fig:consilium-stop-reasons}). Importantly, convergence was not confined to reports accepted at the outset: among the \ConsInitiallyDisputed{} initially disputed instances, \ConsInitiallyDisputedConsensus{} (\ConsInitiallyDisputedConsensusPct\%) subsequently reached unanimous acceptance. Overall, the review loop converted most contested initial reports into accepted shared reports, with agreement typically emerging within the first two rounds.

The process pattern varied substantially among models. GPT-5.6-Sol accepted most reports at the first review and required the fewest rounds, whereas Qwen3.5-27B began with dissent from every role but still reached eventual consensus in almost every case. These contrasting trajectories show that the benchmark captures model-specific review behavior in addition to final-answer quality.

Structured reports generally prompted somewhat more review. Structured and free-form pipelines averaged 1.89 and 1.76 rounds, respectively, and reached consensus in 97.3\% and 98.4\% of instances. The structured and free-form pipelines reached the five-round limit in 1.6\% and 1.1\% of cases. This aggregate pattern was model dependent (Table~\ref{tab:consilium-process}): DeepSeek-V4-Pro, for example, required fewer rounds for structured than for free-form reports (1.27 vs.\ 1.77).

Consensus did not mark the end of generation because a separate finalizer produced the delivered clinical answer after review. A complete path averaged 106.9 thousand provider-reported tokens per free-form case and 133.3 thousand per structured case; the structured pipeline therefore used 26.5 thousand additional tokens. The protocol consequently produced consistent procedural convergence, while structured reporting increased inference requirements. These process outcomes characterize role-conditioned synthetic review behavior and complement, rather than replace, direct assessment of clinical quality.
\definecolor{consBlue}{HTML}{0072B2}
\definecolor{consOrange}{HTML}{D55E00}
\definecolor{consDark}{HTML}{3F4A4F}
\begin{figure}[p]
\centering
\makebox[\linewidth][l]{\textbf{a}\hspace{0.6em}\sffamily\textbf{Unanimous report acceptance by review round}}\par\smallskip
\resizebox{0.63\linewidth}{!}{%
\begin{tikzpicture}[x=1.55cm,y=0.052cm,font=\sffamily\scriptsize]
\draw[->,draw=consDark] (0.75,0) -- (5.35,0);
\draw[->,draw=consDark] (0.75,0) -- (0.75,104);
\node[rotate=90,font=\sffamily\scriptsize\bfseries] at (-0.05,52) {accepted by or before round (\%)};
\draw[draw=consDark] (0.75,0) -- (0.65,0) node[left] {0};
\draw[draw=consDark] (0.75,25) -- (0.65,25) node[left] {25};
\draw[black!10] (0.75,25) -- (5.08,25);
\draw[draw=consDark] (0.75,50) -- (0.65,50) node[left] {50};
\draw[black!10] (0.75,50) -- (5.08,50);
\draw[draw=consDark] (0.75,75) -- (0.65,75) node[left] {75};
\draw[black!10] (0.75,75) -- (5.08,75);
\draw[draw=consDark] (0.75,100) -- (0.65,100) node[left] {100};
\draw[black!10] (0.75,100) -- (5.08,100);
\draw[draw=consDark] (1,0) -- (1,-2.2);
\node[below] at (1,-2.8) {1};
\draw[draw=consDark] (2,0) -- (2,-2.2);
\node[below] at (2,-2.8) {2};
\draw[draw=consDark] (3,0) -- (3,-2.2);
\node[below] at (3,-2.8) {3};
\draw[draw=consDark] (4,0) -- (4,-2.2);
\node[below] at (4,-2.8) {4};
\draw[draw=consDark] (5,0) -- (5,-2.2);
\node[below] at (5,-2.8) {5};
\node[anchor=north,yshift=-18pt,font=\sffamily\scriptsize\bfseries] at (3,0) {review round completed};
\draw[consBlue,thick] plot[mark=*,mark size=2.5pt] coordinates {(1,42.3684) (2,87.3026) (3,95.4605) (4,97.3026) (5,98.3553)};
\draw[consOrange,thick,dashed] plot[mark=square*,mark size=2.5pt] coordinates {(1,36.8421) (2,82.1711) (3,92.8947) (4,95.6579) (5,97.3026)};
\node[anchor=south,text=consBlue,font=\sffamily\scriptsize\bfseries] at (1,44.8684) {42.4\%};
\node[anchor=north,text=consOrange,font=\sffamily\scriptsize\bfseries] at (1,34.3421) {36.8\%};
\node[anchor=south,text=consBlue,font=\sffamily\scriptsize\bfseries] at (2,89.8026) {87.3\%};
\node[anchor=north,text=consOrange,font=\sffamily\scriptsize\bfseries] at (2,79.6711) {82.2\%};
\node[anchor=south west,text=consBlue,font=\sffamily\scriptsize\bfseries] at (5.08,98.3553) {Free-form 98.4\%};
\node[anchor=north west,text=consOrange,font=\sffamily\scriptsize\bfseries] at (5.08,97.3026) {Structured 97.3\%};
\draw[black!18] (0.75,-21) -- (5.08,-21);
\node[font=\sffamily\scriptsize\bfseries] at (3,-26) {arms entering each round, $n$};
\node[anchor=east,text=consBlue,font=\sffamily\scriptsize\bfseries] at (0.72,-34) {Free-form};
\node at (1,-34) {1,520};
\node at (2,-34) {874};
\node at (3,-34) {189};
\node at (4,-34) {63};
\node at (5,-34) {33};
\node[anchor=east,text=consOrange,font=\sffamily\scriptsize\bfseries] at (0.72,-42) {Structured};
\node at (1,-42) {1,520};
\node at (2,-42) {952};
\node at (3,-42) {260};
\node at (4,-42) {95};
\node at (5,-42) {51};
\end{tikzpicture}%
}\par\smallskip
\makebox[\linewidth][l]{\textbf{b}\hspace{0.6em}\sffamily\textbf{Dissent reduction after each report revision}}\par\vspace{1pt}
\makebox[\linewidth][c]{%
\begin{tikzpicture}[x=0.096cm,y=0.70cm,font=\sffamily\scriptsize]
\draw[->,draw=consDark] (0,0) -- (52,0);
\node[anchor=west,font=\sffamily\scriptsize\bfseries] at (58,4.75) {mean reduction (95\% CI)};
\node[anchor=west,font=\sffamily\scriptsize\bfseries] at (92,4.75) {arms compared};
\draw[draw=consDark] (0,0) -- (0,-0.17) node[below] {0};
\draw[draw=consDark] (10,0) -- (10,-0.17) node[below] {10};
\draw[draw=consDark] (20,0) -- (20,-0.17) node[below] {20};
\draw[draw=consDark] (30,0) -- (30,-0.17) node[below] {30};
\draw[draw=consDark] (40,0) -- (40,-0.17) node[below] {40};
\draw[draw=consDark] (50,0) -- (50,-0.17) node[below] {50};
\node[anchor=east,font=\sffamily\scriptsize\bfseries] at (-4,4) {Round 1 to 2};
\draw[consBlue,very thick] (47.7009,4) -- (50.6907,4);
\draw[consBlue,thick] (47.7009,3.87) -- (47.7009,4.13); \draw[consBlue,thick] (50.6907,3.87) -- (50.6907,4.13);
\node[circle,fill=consBlue,inner sep=0pt,minimum size=3.6pt] at (49.2141,4) {};
\node[anchor=west] at (58,4) {$ 49.2\ (47.7,\ 50.7) $};
\node[anchor=west] at (92,4) {1,826};
\node[anchor=east,font=\sffamily\scriptsize\bfseries] at (-4,3) {Round 2 to 3};
\draw[consBlue,very thick] (29.6010,3) -- (37.7742,3);
\draw[consBlue,thick] (29.6010,2.87) -- (29.6010,3.13); \draw[consBlue,thick] (37.7742,2.87) -- (37.7742,3.13);
\node[circle,fill=consBlue,inner sep=0pt,minimum size=3.6pt] at (33.7082,3) {};
\node[anchor=west] at (58,3) {$ 33.7\ (29.6,\ 37.8) $};
\node[anchor=west] at (92,3) {449};
\node[anchor=east,font=\sffamily\scriptsize\bfseries] at (-4,2) {Round 3 to 4};
\draw[consBlue,very thick] (14.3030,2) -- (28.4570,2);
\draw[consBlue,thick] (14.3030,1.87) -- (14.3030,2.13); \draw[consBlue,thick] (28.4570,1.87) -- (28.4570,2.13);
\node[circle,fill=consBlue,inner sep=0pt,minimum size=3.6pt] at (21.3924,2) {};
\node[anchor=west] at (58,2) {$ 21.4\ (14.3,\ 28.5) $};
\node[anchor=west] at (92,2) {158};
\node[anchor=east,font=\sffamily\scriptsize\bfseries] at (-4,1) {Round 4 to 5};
\draw[consBlue,very thick] (20.1922,1) -- (38.0321,1);
\draw[consBlue,thick] (20.1922,0.87) -- (20.1922,1.13); \draw[consBlue,thick] (38.0321,0.87) -- (38.0321,1.13);
\node[circle,fill=consBlue,inner sep=0pt,minimum size=3.6pt] at (29.2857,1) {};
\node[anchor=west] at (58,1) {$ 29.3\ (20.2,\ 38.0) $};
\node[anchor=west] at (92,1) {84};
\draw[dashed,black!45] (0,0.28) -- (0,4.45);
\node[anchor=north,yshift=-18pt,font=\sffamily\scriptsize\bfseries,text=consDark!85] at (26,0) {mean drop in dissent (percentage points)};
\end{tikzpicture}%
}\par\vspace{1pt}
\makebox[\linewidth][l]{\textbf{c}\hspace{0.6em}\sffamily\textbf{Model-specific process patterns}}\par\smallskip
\makebox[\linewidth][c]{%
\begin{tikzpicture}[x=0.90cm,y=0.39cm,font=\sffamily\scriptsize]
\node[circle,fill=consBlue,inner sep=0pt,minimum size=3.4pt] at (5.5,11.15) {}; \node[anchor=west] at (5.72,11.15) {Free-form};
\node[rectangle,fill=consOrange,inner sep=0pt,minimum size=3.4pt] at (7.65,11.15) {}; \node[anchor=west] at (8.0,11.15) {Structured};
\node[font=\sffamily\scriptsize\bfseries] at (6.1,10.25) {Round-1 dissent (\%)};
\node[font=\sffamily\scriptsize\bfseries] at (11.0,10.25) {Mean rounds};
\node[font=\sffamily\scriptsize\bfseries] at (15.9,10.25) {Round-limit stops (\%)};
\draw[draw=consDark] (4.5,-0.25) -- (7.7,-0.25);
\draw[black!12] (4.5,0.0) -- (4.5,9.55);
\draw[draw=consDark] (4.5,-0.25) -- (4.5,-0.43) node[below] {0};
\draw[black!12] (6.1,0.0) -- (6.1,9.55);
\draw[draw=consDark] (6.1,-0.25) -- (6.1,-0.43) node[below] {50};
\draw[black!12] (7.7,0.0) -- (7.7,9.55);
\draw[draw=consDark] (7.7,-0.25) -- (7.7,-0.43) node[below] {100};
\draw[draw=consDark] (9.4,-0.25) -- (12.6,-0.25);
\draw[black!12] (9.4,0.0) -- (9.4,9.55);
\draw[draw=consDark] (9.4,-0.25) -- (9.4,-0.43) node[below] {1};
\draw[black!12] (11.0,0.0) -- (11.0,9.55);
\draw[draw=consDark] (11.0,-0.25) -- (11.0,-0.43) node[below] {3};
\draw[black!12] (12.6,0.0) -- (12.6,9.55);
\draw[draw=consDark] (12.6,-0.25) -- (12.6,-0.43) node[below] {5};
\draw[draw=consDark] (14.3,-0.25) -- (17.5,-0.25);
\draw[black!12] (14.3,0.0) -- (14.3,9.55);
\draw[draw=consDark] (14.3,-0.25) -- (14.3,-0.43) node[below] {0};
\draw[black!12] (15.366666666666667,0.0) -- (15.366666666666667,9.55);
\draw[draw=consDark] (15.366666666666667,-0.25) -- (15.366666666666667,-0.43) node[below] {5};
\draw[black!12] (16.433333333333334,0.0) -- (16.433333333333334,9.55);
\draw[draw=consDark] (16.433333333333334,-0.25) -- (16.433333333333334,-0.43) node[below] {10};
\draw[black!12] (17.5,0.0) -- (17.5,9.55);
\draw[draw=consDark] (17.5,-0.25) -- (17.5,-0.43) node[below] {15};
\draw[black!7] (4.35,9) -- (17.55,9);
\node[anchor=east] at (4.05,9) {DeepSeek-V4-Pro};
\draw[black!35,thin] (5.4513,9.130) -- (4.7414,8.870);
\node[circle,fill=consBlue,inner sep=0pt,minimum size=3.2pt] at (5.4513,9.130) {};
\node[rectangle,fill=consOrange,inner sep=0pt,minimum size=3.2pt] at (4.7414,8.870) {};
\draw[black!35,thin] (10.0189,9.130) -- (9.6147,8.870);
\node[circle,fill=consBlue,inner sep=0pt,minimum size=3.2pt] at (10.0189,9.130) {};
\node[rectangle,fill=consOrange,inner sep=0pt,minimum size=3.2pt] at (9.6147,8.870) {};
\draw[black!35,thin] (14.3000,9.130) -- (14.3000,8.870);
\node[circle,fill=consBlue,inner sep=0pt,minimum size=3.2pt] at (14.3000,9.130) {};
\node[rectangle,fill=consOrange,inner sep=0pt,minimum size=3.2pt] at (14.3000,8.870) {};
\draw[black!7] (4.35,8) -- (17.55,8);
\node[anchor=east] at (4.05,8) {Gemma-4-31B};
\draw[black!35,thin] (5.1529,8.130) -- (5.4898,7.870);
\node[circle,fill=consBlue,inner sep=0pt,minimum size=3.2pt] at (5.1529,8.130) {};
\node[rectangle,fill=consOrange,inner sep=0pt,minimum size=3.2pt] at (5.4898,7.870) {};
\draw[black!35,thin] (9.7495,8.130) -- (9.8505,7.870);
\node[circle,fill=consBlue,inner sep=0pt,minimum size=3.2pt] at (9.7495,8.130) {};
\node[rectangle,fill=consOrange,inner sep=0pt,minimum size=3.2pt] at (9.8505,7.870) {};
\draw[black!35,thin] (14.3000,8.130) -- (14.3000,7.870);
\node[circle,fill=consBlue,inner sep=0pt,minimum size=3.2pt] at (14.3000,8.130) {};
\node[rectangle,fill=consOrange,inner sep=0pt,minimum size=3.2pt] at (14.3000,7.870) {};
\draw[black!7] (4.35,7) -- (17.55,7);
\node[anchor=east] at (4.05,7) {GLM-5.2};
\draw[black!35,thin] (5.8269,7.130) -- (5.7424,6.870);
\node[circle,fill=consBlue,inner sep=0pt,minimum size=3.2pt] at (5.8269,7.130) {};
\node[rectangle,fill=consOrange,inner sep=0pt,minimum size=3.2pt] at (5.7424,6.870) {};
\draw[black!35,thin] (10.1621,7.130) -- (10.0653,6.870);
\node[circle,fill=consBlue,inner sep=0pt,minimum size=3.2pt] at (10.1621,7.130) {};
\node[rectangle,fill=consOrange,inner sep=0pt,minimum size=3.2pt] at (10.0653,6.870) {};
\draw[black!35,thin] (14.3000,7.130) -- (14.3000,6.870);
\node[circle,fill=consBlue,inner sep=0pt,minimum size=3.2pt] at (14.3000,7.130) {};
\node[rectangle,fill=consOrange,inner sep=0pt,minimum size=3.2pt] at (14.3000,6.870) {};
\draw[black!7] (4.35,6) -- (17.55,6);
\node[anchor=east] at (4.05,6) {GPT-5.6-Sol};
\draw[black!35,thin] (4.7779,6.130) -- (4.7695,5.870);
\node[circle,fill=consBlue,inner sep=0pt,minimum size=3.2pt] at (4.7779,6.130) {};
\node[rectangle,fill=consOrange,inner sep=0pt,minimum size=3.2pt] at (4.7695,5.870) {};
\draw[black!35,thin] (9.5726,6.130) -- (9.5853,5.870);
\node[circle,fill=consBlue,inner sep=0pt,minimum size=3.2pt] at (9.5726,6.130) {};
\node[rectangle,fill=consOrange,inner sep=0pt,minimum size=3.2pt] at (9.5853,5.870) {};
\draw[black!35,thin] (14.3000,6.130) -- (14.3000,5.870);
\node[circle,fill=consBlue,inner sep=0pt,minimum size=3.2pt] at (14.3000,6.130) {};
\node[rectangle,fill=consOrange,inner sep=0pt,minimum size=3.2pt] at (14.3000,5.870) {};
\draw[black!7] (4.35,5) -- (17.55,5);
\node[anchor=east] at (4.05,5) {GPT-OSS-120B};
\draw[black!35,thin] (5.8684,5.130) -- (6.8680,4.870);
\node[circle,fill=consBlue,inner sep=0pt,minimum size=3.2pt] at (5.8684,5.130) {};
\node[rectangle,fill=consOrange,inner sep=0pt,minimum size=3.2pt] at (6.8680,4.870) {};
\draw[black!35,thin] (10.5874,5.130) -- (11.0968,4.870);
\node[circle,fill=consBlue,inner sep=0pt,minimum size=3.2pt] at (10.5874,5.130) {};
\node[rectangle,fill=consOrange,inner sep=0pt,minimum size=3.2pt] at (11.0968,4.870) {};
\draw[black!35,thin] (15.6474,5.130) -- (16.5456,4.870);
\node[circle,fill=consBlue,inner sep=0pt,minimum size=3.2pt] at (15.6474,5.130) {};
\node[rectangle,fill=consOrange,inner sep=0pt,minimum size=3.2pt] at (16.5456,4.870) {};
\draw[black!7] (4.35,4) -- (17.55,4);
\node[anchor=east] at (4.05,4) {Qwen3.5-27B};
\draw[black!35,thin] (7.7000,4.130) -- (7.7000,3.870);
\node[circle,fill=consBlue,inner sep=0pt,minimum size=3.2pt] at (7.7000,4.130) {};
\node[rectangle,fill=consOrange,inner sep=0pt,minimum size=3.2pt] at (7.7000,3.870) {};
\draw[black!35,thin] (10.5705,4.130) -- (10.6000,3.870);
\node[circle,fill=consBlue,inner sep=0pt,minimum size=3.2pt] at (10.5705,4.130) {};
\node[rectangle,fill=consOrange,inner sep=0pt,minimum size=3.2pt] at (10.6000,3.870) {};
\draw[black!35,thin] (14.6368,4.130) -- (14.5246,3.870);
\node[circle,fill=consBlue,inner sep=0pt,minimum size=3.2pt] at (14.6368,4.130) {};
\node[rectangle,fill=consOrange,inner sep=0pt,minimum size=3.2pt] at (14.5246,3.870) {};
\draw[black!7] (4.35,3) -- (17.55,3);
\node[anchor=east] at (4.05,3) {Qwen3-235B};
\draw[black!35,thin] (4.8234,3.130) -- (5.2006,2.870);
\node[circle,fill=consBlue,inner sep=0pt,minimum size=3.2pt] at (4.8234,3.130) {};
\node[rectangle,fill=consOrange,inner sep=0pt,minimum size=3.2pt] at (5.2006,2.870) {};
\draw[black!35,thin] (9.6063,3.130) -- (9.8126,2.870);
\node[circle,fill=consBlue,inner sep=0pt,minimum size=3.2pt] at (9.6063,3.130) {};
\node[rectangle,fill=consOrange,inner sep=0pt,minimum size=3.2pt] at (9.8126,2.870) {};
\draw[black!35,thin] (14.3000,3.130) -- (14.4123,2.870);
\node[circle,fill=consBlue,inner sep=0pt,minimum size=3.2pt] at (14.3000,3.130) {};
\node[rectangle,fill=consOrange,inner sep=0pt,minimum size=3.2pt] at (14.4123,2.870) {};
\draw[black!7] (4.35,2) -- (17.55,2);
\node[anchor=east] at (4.05,2) {Qwen3-Next-80B};
\draw[black!35,thin] (4.9598,2.130) -- (6.1253,1.870);
\node[circle,fill=consBlue,inner sep=0pt,minimum size=3.2pt] at (4.9598,2.130) {};
\node[rectangle,fill=consOrange,inner sep=0pt,minimum size=3.2pt] at (6.1253,1.870) {};
\draw[black!35,thin] (9.8126,2.130) -- (10.2926,1.870);
\node[circle,fill=consBlue,inner sep=0pt,minimum size=3.2pt] at (9.8126,2.130) {};
\node[rectangle,fill=consOrange,inner sep=0pt,minimum size=3.2pt] at (10.2926,1.870) {};
\draw[black!35,thin] (14.5246,2.130) -- (14.4123,1.870);
\node[circle,fill=consBlue,inner sep=0pt,minimum size=3.2pt] at (14.5246,2.130) {};
\node[rectangle,fill=consOrange,inner sep=0pt,minimum size=3.2pt] at (14.4123,1.870) {};
\draw[black!7] (4.35,1) -- (17.55,1);
\node[anchor=east,font=\sffamily\scriptsize\bfseries] at (4.05,1) {All eight models};
\draw[black!35,semithick] (5.5701,1.130) -- (5.8296,0.870);
\node[circle,fill=consBlue,inner sep=0pt,minimum size=3.2pt] at (5.5701,1.130) {};
\node[rectangle,fill=consOrange,inner sep=0pt,minimum size=3.2pt] at (5.8296,0.870) {};
\draw[black!35,semithick] (10.0100,1.130) -- (10.1147,0.870);
\node[circle,fill=consBlue,inner sep=0pt,minimum size=3.2pt] at (10.0100,1.130) {};
\node[rectangle,fill=consOrange,inner sep=0pt,minimum size=3.2pt] at (10.1147,0.870) {};
\draw[black!35,semithick] (14.5386,1.130) -- (14.6368,0.870);
\node[circle,fill=consBlue,inner sep=0pt,minimum size=3.2pt] at (14.5386,1.130) {};
\node[rectangle,fill=consOrange,inner sep=0pt,minimum size=3.2pt] at (14.6368,0.870) {};
\draw[black!45] (0.4,1.550) -- (17.55,1.550);
\end{tikzpicture}%
}
\caption{Interactive-consilium dynamics in 3,040 case--model--pipeline arms. (a) Cumulative percentage of 1,520 arms per response format with unanimous acceptance of the reviewed report; round~1 evaluates the initial report. Counts show arms entering each round. (b) Mean reduction in dissent among the same arms before and after each revision, with 95\% CIs from 10,000 DOI-clustered bootstrap resamples; the right column gives the number of arms compared. (c) Free-form (blue circles) and structured (orange squares) outcomes by model; lines connect response formats within models, and the final row pools the fixed eight-model set.}
\label{fig:consilium-dynamics}
\end{figure}

\begin{table}[tbp]
\caption{Model-specific controller-recorded interactive-consilium outcomes in the held-out evaluation cohort.}\label{tab:consilium-process}
\centering
\scriptsize
\setlength{\tabcolsep}{2.6pt}
\begin{tabular}{@{}p{0.238\linewidth}lrrrrrr@{}}
\toprule
\textbf{Model} & \textbf{Pipeline} & \shortstack{\textbf{R1 dissent,}\\\textbf{mean \%}} & \shortstack{\textbf{R1 unanimous,}\\\textbf{$n$ (\%)}} & \shortstack{\textbf{Mean}\\\textbf{rounds}} & \shortstack{\textbf{Report consensus,}\\\textbf{$n$ (\%)}} & \shortstack{\textbf{No-change stop,}\\\textbf{$n$ (\%)}} & \shortstack{\textbf{Round-limit}\\\textbf{stop, $n$ (\%)}} \\
\midrule
DeepSeek-V4-Pro & Free-form & 29.7 & 58 (30.5\%) & 1.77 & 189 (99.5\%) & 1 (0.5\%) & 0 (0.0\%) \\
 & Structured & 7.5 & 144 (75.8\%) & 1.27 & 189 (99.5\%) & 1 (0.5\%) & 0 (0.0\%) \\
\addlinespace[1.5pt]
Gemma-4-31B & Free-form & 20.4 & 110 (57.9\%) & 1.44 & 190 (100.0\%) & 0 (0.0\%) & 0 (0.0\%) \\
 & Structured & 30.9 & 87 (45.8\%) & 1.56 & 190 (100.0\%) & 0 (0.0\%) & 0 (0.0\%) \\
\addlinespace[1.5pt]
GLM-5.2 & Free-form & 41.5 & 33 (17.4\%) & 1.95 & 189 (99.5\%) & 1 (0.5\%) & 0 (0.0\%) \\
 & Structured & 38.8 & 40 (21.1\%) & 1.83 & 186 (97.9\%) & 4 (2.1\%) & 0 (0.0\%) \\
\addlinespace[1.5pt]
GPT-5.6-Sol & Free-form & 8.7 & 155 (81.6\%) & 1.22 & 190 (100.0\%) & 0 (0.0\%) & 0 (0.0\%) \\
 & Structured & 8.4 & 151 (79.5\%) & 1.23 & 188 (98.9\%) & 2 (1.1\%) & 0 (0.0\%) \\
\addlinespace[1.5pt]
GPT-OSS-120B & Free-form & 42.8 & 33 (17.4\%) & 2.48 & 173 (91.1\%) & 5 (2.6\%) & 12 (6.3\%) \\
 & Structured & 74.0 & 3 (1.6\%) & 3.12 & 163 (85.8\%) & 7 (3.7\%) & 20 (10.5\%) \\
\addlinespace[1.5pt]
Qwen3.5-27B & Free-form & 100.0 & 0 (0.0\%) & 2.46 & 186 (97.9\%) & 1 (0.5\%) & 3 (1.6\%) \\
 & Structured & 100.0 & 0 (0.0\%) & 2.50 & 188 (98.9\%) & 0 (0.0\%) & 2 (1.1\%) \\
\addlinespace[1.5pt]
Qwen3-235B & Free-form & 10.1 & 143 (75.3\%) & 1.26 & 190 (100.0\%) & 0 (0.0\%) & 0 (0.0\%) \\
 & Structured & 21.9 & 103 (54.2\%) & 1.52 & 189 (99.5\%) & 0 (0.0\%) & 1 (0.5\%) \\
\addlinespace[1.5pt]
Qwen3-Next-80B & Free-form & 14.4 & 112 (58.9\%) & 1.52 & 188 (98.9\%) & 0 (0.0\%) & 2 (1.1\%) \\
 & Structured & 50.8 & 32 (16.8\%) & 2.12 & 186 (97.9\%) & 3 (1.6\%) & 1 (0.5\%) \\
\midrule
\textbf{All eight models} & Free-form & 33.4 & 644 (42.4\%) & 1.76 & 1,495 (98.4\%) & 8 (0.5\%) & 17 (1.1\%) \\
 & Structured & 41.6 & 560 (36.8\%) & 1.89 & 1,479 (97.3\%) & 17 (1.1\%) & 24 (1.6\%) \\
\bottomrule
\end{tabular}
\begin{flushleft}
\footnotesize R1 denotes the first review round. Dissent is the mean arm-level percentage of assigned roles voting \texttt{no}; every case--model arm therefore receives equal weight irrespective of panel size. Report consensus denotes unanimous role acceptance of the currently reviewed shared report before finalization. At a round-limit stop, the moderator's revision after the fifth review was retained without an additional role vote. Model rows contain 190 cases per pipeline and pooled rows contain 1,520 case--model arms per pipeline. No-change denotes the controller stop reason \texttt{no\_substantive\_change}. Every model completed all 190 cases in both pipelines.
\end{flushleft}
\end{table}

\begin{table}[tbp]
\caption{Role-level vote transitions among interactive consilia continuing to the next review round.}\label{tab:consilium-process-audit}
\centering
\scriptsize
\setlength{\tabcolsep}{7.0pt}
\begin{tabular}{@{}lrrrr@{}}
\toprule
\textbf{Transition} & \textbf{R1$\rightarrow$R2} & \textbf{R2$\rightarrow$R3} & \textbf{R3$\rightarrow$R4} & \textbf{R4$\rightarrow$R5} \\
\midrule
\texttt{no}$\rightarrow$\texttt{yes} & 4,582 (83.1\%) & 937 (79.9\%) & 278 (66.0\%) & 153 (62.4\%) \\
\texttt{no}$\rightarrow$\texttt{no} & 932 (16.9\%) & 235 (20.1\%) & 143 (34.0\%) & 92 (37.6\%) \\
\addlinespace[1.5pt]
\texttt{yes}$\rightarrow$\texttt{no} & 246 (7.4\%) & 194 (18.5\%) & 109 (29.7\%) & 31 (17.9\%) \\
\texttt{yes}$\rightarrow$\texttt{yes} & 3,064 (92.6\%) & 854 (81.5\%) & 258 (70.3\%) & 142 (82.1\%) \\
\bottomrule
\end{tabular}
\begin{flushleft}
\footnotesize Percentages are conditional on the preceding vote, so the two rows beginning with the same vote sum to 100\% within a transition. Each continuing role contributes one observation; larger panels therefore contribute more transitions. Figure~\ref{fig:consilium-dynamics}b instead gives equal weight to each case--model--format instance.
\end{flushleft}
\end{table}

\definecolor{stopBlue}{HTML}{4C78A8}
\definecolor{stopOrange}{HTML}{E45756}
\definecolor{stopGray}{HTML}{8F969C}
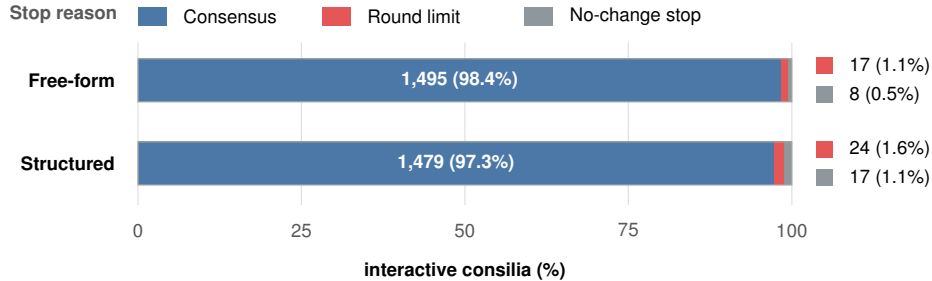
\begin{figure}[tbp]
\centering
\begin{tikzpicture}[x=1cm,y=1cm,font=\sffamily\scriptsize]
\node[anchor=east,font=\sffamily\scriptsize\bfseries,text=black!65] at (2.00,2.82) {Stop reason};
\fill[stopBlue] (2.15,2.68) rectangle (2.53,2.92);
\node[anchor=west] at (2.63,2.80) {Consensus};
\fill[stopOrange] (4.58,2.68) rectangle (4.96,2.92);
\node[anchor=west] at (5.06,2.80) {Round limit};
\fill[stopGray] (7.26,2.68) rectangle (7.64,2.92);
\node[anchor=west] at (7.74,2.80) {No-change stop};
\draw[black!12,line width=0.45pt] (2.150,0.30) -- (2.150,2.45);
\node[anchor=north,text=black!65] at (2.150,0.18) {0};
\draw[black!12,line width=0.45pt] (4.312,0.30) -- (4.312,2.45);
\node[anchor=north,text=black!65] at (4.312,0.18) {25};
\draw[black!12,line width=0.45pt] (6.475,0.30) -- (6.475,2.45);
\node[anchor=north,text=black!65] at (6.475,0.18) {50};
\draw[black!12,line width=0.45pt] (8.637,0.30) -- (8.637,2.45);
\node[anchor=north,text=black!65] at (8.637,0.18) {75};
\draw[black!12,line width=0.45pt] (10.800,0.30) -- (10.800,2.45);
\node[anchor=north,text=black!65] at (10.800,0.18) {100};
\node[anchor=east,font=\sffamily\scriptsize\bfseries] at (1.970,1.950) {Free-form};
\fill[stopBlue] (2.150,1.665) rectangle (10.658,2.235);
\fill[stopOrange] (10.658,1.665) rectangle (10.754,2.235);
\fill[stopGray] (10.754,1.665) rectangle (10.800,2.235);
\draw[black!35,line width=0.45pt] (2.150,1.665) rectangle (10.800,2.235);
\node[font=\sffamily\scriptsize\bfseries,text=white] at (6.404,1.950) {1,495 (98.4\%)};
\fill[stopOrange] (11.12,2.030) rectangle (11.34,2.250);
\node[anchor=west] at (11.44,2.140) {17 (1.1\%)};
\fill[stopGray] (11.12,1.650) rectangle (11.34,1.870);
\node[anchor=west] at (11.44,1.760) {8 (0.5\%)};
\node[anchor=east,font=\sffamily\scriptsize\bfseries] at (1.970,0.850) {Structured};
\fill[stopBlue] (2.150,0.565) rectangle (10.567,1.135);
\fill[stopOrange] (10.567,0.565) rectangle (10.703,1.135);
\fill[stopGray] (10.703,0.565) rectangle (10.800,1.135);
\draw[black!35,line width=0.45pt] (2.150,0.565) rectangle (10.800,1.135);
\node[font=\sffamily\scriptsize\bfseries,text=white] at (6.358,0.850) {1,479 (97.3\%)};
\fill[stopOrange] (11.12,0.930) rectangle (11.34,1.150);
\node[anchor=west] at (11.44,1.040) {24 (1.6\%)};
\fill[stopGray] (11.12,0.550) rectangle (11.34,0.770);
\node[anchor=west] at (11.44,0.660) {17 (1.1\%)};
\node[anchor=north,yshift=-18pt,font=\sffamily\scriptsize\bfseries] at (6.475,0.30) {interactive consilia (\%)};
\end{tikzpicture}
\caption{Controller-recorded stop reasons by response format. Each continuous 100\% stacked bar partitions 1,520 interactive consilia into mutually exclusive stop reasons; labels give counts and percentages. Values for the narrow round-limit and no-change segments are repeated at right with matching color keys. Consensus indicates unanimous acceptance of the reviewed shared report. A no-substantive-change stop indicates that the moderator identified no further clinically meaningful revision; a round-limit stop indicates persistent dissent through review round~5.}
\label{fig:consilium-stop-reasons}
\end{figure}

\definecolor{usageInput}{HTML}{4C78A8}
\definecolor{usageOutput}{HTML}{E69F00}
\begin{figure}[p]
\centering
\makebox[\linewidth][l]{\textbf{a}\hspace{0.6em}\sffamily\textbf{Complete-pipeline token use by model and response format}}\par\smallskip
\makebox[\linewidth][c]{%
\begin{tikzpicture}[x=1.40cm,y=0.70cm,font=\sffamily\scriptsize]
\node[font=\sffamily\scriptsize\bfseries] at (3.82,9.15) {Format};
\fill[usageInput] (5.25,9.00) rectangle (5.62,9.26);
\node[anchor=west] at (5.72,9.13) {Prompt/input tokens};
\fill[usageOutput] (8.55,9.00) rectangle (8.92,9.26);
\node[anchor=west] at (9.02,9.13) {Completion/output tokens};
\draw[black!12,line width=0.45pt] (4.120,0.48) -- (4.120,8.72);
\node[anchor=north,text=black!65] at (4.120,0.32) {0};
\draw[black!12,line width=0.45pt] (5.940,0.48) -- (5.940,8.72);
\node[anchor=north,text=black!65] at (5.940,0.32) {50};
\draw[black!12,line width=0.45pt] (7.760,0.48) -- (7.760,8.72);
\node[anchor=north,text=black!65] at (7.760,0.32) {100};
\draw[black!12,line width=0.45pt] (9.580,0.48) -- (9.580,8.72);
\node[anchor=north,text=black!65] at (9.580,0.32) {150};
\draw[black!12,line width=0.45pt] (11.400,0.48) -- (11.400,8.72);
\node[anchor=north,text=black!65] at (11.400,0.32) {200};
\draw[black!12,line width=0.45pt] (13.220,0.48) -- (13.220,8.72);
\node[anchor=north,text=black!65] at (13.220,0.32) {250};
\node[anchor=east,font=\sffamily\scriptsize\bfseries] at (3.48,8.000) {DeepSeek-V4-Pro};
\node[font=\sffamily\scriptsize\bfseries] at (3.82,8.190) {F};
\fill[usageInput] (4.120,8.055) rectangle (7.194,8.325);
\fill[usageOutput] (7.194,8.055) rectangle (7.703,8.325);
\node[anchor=west,font=\sffamily\scriptsize\bfseries] at (7.803,8.190) {98};
\node[font=\sffamily\scriptsize\bfseries] at (3.82,7.810) {S};
\fill[usageInput] (4.120,7.675) rectangle (7.285,7.945);
\fill[usageOutput] (7.285,7.675) rectangle (7.902,7.945);
\node[anchor=west,font=\sffamily\scriptsize\bfseries] at (8.002,7.810) {104};
\node[anchor=east,font=\sffamily\scriptsize\bfseries] at (3.48,7.000) {Gemma-4-31B};
\node[font=\sffamily\scriptsize\bfseries] at (3.82,7.190) {F};
\fill[usageInput] (4.120,7.055) rectangle (5.690,7.325);
\fill[usageOutput] (5.690,7.055) rectangle (5.921,7.325);
\node[anchor=west,font=\sffamily\scriptsize\bfseries] at (6.021,7.190) {49};
\node[font=\sffamily\scriptsize\bfseries] at (3.82,6.810) {S};
\fill[usageInput] (4.120,6.675) rectangle (6.324,6.945);
\fill[usageOutput] (6.324,6.675) rectangle (6.655,6.945);
\node[anchor=west,font=\sffamily\scriptsize\bfseries] at (6.755,6.810) {70};
\node[anchor=east,font=\sffamily\scriptsize\bfseries] at (3.48,6.000) {GLM-5.2};
\node[font=\sffamily\scriptsize\bfseries] at (3.82,6.190) {F};
\fill[usageInput] (4.120,6.055) rectangle (7.461,6.325);
\fill[usageOutput] (7.461,6.055) rectangle (8.016,6.325);
\node[anchor=west,font=\sffamily\scriptsize\bfseries] at (8.116,6.190) {107};
\node[font=\sffamily\scriptsize\bfseries] at (3.82,5.810) {S};
\fill[usageInput] (4.120,5.675) rectangle (7.985,5.945);
\fill[usageOutput] (7.985,5.675) rectangle (8.642,5.945);
\node[anchor=west,font=\sffamily\scriptsize\bfseries] at (8.742,5.810) {124};
\node[anchor=east,font=\sffamily\scriptsize\bfseries] at (3.48,5.000) {GPT-5.6-Sol};
\node[font=\sffamily\scriptsize\bfseries] at (3.82,5.190) {F};
\fill[usageInput] (4.120,5.055) rectangle (6.133,5.325);
\fill[usageOutput] (6.133,5.055) rectangle (6.492,5.325);
\node[anchor=west,font=\sffamily\scriptsize\bfseries] at (6.592,5.190) {65};
\node[font=\sffamily\scriptsize\bfseries] at (3.82,4.810) {S};
\fill[usageInput] (4.120,4.675) rectangle (8.071,4.945);
\fill[usageOutput] (8.071,4.675) rectangle (8.927,4.945);
\node[anchor=west,font=\sffamily\scriptsize\bfseries] at (9.027,4.810) {132};
\node[anchor=east,font=\sffamily\scriptsize\bfseries] at (3.48,4.000) {GPT-OSS-120B};
\node[font=\sffamily\scriptsize\bfseries] at (3.82,4.190) {F};
\fill[usageInput] (4.120,4.055) rectangle (10.123,4.325);
\fill[usageOutput] (10.123,4.055) rectangle (11.661,4.325);
\node[anchor=west,font=\sffamily\scriptsize\bfseries] at (11.761,4.190) {207};
\node[font=\sffamily\scriptsize\bfseries] at (3.82,3.810) {S};
\fill[usageInput] (4.120,3.675) rectangle (10.084,3.945);
\fill[usageOutput] (10.084,3.675) rectangle (11.739,3.945);
\node[anchor=west,font=\sffamily\scriptsize\bfseries] at (11.839,3.810) {209};
\node[anchor=east,font=\sffamily\scriptsize\bfseries] at (3.48,3.000) {Qwen3.5-27B};
\node[font=\sffamily\scriptsize\bfseries] at (3.82,3.190) {F};
\fill[usageInput] (4.120,3.055) rectangle (8.644,3.325);
\fill[usageOutput] (8.644,3.055) rectangle (9.293,3.325);
\node[anchor=west,font=\sffamily\scriptsize\bfseries] at (9.393,3.190) {142};
\node[font=\sffamily\scriptsize\bfseries] at (3.82,2.810) {S};
\fill[usageInput] (4.120,2.675) rectangle (9.022,2.945);
\fill[usageOutput] (9.022,2.675) rectangle (9.687,2.945);
\node[anchor=west,font=\sffamily\scriptsize\bfseries] at (9.787,2.810) {153};
\node[anchor=east,font=\sffamily\scriptsize\bfseries] at (3.48,2.000) {Qwen3-235B};
\node[font=\sffamily\scriptsize\bfseries] at (3.82,2.190) {F};
\fill[usageInput] (4.120,2.055) rectangle (6.921,2.325);
\fill[usageOutput] (6.921,2.055) rectangle (7.480,2.325);
\node[anchor=west,font=\sffamily\scriptsize\bfseries] at (7.580,2.190) {92};
\node[font=\sffamily\scriptsize\bfseries] at (3.82,1.810) {S};
\fill[usageInput] (4.120,1.675) rectangle (8.259,1.945);
\fill[usageOutput] (8.259,1.675) rectangle (9.049,1.945);
\node[anchor=west,font=\sffamily\scriptsize\bfseries] at (9.149,1.810) {135};
\node[anchor=east,font=\sffamily\scriptsize\bfseries] at (3.48,1.000) {Qwen3-Next-80B};
\node[font=\sffamily\scriptsize\bfseries] at (3.82,1.190) {F};
\fill[usageInput] (4.120,1.055) rectangle (7.012,1.325);
\fill[usageOutput] (7.012,1.055) rectangle (7.514,1.325);
\node[anchor=west,font=\sffamily\scriptsize\bfseries] at (7.614,1.190) {93};
\node[font=\sffamily\scriptsize\bfseries] at (3.82,0.810) {S};
\fill[usageInput] (4.120,0.675) rectangle (8.525,0.945);
\fill[usageOutput] (8.525,0.675) rectangle (9.183,0.945);
\node[anchor=west,font=\sffamily\scriptsize\bfseries] at (9.283,0.810) {139};
\node[anchor=north,yshift=-18pt,font=\sffamily\scriptsize\bfseries] at (8.670,0.48) {mean tokens per case (thousands)};
\end{tikzpicture}%
}\par\medskip
\makebox[\linewidth][l]{\textbf{b}\hspace{0.6em}\sffamily\textbf{Token use by pipeline stage and response format}}\par\smallskip
\makebox[\linewidth][c]{%
\begin{tikzpicture}[x=1.40cm,y=0.80cm,font=\sffamily\scriptsize]
\node[font=\sffamily\scriptsize\bfseries] at (4.12,7.12) {Format};
\draw[black!12,line width=0.45pt] (4.420,0.48) -- (4.420,6.65);
\node[anchor=north,text=black!65] at (4.420,0.32) {0};
\draw[black!12,line width=0.45pt] (6.408,0.48) -- (6.408,6.65);
\node[anchor=north,text=black!65] at (6.408,0.32) {25};
\draw[black!12,line width=0.45pt] (8.396,0.48) -- (8.396,6.65);
\node[anchor=north,text=black!65] at (8.396,0.32) {50};
\draw[black!12,line width=0.45pt] (10.384,0.48) -- (10.384,6.65);
\node[anchor=north,text=black!65] at (10.384,0.32) {75};
\draw[black!12,line width=0.45pt] (12.372,0.48) -- (12.372,6.65);
\node[anchor=north,text=black!65] at (12.372,0.32) {100};
\node[anchor=east,font=\sffamily\scriptsize\bfseries] at (3.78,6.000) {Role assignment (shared)};
\node[font=\sffamily\scriptsize\bfseries] at (4.12,6.190) {F};
\fill[usageInput] (4.420,6.055) rectangle (4.562,6.325);
\fill[usageOutput] (4.562,6.055) rectangle (4.589,6.325);
\node[anchor=west,font=\sffamily\scriptsize\bfseries] at (4.689,6.190) {2.1};
\node[font=\sffamily\scriptsize\bfseries] at (4.12,5.810) {S};
\fill[usageInput] (4.420,5.675) rectangle (4.562,5.945);
\fill[usageOutput] (4.562,5.675) rectangle (4.589,5.945);
\node[anchor=west,font=\sffamily\scriptsize\bfseries] at (4.689,5.810) {2.1};
\node[anchor=east,font=\sffamily\scriptsize\bfseries] at (3.78,5.000) {Role-specific opinions};
\node[font=\sffamily\scriptsize\bfseries] at (4.12,5.190) {F};
\fill[usageInput] (4.420,5.055) rectangle (5.056,5.325);
\fill[usageOutput] (5.056,5.055) rectangle (5.698,5.325);
\node[anchor=west,font=\sffamily\scriptsize\bfseries] at (5.798,5.190) {16.1};
\node[font=\sffamily\scriptsize\bfseries] at (4.12,4.810) {S};
\fill[usageInput] (4.420,4.675) rectangle (5.662,4.945);
\fill[usageOutput] (5.662,4.675) rectangle (6.602,4.945);
\node[anchor=west,font=\sffamily\scriptsize\bfseries] at (6.702,4.810) {27.4};
\node[anchor=east,font=\sffamily\scriptsize\bfseries] at (3.78,4.000) {Initial shared report};
\node[font=\sffamily\scriptsize\bfseries] at (4.12,4.190) {F};
\fill[usageInput] (4.420,4.055) rectangle (5.179,4.325);
\fill[usageOutput] (5.179,4.055) rectangle (5.317,4.325);
\node[anchor=west,font=\sffamily\scriptsize\bfseries] at (5.417,4.190) {11.3};
\node[font=\sffamily\scriptsize\bfseries] at (4.12,3.810) {S};
\fill[usageInput] (4.420,3.675) rectangle (5.452,3.945);
\fill[usageOutput] (5.452,3.675) rectangle (5.578,3.945);
\node[anchor=west,font=\sffamily\scriptsize\bfseries] at (5.678,3.810) {14.6};
\node[anchor=east,font=\sffamily\scriptsize\bfseries] at (3.78,3.000) {Role review rounds};
\node[font=\sffamily\scriptsize\bfseries] at (4.12,3.190) {F};
\fill[usageInput] (4.420,3.055) rectangle (8.665,3.325);
\fill[usageOutput] (8.665,3.055) rectangle (8.927,3.325);
\node[anchor=west,font=\sffamily\scriptsize\bfseries] at (9.027,3.190) {56.7};
\node[font=\sffamily\scriptsize\bfseries] at (4.12,2.810) {S};
\fill[usageInput] (4.420,2.675) rectangle (9.201,2.945);
\fill[usageOutput] (9.201,2.675) rectangle (9.547,2.945);
\node[anchor=west,font=\sffamily\scriptsize\bfseries] at (9.647,2.810) {64.5};
\node[anchor=east,font=\sffamily\scriptsize\bfseries] at (3.78,2.000) {Moderator revisions};
\node[font=\sffamily\scriptsize\bfseries] at (4.12,2.190) {F};
\fill[usageInput] (4.420,2.055) rectangle (4.730,2.325);
\fill[usageOutput] (4.730,2.055) rectangle (4.873,2.325);
\node[anchor=west,font=\sffamily\scriptsize\bfseries] at (4.973,2.190) {5.7};
\node[font=\sffamily\scriptsize\bfseries] at (4.12,1.810) {S};
\fill[usageInput] (4.420,1.675) rectangle (4.761,1.945);
\fill[usageOutput] (4.761,1.675) rectangle (4.897,1.945);
\node[anchor=west,font=\sffamily\scriptsize\bfseries] at (4.997,1.810) {6.0};
\node[anchor=east,font=\sffamily\scriptsize\bfseries] at (3.78,1.000) {Final opinion};
\node[font=\sffamily\scriptsize\bfseries] at (4.12,1.190) {F};
\fill[usageInput] (4.420,1.055) rectangle (5.487,1.325);
\fill[usageOutput] (5.487,1.055) rectangle (5.614,1.325);
\node[anchor=west,font=\sffamily\scriptsize\bfseries] at (5.714,1.190) {15.0};
\node[font=\sffamily\scriptsize\bfseries] at (4.12,0.810) {S};
\fill[usageInput] (4.420,0.675) rectangle (5.783,0.945);
\fill[usageOutput] (5.783,0.675) rectangle (5.910,0.945);
\node[anchor=west,font=\sffamily\scriptsize\bfseries] at (6.010,0.810) {18.7};
\node[anchor=north,yshift=-18pt,font=\sffamily\scriptsize\bfseries] at (8.595,0.48) {mean total tokens per case (thousands)};
\end{tikzpicture}%
}
\caption{Provider-reported token use in the eight single-model evaluation pipelines. Blue and orange segments consistently denote prompt/input and completion/output tokens, respectively, in both panels; end labels give their summed mean in thousands. \textbf{a}, Mean complete-pipeline token use per case by model and response format. Each format-specific path includes role assignment; the same case--model role assignment was reused by both formats. \textbf{b}, Mean token use per case by pipeline stage and response format, averaged over the fixed eight-model set. The shared role-assignment value is repeated to show the resource requirement of a standalone format-specific path, but is counted only once in the experiment-wide total. Token units are model-tokenizer-specific and do not represent model-independent computational operations. F denotes free-form and S structured.}
\label{fig:consilium-token-usage}
\end{figure}

\FloatBarrier
\flushbottom

\subsection{Physician Ratings and Judge Calibration}

Six physicians supplied criterion-level reference ratings for Judge-guidance development and internal validation within the fixed 110-case sampling frame. DOI-disjoint development subsets and independent optimization of each criterion preserved case separation and prevented a favorable result for one clinical dimension from compensating for another. This design produced a Judge that could characterize factual grounding, reasoning, management, safety, and conclusion quality separately rather than collapsing them into one undifferentiated score.

Criterion-local optimization reduced macro-mean validation MAE from 0.8601 without additional guidance and 0.9033 with the initial prompts to 0.7798 with the selected guidance, a 9.3\% improvement over the no-additional-guidance configuration. The clearest validation gains occurred for factual grounding, differential diagnosis, and management. 

The held-out benchmark used one fixed Judge configuration across all candidate outputs. Calibration performance remained criterion specific, as intended by the evaluation design. Criterion~6, which jointly assesses conclusion clarity, diagnostic concordance, evidence grounding, and uncertainty, was the most demanding target (MAE, 1.6667). We therefore use the Judge to identify consistent criterion-level and model-level patterns, while interpreting small differences in this composite criterion conservatively.

The evolutionary search improved or preserved training re-evaluation MAE for every criterion and transferred its gains to disjoint validation cases for factual grounding, data use, differential diagnosis, and management (Table~\ref*{tab:evolution-initial-descendant}). The largest validation reductions were observed for factual grounding (1.2222 to 0.6296) and management (0.8889 to 0.5000). These results show that criterion-local guidance can produce substantial numerical agreement gains for clinically interpretable dimensions even when a single universal prompt would obscure them.

\FloatBarrier
\begin{table}[!ht]
\caption{Criterion-specific MAE of the best initial candidates and evolved descendants in the criterion-local calibration run.}\label{tab:evolution-initial-descendant}
\centering
\small
\begin{tabular}{@{}lcccc@{}}
\toprule
 & \multicolumn{2}{c}{\textbf{Training re-evaluation MAE}} & \multicolumn{2}{c}{\textbf{Validation MAE}} \\
\cmidrule(lr){2-3}\cmidrule(lr){4-5}
\textbf{Criterion} & \textbf{Best initial} & \textbf{Best descendant} & \textbf{Best initial} & \textbf{Best descendant} \\
\midrule
1 & $1.0494 \pm 0.1132$ & $0.6296 \pm 0.1481$ & $1.2222 \pm 0.1925$ & $0.6296 \pm 0.1283$ \\
2 & $0.5556 \pm 0.0980$ & $0.5556 \pm 0.0980$ & $0.6296 \pm 0.0642$ & $0.7037 \pm 0.0642$ \\
3 & $0.7963 \pm 0.1604$ & $0.6667 \pm 0.0556$ & $0.7778 \pm 0.0962$ & $0.6667 \pm 0.1667$ \\
4 & $0.8642 \pm 0.0855$ & $0.8395 \pm 0.0771$ & $0.8889 \pm 0.0000$ & $0.8889 \pm 0.0000$ \\
5 & $0.8889 \pm 0.1111$ & $0.5926 \pm 0.0642$ & $0.7222 \pm 0.0962$ & $0.5556 \pm 0.0962$ \\
6 & $1.2963 \pm 0.0642$ & $1.1296 \pm 0.1398$ & $1.6667 \pm 0.1667$ & $1.6667 \pm 0.2887$ \\
7 & $1.0741 \pm 0.0642$ & $0.7222 \pm 0.0000$ & $0.8889 \pm 0.0962$ & $0.5000 \pm 0.1667$ \\
8 & $0.7778 \pm 0.0980$ & $0.7284 \pm 0.0566$ & $1.0000 \pm 0.1111$ & $1.0741 \pm 0.0642$ \\
9 & $0.4444 \pm 0.1925$ & $0.4074 \pm 0.0642$ & $0.3333 \pm 0.0000$ & $0.3333 \pm 0.0000$ \\
\bottomrule
\end{tabular}
\begin{flushleft}
\footnotesize Values are mean $\pm$ sample standard deviation (SD) of criterion-specific penalized MAE over three Judge runs (lower is better). SD was calculated across the three run-level MAEs with divisor $3-1$, not across cases. The initial candidate had the lowest MAE among the two starting prompts in training re-evaluation. The descendant had the lowest MAE among evolved candidates during the search and was then re-evaluated with new random seeds. The same candidates were evaluated on validation cases without further selection. 
\end{flushleft}
\end{table}
\FloatBarrier

\subsection{Held-Out Benchmark Performance}

All eight model pipelines were evaluated on the same 190 held-out cases in both response formats, with complete generation coverage and no final Judge failures. The analysis included 6,080 response components and 54,720 recorded criterion outcomes. By evaluating the same cases at the role-opinion and final-synthesis checkpoints, the benchmark allowed model capability, response format, and the contribution of multidisciplinary synthesis to be examined separately on the original 0--3 criterion scale (Table~\ref{tab:model-leaderboard}).

\begin{table}[!ht]
\centering\small
\caption{Descriptive Judge-score summaries across the three evaluation components in the \ConsModelCountWord{}-model evaluation on 190 cases per model.}\label{tab:model-leaderboard}
\begin{tabular}{@{}lcccc@{}}
\toprule
Model & \shortstack{Role reasoning\\Structured} & \shortstack{Role clinical opinion\\Structured} & \shortstack{Conclusion\\Free-form} & \shortstack{Conclusion\\Structured} \\
\midrule
GPT-5.6-Sol & {\bfseries\boldmath $2.892 \pm 0.409$} & {\bfseries\boldmath $2.486 \pm 0.906$} & {\bfseries\boldmath $2.708 \pm 0.700$} & {\bfseries\boldmath $2.772 \pm 0.643$} \\
GPT-OSS-120B & $2.417 \pm 0.887$ & $1.983 \pm 1.108$ & $2.543 \pm 0.861$ & $2.569 \pm 0.839$ \\
DeepSeek-V4-Pro & $2.441 \pm 0.841$ & $2.033 \pm 1.053$ & $2.514 \pm 0.833$ & $2.526 \pm 0.863$ \\
GLM-5.2 & $2.462 \pm 0.806$ & $2.082 \pm 1.036$ & $2.491 \pm 0.793$ & $2.516 \pm 0.800$ \\
Qwen3.5-27B & $2.259 \pm 0.937$ & $1.943 \pm 1.086$ & $2.448 \pm 0.894$ & $2.451 \pm 0.885$ \\
Qwen3-235B & $2.323 \pm 0.948$ & $1.863 \pm 1.084$ & $2.365 \pm 0.929$ & $2.389 \pm 0.988$ \\
Gemma-4-31B & $2.252 \pm 0.861$ & $2.007 \pm 1.047$ & $2.322 \pm 0.877$ & $2.360 \pm 0.862$ \\
Qwen3-Next-80B & $2.091 \pm 1.032$ & $1.740 \pm 1.137$ & $2.245 \pm 1.011$ & $2.200 \pm 1.054$ \\
\bottomrule
\end{tabular}
\begin{flushleft}\footnotesize
Cells show the numeric mean $\pm$ population SD on the original 0--3 scale. Models are ordered by their identical final-report ranking in the free-form and structured conditions. The SD denominator is the number of numeric, structurally applicable Judge scores in the cell; response-dependent not-evaluable outcomes are excluded. Bold denotes the largest mean in each component column. Criterion applicability and numeric coverage are reported in Appendix Tables~\ref{app:judge-applicability} and~\ref{app:judge-coverage}. Role-level results concern one primary role per case; no role-level Judge scores were available for free-form outputs. GPT-OSS-120B also served as the Judge, so its comparison is not evaluator-independent and may reflect self-preference.
\end{flushleft}
\end{table}
\FloatBarrier

Across this fixed benchmark, model identity produced the largest descriptive separation in overall performance. GPT-5.6-Sol ranked first in structured role reasoning, role clinical opinion, free-form conclusions, and structured conclusions; Qwen3-Next-80B ranked last in all four components. GPT-5.6-Sol exceeded the next-highest model by 0.430 points for role reasoning, 0.404 for role clinical opinion, 0.165 for free-form conclusions, and 0.203 for structured conclusions. The best-to-worst model spread in final reports was 0.463 points for free-form and 0.572 for structured outputs, substantially larger than every within-model format difference ($-0.045$ to $+0.064$). The model ordering of final reports was identical in the two response formats, showing that the benchmark recovered a stable model hierarchy and that model choice was more strongly associated with final-report quality than the global response schema.

Structured conclusions had higher overall means than their free-form counterparts in seven of the eight model pipelines. GPT-5.6-Sol showed the largest structured-minus-free-form difference ($+0.064$ points), followed by Gemma-4-31B ($+0.038$); Qwen3-Next-80B was the only pipeline with a lower structured mean ($-0.045$). Thus, the model-level comparison showed a broad, although modest, advantage for structured final reporting.

To contextualize these quality differences, Table~\ref{tab:quality-inference} presents final-report Judge scores alongside complete-pipeline token use for each model and response format on the same 190 held-out cases. For GPT-5.6-Sol, the mean score increased from 2.708 with free-form reporting to 2.772 with structured reporting, while mean token use increased from 65.2 thousand to 132.1 thousand per case. These comparisons describe the inference requirements accompanying the observed quality differences.

\begin{table}[!ht]
\centering\small
\caption{Judge-rated quality and observed inference requirements by model and response format.}\label{tab:quality-inference}
\begin{tabular}{@{}llrrrr@{}}
\toprule
Model & Format & \shortstack{Mean Judge\\score} & \shortstack{Input tokens/case\\(thousands)} & \shortstack{Output tokens/case\\(thousands)} & \shortstack{Total tokens/case\\(thousands)} \\
\midrule
GPT-5.6-Sol & Free-form & 2.708 & 55.3 & 9.9 & 65.2 \\
 & Structured & 2.772 & 108.5 & 23.5 & 132.1 \\
\addlinespace
GPT-OSS-120B & Free-form & 2.543 & 164.9 & 42.3 & 207.2 \\
 & Structured & 2.569 & 163.8 & 45.5 & 209.3 \\
\addlinespace
DeepSeek-V4-Pro & Free-form & 2.514 & 84.5 & 14.0 & 98.4 \\
 & Structured & 2.526 & 87.0 & 17.0 & 103.9 \\
\addlinespace
GLM-5.2 & Free-form & 2.491 & 91.8 & 15.3 & 107.0 \\
 & Structured & 2.516 & 106.2 & 18.1 & 124.2 \\
\addlinespace
Qwen3.5-27B & Free-form & 2.448 & 124.3 & 17.8 & 142.1 \\
 & Structured & 2.451 & 134.7 & 18.3 & 152.9 \\
\addlinespace
Qwen3-235B & Free-form & 2.365 & 77.0 & 15.3 & 92.3 \\
 & Structured & 2.389 & 113.7 & 21.7 & 135.4 \\
\addlinespace
Gemma-4-31B & Free-form & 2.322 & 43.1 & 6.3 & 49.5 \\
 & Structured & 2.360 & 60.6 & 9.1 & 69.6 \\
\addlinespace
Qwen3-Next-80B & Free-form & 2.245 & 79.4 & 13.8 & 93.2 \\
 & Structured & 2.200 & 121.0 & 18.1 & 139.1 \\
\bottomrule
\end{tabular}
\begin{flushleft}\footnotesize
Judge scores pool available numeric ratings for criteria~1--8 on the original 0--3 scale, with not-evaluable outcomes excluded, as in Table~\ref{tab:model-leaderboard}. Token values are provider-reported means per case over 190 cases per configuration, expressed in thousands. They include the complete generation pipeline, from role assignment through finalization, and exclude Judge/evaluation calls. The shared case--model role assignment is included in each standalone format-specific path. Columns are rounded independently. Token units are model-tokenizer-specific; these descriptive comparisons do not represent matched compute budgets.
\end{flushleft}
\end{table}
\FloatBarrier

The complete multidisciplinary synthesis had higher pooled Judge means for every criterion shared with the isolated primary-role components (Table~\ref{tab:component-criteria}). Relative to structured role reasoning, the structured final report was higher by 0.650 points in factual grounding, 0.302 in reasoning coherence, 0.225 in safety, and 0.210 in problem prioritization; data use, differential diagnosis, and management were also higher. Relative to the role-specific clinical opinion, the final report was higher by 0.515 points in problem prioritization, 0.377 in factual grounding, 0.357 in reasoning coherence, 0.111 in clinical-conclusion quality, and 0.104 in safety. The largest gains therefore concerned integration of case facts, prioritization of the clinical problem, and construction of a coherent account. These findings identify a clear value of the complete synthesis stage: it transformed isolated role-specific material into a more highly rated, integrated clinical second opinion across every comparable dimension.

Structured prompting had its clearest descriptive advantage in actionability. Management was higher by 0.304 points (2.740 vs.\ 2.436), with the same direction in seven of eight model pipelines; problem prioritization was higher by 0.056 and factual grounding by 0.040. The overall final-report mean was likewise higher in seven of eight models. The criterion-level advantage was concentrated rather than uniform: reasoning coherence was essentially unchanged, while data use, differential diagnosis, clinical-conclusion quality, and safety were modestly lower. The broad model-level advantage of structured reports was therefore driven primarily by the comparatively large management gain, which outweighed smaller decreases in several other criteria. Structured prompting improved the actionability and organization of the second opinion rather than every aspect of its clinical profile.

\begin{table}[!ht]
\centering\small
\caption{Criterion-specific profiles pooled across the \ConsModelCountWord{} models.}\label{tab:component-criteria}
\begin{tabular}{@{}lcccc@{}}
\toprule
Criterion & \shortstack{Role reasoning\\Structured} & \shortstack{Role clinical opinion\\Structured} & \shortstack{Conclusion\\Free-form} & \shortstack{Conclusion\\Structured} \\
\midrule
1: Factual grounding & $1.743 \pm 1.251$ & $2.016 \pm 1.169$ & $2.353 \pm 1.052$ & {\bfseries\boldmath $2.393 \pm 1.055$} \\
2: Problem prioritization & $2.618 \pm 0.570$ & $2.313 \pm 0.672$ & $2.772 \pm 0.477$ & {\bfseries\boldmath $2.828 \pm 0.406$} \\
3: Data use & $2.517 \pm 0.607$ & - & {\bfseries\boldmath $2.692 \pm 0.506$} & $2.636 \pm 0.533$ \\
4: Reasoning coherence & $2.517 \pm 0.851$ & $2.462 \pm 0.803$ & {\bfseries\boldmath $2.822 \pm 0.564$} & $2.819 \pm 0.566$ \\
5: Differential diagnosis & $2.735 \pm 0.670$ & - & {\bfseries\boldmath $2.807 \pm 0.509$} & $2.765 \pm 0.483$ \\
6: Clinical conclusion & - & $1.270 \pm 1.283$ & {\bfseries\boldmath $1.416 \pm 1.149$} & $1.381 \pm 1.208$ \\
7: Management & $2.712 \pm 0.543$ & - & $2.436 \pm 0.672$ & {\bfseries\boldmath $2.740 \pm 0.484$} \\
8: Safety & $1.903 \pm 0.899$ & $2.024 \pm 0.929$ & {\bfseries\boldmath $2.243 \pm 0.908$} & $2.128 \pm 0.957$ \\
\bottomrule
\end{tabular}
\begin{flushleft}\footnotesize
Cells show the pooled numeric mean $\pm$ population SD on the original 0--3 scale across the eight model pipelines. The SD denominator is the number of numeric, structurally applicable Judge scores in the cell, response-dependent not-evaluable outcomes are excluded. A dash denotes structural non-applicability. Bold denotes the largest mean across applicable evaluation components in each criterion row. Applicability, coverage, and non-pooled model-specific profiles are reported in Appendix Tables~\ref{app:judge-applicability}--\ref{app:judge-profile-glm52}.
\end{flushleft}
\end{table}
\FloatBarrier

\subsection{Distribution of Judge Scores Across Cases and Criteria}

\begin{figure}[!ht]
\centering
\includegraphics[width=\textwidth]{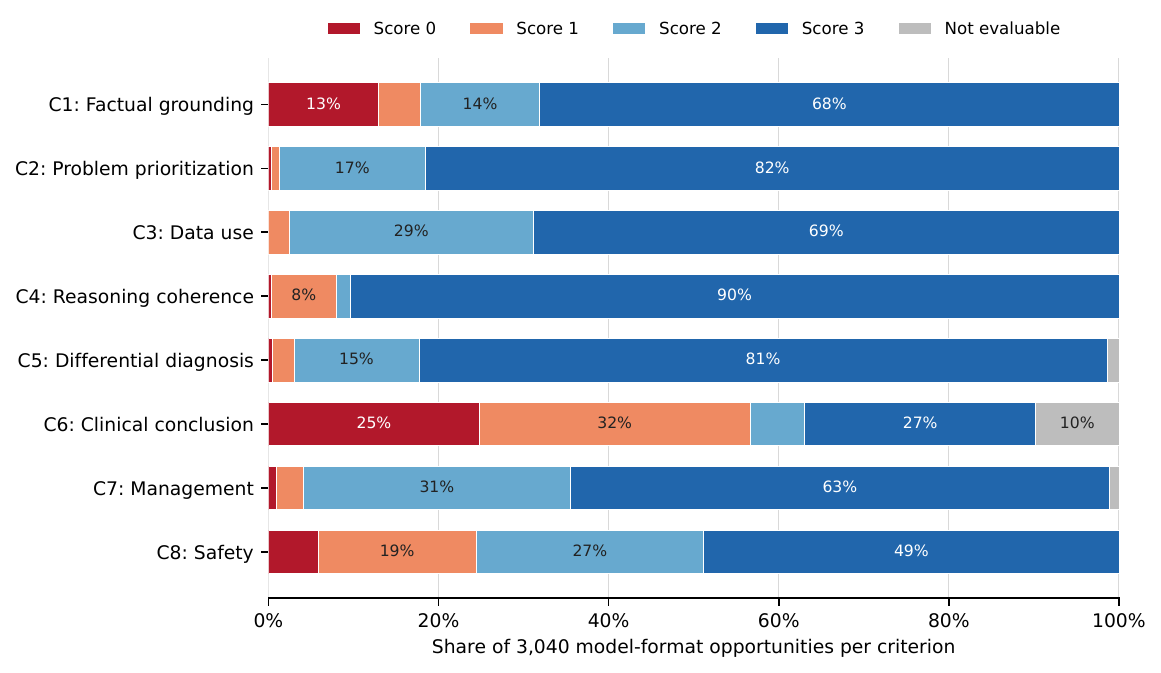}
\caption{Distribution of Judge outcomes for multidisciplinary conclusions by criterion. Each horizontal bar represents all \ConsTotalArms{} model-format opportunities for one criterion; the score categories and not-evaluable outcomes therefore sum to 100\%.}
\label{fig:judge-score-outcomes}
\end{figure}

\begin{figure}[!ht]
\centering
\includegraphics[width=\textwidth]{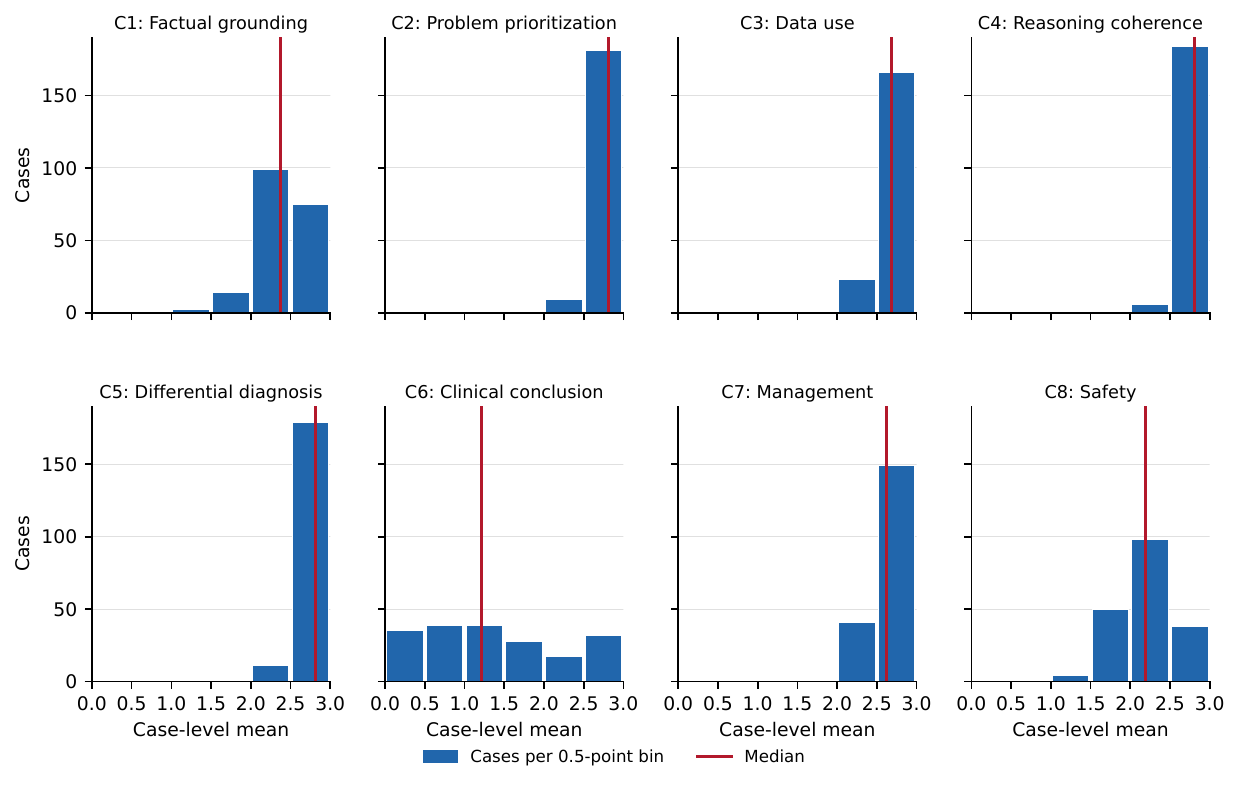}
\caption{Distribution of case-level mean Judge scores for multidisciplinary conclusions. For each case and criterion, the mean was calculated over numeric scores from \ConsModelCountWord{} models and two response formats. Bars count cases in 0.5-point bins; vertical red lines show the median. Not-evaluable outcomes were excluded from the means.}
\label{fig:judge-case-means}
\end{figure}

The criterion distributions revealed a clear aggregate pattern across the \ConsTotalArms{} final reports. Problem prioritization, reasoning coherence, and differential diagnosis were concentrated at the top of the scale, whereas final-conclusion quality and safety accounted for most of the clinically important residual variation (Figures~\ref{fig:judge-score-outcomes} and~\ref{fig:judge-case-means}). Each case contributed once to the case-level distributions after averaging the available scores across the \ConsModelFormatCombinations{} model--format combinations.

\FloatBarrier
\FloatBarrier

The criterion-level results were strongest for problem prioritization, reasoning coherence, and differential diagnosis: 81.5\% received the maximum score for problem prioritization, 90.4\% for reasoning coherence, and 80.9\% for differential diagnosis. Under the frozen Judge, the principal measured weakness appeared when this strong reasoning had to be converted into a definitive, evidence-calibrated clinical conclusion. Among numeric criterion~6 scores, 62.8\% were 0 or~1. Safety was the second least-saturated dimension: 48.8\% of reports received the maximum score, leaving substantial room to strengthen monitoring, contraindication handling, and escalation guidance. The benchmark therefore distinguishes fluent diagnostic reasoning from the harder task of producing a highly rated, calibrated, and safe final decision.

\FloatBarrier
\FloatBarrier

Case-level distributions reinforced this separation. Problem prioritization, reasoning coherence, and differential diagnosis each had a median of 2.812, with narrow interquartile ranges near the top of the scale. Clinical-conclusion quality had a markedly lower median of 1.215 (IQR, 0.667--2.000) and the widest between-case spread. Factual grounding (median, 2.375) and safety (median, 2.188) occupied intermediate positions. These profiles identify conclusion formation, rather than generation of a plausible differential, as the main target for further improvement.

\subsection{Judge Scores and Consilium Completion Rounds}
\label{sec:judge-rounds}

Review duration varied strongly by model--format configuration and showed no monotonic relation with Judge-rated case complexity. Qwen3.5-27B produced no one-round reports in either format, and rounds~4--5 were concentrated in GPT-OSS-120B and Qwen3.5-27B (Table~\ref{tab:consilium-process}). Completed rounds therefore captured persistent role-conditioned disagreement more clearly than case difficulty.

GPT-5.6-Sol combined the highest Judge scores with the shortest review trajectories. It averaged 1.22 review rounds for free-form reports and 1.23 for structured reports, with first-round unanimous acceptance in 81.6\% and 79.5\% of cases and eventual report consensus in 100.0\% and 98.9\%, respectively. This combination identifies GPT-5.6-Sol as both the quality leader and the most rapidly converging pipeline in the evaluated cohort.

\begin{table}[!ht]
\centering\small
\caption{Final-answer Judge scores and case complexity by completed review rounds.}\label{tab:judge-rounds-summary}
\begin{tabular}{@{}lrrrr@{}}
\toprule
Completed rounds & Outputs & \shortstack{C1--C8 output mean\\Mean $\pm$ population SD} & \shortstack{Output mean $<2$\\No. (\%)} & \shortstack{C10 complexity\\Mean $\pm$ population SD} \\
\midrule
1 & 1,214 & 2.525 $\pm$ 0.326 & 62 (5.1) & 2.148 $\pm$ 0.543 \\
2 & 1,377 & 2.419 $\pm$ 0.338 & 123 (8.9) & 2.131 $\pm$ 0.537 \\
$\geq$3 & 449 & 2.438 $\pm$ 0.330 & 33 (7.3) & 2.158 $\pm$ 0.494 \\
\bottomrule
\end{tabular}
\begin{flushleft}\footnotesize
The C1--C8 output mean averages available numeric scores for one case, model, and response format; not-evaluable outcomes were excluded. A mean below~2 is a descriptive rubric boundary, not a readiness threshold. Criterion~10 was analyzed separately.
\end{flushleft}
\end{table}
\FloatBarrier

\begin{figure}[!ht]
\centering
\includegraphics[width=\textwidth]{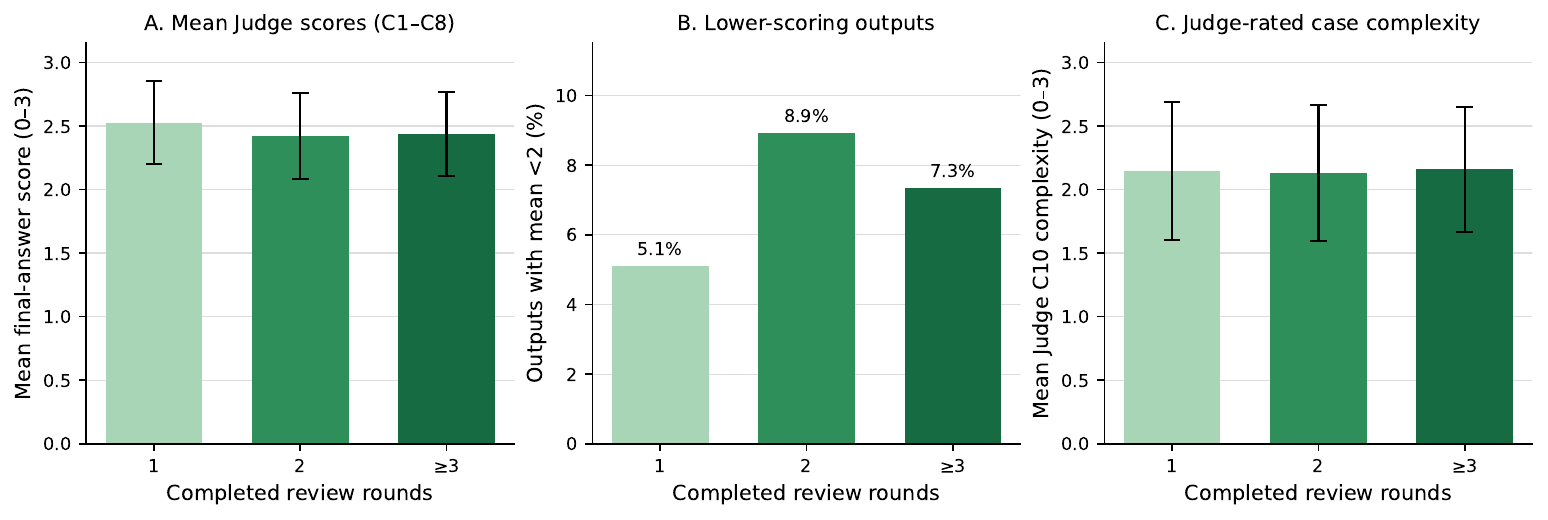}
\caption{Judge scores and case complexity by completed consilium review rounds. Panels show descriptive summaries for one, two, and at least three rounds. Error bars in panels A and C show population SD; panel B shows the proportion of outputs with a mean score below~2. The mean in panels A and B averages available numeric scores for criteria~1--8; criterion~10 was analyzed separately.}
\label{fig:judge-rounds-quality}
\end{figure}

Longer review trajectories were not associated with higher Judge-rated quality. Mean final-report scores were 2.525 after one round, 2.419 after two rounds, and 2.438 after three or more rounds. The proportion of outputs with a mean score below 2 was also lowest after one round (5.1\% vs.\ 8.9\% and 7.3\%). Mean Judge-rated case-complexity scores were nearly identical across these groups (2.148, 2.131, and 2.158). In these descriptive comparisons, outputs completed after additional review rounds showed no quality advantage. This pattern supports prospective evaluation of an early-stopping strategy but does not estimate the effect of terminating an ongoing review earlier (Table~\ref{tab:judge-rounds-summary}; Figure~\ref{fig:judge-rounds-quality}).

\FloatBarrier
\FloatBarrier

\section{Discussion}


\subsection{Principal Findings}

This study produced four principal findings. First, model identity separated final-report quality more strongly than response format: GPT-5.6-Sol led every evaluated component, the model ordering was identical across final-report formats, and the best-to-worst spread substantially exceeded all within-model format differences. Second, the complete structured synthesis had higher pooled Judge means than the pre-consilium primary-role opinion for every shared criterion, with the largest gains in problem prioritization, factual grounding, and reasoning coherence. Third, structured final reports achieved higher overall means in seven of eight model pipelines; this broad model-level advantage was driven primarily by a 0.304-point improvement in management, which was also observed in seven pipelines. Fourth, the review protocol reached eventual consensus in 97.8\% of reports, whereas longer review trajectories had no final-report score advantage. GPT-5.6-Sol combined the highest report quality with the fewest review rounds, linking strong initial performance with rapid convergence in this protocol.

The stage comparison showed a favorable aggregate pattern for the full synthesis pipeline. Its final reports had higher pooled Judge means for every criterion comparable with the structured primary-role output. The largest differences occurred in problem prioritization, grounding, and coherence, consistent with the consilium integrating case evidence and role-specific perspectives into a clearer and more clinically organized account. The result is important because it demonstrates value beyond generation of additional text: the final synthesis more effectively assembled the available material into a complete clinical second opinion. The comparison represents the contribution of the full synthesis pipeline, including aggregation, review, revision, and finalization, rather than an isolated causal effect of discussion alone.

Structured prompting had its clearest advantage when the response required explicit action planning. Its management difference was the largest format-associated criterion contrast and had the same direction across seven of eight models. This comparatively large gain explains why structured reports had higher overall means in seven pipelines even though several smaller criterion differences favored free-form reports. Smaller advantages in prioritization and grounding further indicate that the implemented structured workflow produced a more organized and actionable plan. Structured prompting therefore added value through management quality and organization rather than through uniform improvement of every criterion.

\subsection{Value of the Benchmark}

The benchmark was designed to evaluate a clinical second opinion as a complete decision-support artifact rather than as a single diagnostic answer. Its eight response-quality criteria separately measure whether a model uses the supplied facts correctly, identifies and prioritizes the clinical problems, integrates relevant data, constructs coherent reasoning, develops an appropriate differential diagnosis, reaches a calibrated conclusion, proposes an actionable management plan, and addresses safety. This profile makes it possible to distinguish models that produce fluent diagnostic reasoning from those that can also convert that reasoning into a grounded, usable, and safe clinical recommendation.

The two evaluation checkpoints add a second level of resolution. Retaining the primary-role opinion before consilium and evaluating the final multidisciplinary synthesis on the same cases makes it possible to determine whether the complete synthesis pipeline adds value and which clinical dimensions change. Likewise, applying free-form and structured conditions to the same case--model panels separates a broad response-format effect from the substantially larger effect of model identity. Table~\ref{tab:model-leaderboard} demonstrates this capability directly: it recovers a stable model hierarchy while Table~\ref{tab:component-criteria} shows that the apparent advantage of structured reports is specifically concentrated in management rather than distributed uniformly across the rubric.

Finally, the benchmark records the process that produces the final answer. First-round dissent, consensus frequency, review rounds, the distinct finalization stage, and token use are reported alongside clinical-quality scores. This makes it possible to distinguish procedural agreement from output quality and to identify configurations, such as GPT-5.6-Sol, that combine strong clinical scores with rapid convergence. Together, these features provide a comprehensive and extensible framework for comparing clinically relevant model capabilities, prompting strategies, and synthesis pipelines without reducing performance to diagnostic accuracy alone.

\section{Conclusion}

This benchmark provides a multidimensional evaluation of LLM-generated clinical second opinions across factual grounding, problem prioritization, data use, reasoning coherence, differential diagnosis, final-conclusion quality, management, and safety. On the same 190 held-out cases, it distinguished a stable hierarchy among eight model pipelines, separated model capability from response-format effects, and quantified the contribution of the complete multidisciplinary synthesis. The structured final consilium report was more highly rated than the pre-consilium primary-role opinion across every shared criterion, while structured reports achieved higher overall means in seven of eight models, primarily through improved management planning. GPT-5.6-Sol led every evaluated component and reached consensus in the fewest review rounds.

These results show that a clinically informative benchmark can capture differences that diagnostic accuracy alone would miss: whether a model can transform case information and multiple perspectives into a grounded, coherent, actionable, and safe second opinion. At the same time, structured reporting increased inference requirements, and additional review rounds were not associated with higher final-report scores. Clinical quality, procedural convergence, response structure, and inference cost should therefore be evaluated as complementary outcomes. The benchmark offers a comprehensive framework for comparing future models and synthesis strategies, while prospective studies remain necessary to establish effects on clinician decisions and patient care.

\section{Limitations}

NEJM Case Records overrepresent unusual, difficult cases and do not reproduce routine prevalence, documentation, time pressure, or longitudinal uncertainty. Their published diagnoses are useful targets but not exhaustive ground truth for management. The public cases may also have appeared in model training data; because DOI and title were supplied before the consilium, we cannot separate memorization from de novo reasoning. Finally, prompted roles are simulated perspectives, not independently validated specialists, and may share correlated errors.

The consilium used more calls, context, and tokens than a standalone opinion, without a compute-matched design that could isolate sampling, review, and interaction. The structured condition also changed instructions, output length, and downstream context, so observed differences cannot be attributed to format alone.

Judge scores may reflect knowledge, style preferences, or failure modes shared with evaluated models; GPT-OSS-120B also judged its own outputs. Calibration used a limited physician-rated sample, error varied by criterion, and several evolved prompts showed semantic drift. The rubric was not externally validated, the held-out cohort was published later than the reference cohort, and exclusion of non-evaluable scores could affect means. Panel size showed little variation. Most importantly, we evaluated artifacts rather than effects on clinician behavior, workflow, or patient outcomes.

\bibliographystyle{unsrtnat}
\bibliography{references}

\clearpage
\appendix

\section{Expanded Judge Score Summaries}
\label{app:judge-score-summaries}
The main-text tables omit applicability counts to keep the score summaries comparable. Table~\ref{app:judge-applicability} gives the prespecified criterion sets, and Tables~\ref{app:judge-coverage} and~\ref{app:judge-criterion-coverage} give numeric coverage. Tables~\ref{app:judge-profile-gptoss}--\ref{app:judge-profile-glm52} disaggregate the criterion profiles by model. All standard deviations in this section are population SDs, $\sqrt{\sum_i(x_i-\bar{x})^2/n}$, calculated over the numeric Judge scores shown in the corresponding cell; structurally inapplicable and response-dependent not-evaluable outcomes are not assigned numeric values.

\begin{table}[!ht]
\centering\small
\caption{Criterion applicability by evaluation component.}\label{app:judge-applicability}
\begin{tabular}{@{}llp{4.2cm}rr@{}}
\toprule
Component & Format & Applicable criteria & No. of criteria & Opportunities per model \\
\midrule
Role reasoning & Structured & 1, 2, 3, 4, 5, 7, 8 & 7 & 1330 \\
Role clinical opinion & Structured & 1, 2, 4, 6, 8 & 5 & 950 \\
Conclusion & Free-form & 1, 2, 3, 4, 5, 6, 7, 8 & 8 & 1520 \\
Conclusion & Structured & 1, 2, 3, 4, 5, 6, 7, 8 & 8 & 1520 \\
\bottomrule
\end{tabular}
\begin{flushleft}\footnotesize
Criterion numbers correspond to the eight response-quality criteria in Table~\ref{tab:component-criteria}. Opportunities per model equal 190 cases multiplied by the number of applicable criteria. Panel adequacy and source-case complexity were scored separately and do not enter the response-quality means.
\end{flushleft}
\end{table}
\FloatBarrier

\begin{table}[!ht]
\centering\small
\caption{Numeric Judge-score coverage by model and evaluation component.}\label{app:judge-coverage}
\begin{tabular}{@{}lcccc@{}}
\toprule
Model & \shortstack{Role reasoning\\Structured} & \shortstack{Role clinical opinion\\Structured} & \shortstack{Conclusion\\Free-form} & \shortstack{Conclusion\\Structured} \\
\midrule
GPT-OSS-120B & 1330/1330 & 949/950 & 1498/1520 & 1491/1520 \\
Qwen3.5-27B & 1330/1330 & 950/950 & 1502/1520 & 1503/1520 \\
Qwen3-235B & 1330/1330 & 950/950 & 1513/1520 & 1509/1520 \\
Gemma-4-31B & 1330/1330 & 950/950 & 1418/1520 & 1503/1520 \\
Qwen3-Next-80B & 1330/1330 & 950/950 & 1501/1520 & 1500/1520 \\
GPT-5.6-Sol & 1330/1330 & 949/950 & 1518/1520 & 1519/1520 \\
DeepSeek-V4-Pro & 1330/1330 & 950/950 & 1502/1520 & 1504/1520 \\
GLM-5.2 & 1330/1330 & 950/950 & 1470/1520 & 1495/1520 \\
\bottomrule
\end{tabular}
\begin{flushleft}\footnotesize
The numerator is the number of numeric scores and the denominator is the number of structurally applicable criterion opportunities. The difference is the number of response-dependent not-evaluable outcomes.
\end{flushleft}
\end{table}
\FloatBarrier

\begin{table}[!ht]
\centering\small
\caption{Criterion-specific numeric Judge-score coverage pooled across the \ConsModelCountWord{} models.}\label{app:judge-criterion-coverage}
\begin{tabular}{@{}lcccc@{}}
\toprule
Criterion & \shortstack{Role reasoning\\Structured} & \shortstack{Role clinical opinion\\Structured} & \shortstack{Conclusion\\Free-form} & \shortstack{Conclusion\\Structured} \\
\midrule
1: Factual grounding & 1520/1520 & 1520/1520 & 1520/1520 & 1520/1520 \\
2: Problem prioritization & 1520/1520 & 1520/1520 & 1520/1520 & 1520/1520 \\
3: Data use & 1520/1520 & - & 1520/1520 & 1520/1520 \\
4: Reasoning coherence & 1520/1520 & 1518/1520 & 1520/1520 & 1520/1520 \\
5: Differential diagnosis & 1520/1520 & - & 1480/1520 & 1518/1520 \\
6: Clinical conclusion & - & 1520/1520 & 1355/1520 & 1387/1520 \\
7: Management & 1520/1520 & - & 1487/1520 & 1519/1520 \\
8: Safety & 1520/1520 & 1520/1520 & 1520/1520 & 1520/1520 \\
\bottomrule
\end{tabular}
\begin{flushleft}\footnotesize
The numerator is the number of numeric scores and the denominator is the number of structurally applicable opportunities pooled across \ConsModelCountWord{} models. A dash denotes structural non-applicability.
\end{flushleft}
\end{table}
\FloatBarrier

\begin{table}[!ht]
\centering\small
\setlength{\tabcolsep}{4pt}
\caption{Model-specific criterion profile for GPT-OSS-120B.}\label{app:judge-profile-gptoss}
\begin{tabular}{@{}p{0.28\linewidth}*{4}{>{\centering\arraybackslash}p{0.16\linewidth}}@{}}
\toprule
Criterion & \shortstack{Role reasoning\\Structured} & \shortstack{Role clinical opinion\\Structured} & \shortstack{Conclusion\\Free-form} & \shortstack{Conclusion\\Structured} \\
\midrule
1: Factual grounding & $1.789 \pm 1.196$ & $2.021 \pm 1.147$ & $2.468 \pm 0.868$ & $2.674 \pm 0.807$ \\
2: Problem prioritization & $2.616 \pm 0.547$ & $2.463 \pm 0.646$ & $2.879 \pm 0.326$ & $2.895 \pm 0.339$ \\
3: Data use & $2.642 \pm 0.560$ & - & $2.805 \pm 0.396$ & $2.732 \pm 0.455$ \\
4: Reasoning coherence & $2.584 \pm 0.795$ & $2.593 \pm 0.719$ & $2.937 \pm 0.334$ & $2.911 \pm 0.394$ \\
5: Differential diagnosis & $2.732 \pm 0.686$ & - & $2.878 \pm 0.403$ & $2.718 \pm 0.462$ \\
6: Clinical conclusion & - & $1.000 \pm 1.240$ & $1.294 \pm 1.209$ & $1.258 \pm 1.216$ \\
7: Management & $2.842 \pm 0.499$ & - & $3.000 \pm 0.000$ & $2.958 \pm 0.248$ \\
8: Safety & $1.716 \pm 0.908$ & $1.842 \pm 0.874$ & $2.011 \pm 0.900$ & $2.221 \pm 0.908$ \\
\bottomrule
\end{tabular}
\begin{flushleft}\footnotesize
Cells show the model-specific numeric mean $\pm$ population SD. A dash denotes structural non-applicability. Aggregate component coverage is reported in Table~\ref{app:judge-coverage}; criterion-specific pooled coverage is reported in Table~\ref{app:judge-criterion-coverage}.
\end{flushleft}
\end{table}
\FloatBarrier

\begin{table}[!ht]
\centering\small
\setlength{\tabcolsep}{4pt}
\caption{Model-specific criterion profile for Qwen3.5-27B.}\label{app:judge-profile-qwen35}
\begin{tabular}{@{}p{0.28\linewidth}*{4}{>{\centering\arraybackslash}p{0.16\linewidth}}@{}}
\toprule
Criterion & \shortstack{Role reasoning\\Structured} & \shortstack{Role clinical opinion\\Structured} & \shortstack{Conclusion\\Free-form} & \shortstack{Conclusion\\Structured} \\
\midrule
1: Factual grounding & $1.463 \pm 1.208$ & $1.953 \pm 1.153$ & $2.442 \pm 0.997$ & $2.516 \pm 0.922$ \\
2: Problem prioritization & $2.568 \pm 0.601$ & $2.226 \pm 0.708$ & $2.795 \pm 0.441$ & $2.811 \pm 0.430$ \\
3: Data use & $2.368 \pm 0.642$ & - & $2.663 \pm 0.525$ & $2.505 \pm 0.578$ \\
4: Reasoning coherence & $2.363 \pm 0.906$ & $2.437 \pm 0.770$ & $2.905 \pm 0.425$ & $2.884 \pm 0.456$ \\
5: Differential diagnosis & $2.595 \pm 0.794$ & - & $2.788 \pm 0.535$ & $2.816 \pm 0.450$ \\
6: Clinical conclusion & - & $1.174 \pm 1.296$ & $1.331 \pm 1.171$ & $1.283 \pm 1.270$ \\
7: Management & $2.711 \pm 0.585$ & - & $2.487 \pm 0.681$ & $2.647 \pm 0.510$ \\
8: Safety & $1.747 \pm 0.888$ & $1.926 \pm 0.937$ & $2.132 \pm 0.928$ & $2.042 \pm 0.888$ \\
\bottomrule
\end{tabular}
\begin{flushleft}\footnotesize
Cells show the model-specific numeric mean $\pm$ population SD. A dash denotes structural non-applicability. Aggregate component coverage is reported in Table~\ref{app:judge-coverage}; criterion-specific pooled coverage is reported in Table~\ref{app:judge-criterion-coverage}.
\end{flushleft}
\end{table}
\FloatBarrier

\begin{table}[!ht]
\centering\small
\setlength{\tabcolsep}{4pt}
\caption{Model-specific criterion profile for Qwen3-235B.}\label{app:judge-profile-qwen235}
\begin{tabular}{@{}p{0.28\linewidth}*{4}{>{\centering\arraybackslash}p{0.16\linewidth}}@{}}
\toprule
Criterion & \shortstack{Role reasoning\\Structured} & \shortstack{Role clinical opinion\\Structured} & \shortstack{Conclusion\\Free-form} & \shortstack{Conclusion\\Structured} \\
\midrule
1: Factual grounding & $1.332 \pm 1.257$ & $1.689 \pm 1.207$ & $1.889 \pm 1.202$ & $1.895 \pm 1.273$ \\
2: Problem prioritization & $2.679 \pm 0.510$ & $2.237 \pm 0.617$ & $2.811 \pm 0.405$ & $2.874 \pm 0.332$ \\
3: Data use & $2.547 \pm 0.558$ & - & $2.747 \pm 0.481$ & $2.711 \pm 0.498$ \\
4: Reasoning coherence & $2.379 \pm 0.925$ & $2.305 \pm 0.907$ & $2.663 \pm 0.748$ & $2.647 \pm 0.779$ \\
5: Differential diagnosis & $2.789 \pm 0.639$ & - & $2.847 \pm 0.507$ & $2.879 \pm 0.424$ \\
6: Clinical conclusion & - & $1.079 \pm 1.169$ & $1.228 \pm 1.012$ & $1.257 \pm 1.168$ \\
7: Management & $2.721 \pm 0.563$ & - & $2.368 \pm 0.616$ & $2.795 \pm 0.518$ \\
8: Safety & $1.816 \pm 0.872$ & $2.005 \pm 0.921$ & $2.337 \pm 0.835$ & $1.989 \pm 1.005$ \\
\bottomrule
\end{tabular}
\begin{flushleft}\footnotesize
Cells show the model-specific numeric mean $\pm$ population SD. A dash denotes structural non-applicability. Aggregate component coverage is reported in Table~\ref{app:judge-coverage}; criterion-specific pooled coverage is reported in Table~\ref{app:judge-criterion-coverage}.
\end{flushleft}
\end{table}
\FloatBarrier

\begin{table}[!ht]
\centering\small
\setlength{\tabcolsep}{4pt}
\caption{Model-specific criterion profile for Gemma-4-31B.}\label{app:judge-profile-gemma4}
\begin{tabular}{@{}p{0.28\linewidth}*{4}{>{\centering\arraybackslash}p{0.16\linewidth}}@{}}
\toprule
Criterion & \shortstack{Role reasoning\\Structured} & \shortstack{Role clinical opinion\\Structured} & \shortstack{Conclusion\\Free-form} & \shortstack{Conclusion\\Structured} \\
\midrule
1: Factual grounding & $1.737 \pm 1.220$ & $2.132 \pm 1.142$ & $2.789 \pm 0.587$ & $2.647 \pm 0.825$ \\
2: Problem prioritization & $2.416 \pm 0.689$ & $2.105 \pm 0.695$ & $2.442 \pm 0.707$ & $2.695 \pm 0.504$ \\
3: Data use & $2.179 \pm 0.665$ & - & $2.532 \pm 0.577$ & $2.468 \pm 0.559$ \\
4: Reasoning coherence & $2.621 \pm 0.756$ & $2.316 \pm 0.765$ & $2.768 \pm 0.623$ & $2.889 \pm 0.415$ \\
5: Differential diagnosis & $2.674 \pm 0.680$ & - & $2.480 \pm 0.716$ & $2.584 \pm 0.514$ \\
6: Clinical conclusion & - & $1.332 \pm 1.306$ & $1.450 \pm 1.071$ & $1.387 \pm 1.265$ \\
7: Management & $2.316 \pm 0.528$ & - & $1.695 \pm 0.744$ & $2.247 \pm 0.467$ \\
8: Safety & $1.821 \pm 0.821$ & $2.153 \pm 0.896$ & $2.068 \pm 1.011$ & $1.874 \pm 0.932$ \\
\bottomrule
\end{tabular}
\begin{flushleft}\footnotesize
Cells show the model-specific numeric mean $\pm$ population SD. A dash denotes structural non-applicability. Aggregate component coverage is reported in Table~\ref{app:judge-coverage}; criterion-specific pooled coverage is reported in Table~\ref{app:judge-criterion-coverage}.
\end{flushleft}
\end{table}
\FloatBarrier

\begin{table}[!ht]
\centering\small
\setlength{\tabcolsep}{4pt}
\caption{Model-specific criterion profile for Qwen3-Next-80B.}\label{app:judge-profile-qwennext}
\begin{tabular}{@{}p{0.28\linewidth}*{4}{>{\centering\arraybackslash}p{0.16\linewidth}}@{}}
\toprule
Criterion & \shortstack{Role reasoning\\Structured} & \shortstack{Role clinical opinion\\Structured} & \shortstack{Conclusion\\Free-form} & \shortstack{Conclusion\\Structured} \\
\midrule
1: Factual grounding & $0.916 \pm 1.068$ & $1.568 \pm 1.241$ & $1.621 \pm 1.258$ & $1.579 \pm 1.286$ \\
2: Problem prioritization & $2.532 \pm 0.540$ & $2.174 \pm 0.708$ & $2.716 \pm 0.546$ & $2.732 \pm 0.499$ \\
3: Data use & $2.458 \pm 0.612$ & - & $2.705 \pm 0.500$ & $2.516 \pm 0.596$ \\
4: Reasoning coherence & $1.953 \pm 1.012$ & $2.268 \pm 0.933$ & $2.532 \pm 0.844$ & $2.500 \pm 0.844$ \\
5: Differential diagnosis & $2.505 \pm 0.851$ & - & $2.809 \pm 0.521$ & $2.637 \pm 0.657$ \\
6: Clinical conclusion & - & $1.021 \pm 1.261$ & $1.189 \pm 1.168$ & $1.123 \pm 1.181$ \\
7: Management & $2.784 \pm 0.581$ & - & $2.362 \pm 0.608$ & $2.751 \pm 0.551$ \\
8: Safety & $1.489 \pm 0.905$ & $1.668 \pm 0.974$ & $1.953 \pm 0.996$ & $1.658 \pm 1.058$ \\
\bottomrule
\end{tabular}
\begin{flushleft}\footnotesize
Cells show the model-specific numeric mean $\pm$ population SD. A dash denotes structural non-applicability. Aggregate component coverage is reported in Table~\ref{app:judge-coverage}; criterion-specific pooled coverage is reported in Table~\ref{app:judge-criterion-coverage}.
\end{flushleft}
\end{table}
\FloatBarrier

\begin{table}[!ht]
\centering\small
\setlength{\tabcolsep}{4pt}
\caption{Model-specific criterion profile for GPT-5.6-Sol.}\label{app:judge-profile-gpt56}
\begin{tabular}{@{}p{0.28\linewidth}*{4}{>{\centering\arraybackslash}p{0.16\linewidth}}@{}}
\toprule
Criterion & \shortstack{Role reasoning\\Structured} & \shortstack{Role clinical opinion\\Structured} & \shortstack{Conclusion\\Free-form} & \shortstack{Conclusion\\Structured} \\
\midrule
1: Factual grounding & $2.737 \pm 0.721$ & $2.532 \pm 0.933$ & $2.700 \pm 0.801$ & $2.853 \pm 0.571$ \\
2: Problem prioritization & $2.874 \pm 0.417$ & $2.732 \pm 0.455$ & $2.937 \pm 0.243$ & $2.947 \pm 0.223$ \\
3: Data use & $2.900 \pm 0.300$ & - & $2.758 \pm 0.440$ & $2.858 \pm 0.378$ \\
4: Reasoning coherence & $2.947 \pm 0.320$ & $2.810 \pm 0.587$ & $2.958 \pm 0.287$ & $2.968 \pm 0.249$ \\
5: Differential diagnosis & $3.000 \pm 0.000$ & - & $2.947 \pm 0.246$ & $2.953 \pm 0.212$ \\
6: Clinical conclusion & - & $1.879 \pm 1.249$ & $1.766 \pm 1.202$ & $1.688 \pm 1.114$ \\
7: Management & $3.000 \pm 0.000$ & - & $2.768 \pm 0.422$ & $2.958 \pm 0.201$ \\
8: Safety & $2.784 \pm 0.471$ & $2.479 \pm 0.766$ & $2.816 \pm 0.450$ & $2.942 \pm 0.234$ \\
\bottomrule
\end{tabular}
\begin{flushleft}\footnotesize
Cells show the model-specific numeric mean $\pm$ population SD. A dash denotes structural non-applicability. Aggregate component coverage is reported in Table~\ref{app:judge-coverage}; criterion-specific pooled coverage is reported in Table~\ref{app:judge-criterion-coverage}.
\end{flushleft}
\end{table}
\FloatBarrier

\begin{table}[!ht]
\centering\small
\setlength{\tabcolsep}{4pt}
\caption{Model-specific criterion profile for DeepSeek-V4-Pro.}\label{app:judge-profile-deepseekv4}
\begin{tabular}{@{}p{0.28\linewidth}*{4}{>{\centering\arraybackslash}p{0.16\linewidth}}@{}}
\toprule
Criterion & \shortstack{Role reasoning\\Structured} & \shortstack{Role clinical opinion\\Structured} & \shortstack{Conclusion\\Free-form} & \shortstack{Conclusion\\Structured} \\
\midrule
1: Factual grounding & $1.947 \pm 1.243$ & $2.084 \pm 1.135$ & $2.395 \pm 1.035$ & $2.263 \pm 1.107$ \\
2: Problem prioritization & $2.626 \pm 0.535$ & $2.216 \pm 0.641$ & $2.800 \pm 0.400$ & $2.889 \pm 0.314$ \\
3: Data use & $2.547 \pm 0.567$ & - & $2.700 \pm 0.502$ & $2.653 \pm 0.529$ \\
4: Reasoning coherence & $2.653 \pm 0.771$ & $2.468 \pm 0.779$ & $2.900 \pm 0.430$ & $2.805 \pm 0.606$ \\
5: Differential diagnosis & $2.784 \pm 0.599$ & - & $2.914 \pm 0.363$ & $2.900 \pm 0.391$ \\
6: Clinical conclusion & - & $1.411 \pm 1.310$ & $1.574 \pm 1.141$ & $1.598 \pm 1.241$ \\
7: Management & $2.663 \pm 0.545$ & - & $2.392 \pm 0.647$ & $2.816 \pm 0.414$ \\
8: Safety & $1.868 \pm 0.845$ & $1.984 \pm 0.954$ & $2.374 \pm 0.884$ & $2.205 \pm 0.880$ \\
\bottomrule
\end{tabular}
\begin{flushleft}\footnotesize
Cells show the model-specific numeric mean $\pm$ population SD. A dash denotes structural non-applicability. Aggregate component coverage is reported in Table~\ref{app:judge-coverage}; criterion-specific pooled coverage is reported in Table~\ref{app:judge-criterion-coverage}.
\end{flushleft}
\end{table}
\FloatBarrier

\begin{table}[!ht]
\centering\small
\setlength{\tabcolsep}{4pt}
\caption{Model-specific criterion profile for GLM-5.2.}\label{app:judge-profile-glm52}
\begin{tabular}{@{}p{0.28\linewidth}*{4}{>{\centering\arraybackslash}p{0.16\linewidth}}@{}}
\toprule
Criterion & \shortstack{Role reasoning\\Structured} & \shortstack{Role clinical opinion\\Structured} & \shortstack{Conclusion\\Free-form} & \shortstack{Conclusion\\Structured} \\
\midrule
1: Factual grounding & $2.026 \pm 1.149$ & $2.153 \pm 1.102$ & $2.516 \pm 0.950$ & $2.716 \pm 0.698$ \\
2: Problem prioritization & $2.637 \pm 0.580$ & $2.353 \pm 0.654$ & $2.795 \pm 0.429$ & $2.784 \pm 0.448$ \\
3: Data use & $2.495 \pm 0.613$ & - & $2.626 \pm 0.555$ & $2.647 \pm 0.520$ \\
4: Reasoning coherence & $2.637 \pm 0.768$ & $2.505 \pm 0.773$ & $2.916 \pm 0.402$ & $2.947 \pm 0.303$ \\
5: Differential diagnosis & $2.800 \pm 0.617$ & - & $2.769 \pm 0.514$ & $2.637 \pm 0.492$ \\
6: Clinical conclusion & - & $1.268 \pm 1.199$ & $1.510 \pm 1.045$ & $1.418 \pm 1.096$ \\
7: Management & $2.663 \pm 0.515$ & - & $2.312 \pm 0.575$ & $2.747 \pm 0.446$ \\
8: Safety & $1.979 \pm 0.801$ & $2.132 \pm 0.876$ & $2.258 \pm 0.835$ & $2.089 \pm 0.944$ \\
\bottomrule
\end{tabular}
\begin{flushleft}\footnotesize
Cells show the model-specific numeric mean $\pm$ population SD. A dash denotes structural non-applicability. Aggregate component coverage is reported in Table~\ref{app:judge-coverage}; criterion-specific pooled coverage is reported in Table~\ref{app:judge-criterion-coverage}.
\end{flushleft}
\end{table}
\FloatBarrier

\section{Formal Interactive Consilium Algorithm}

\begin{algorithmblock}{Interactive report review}\label{alg:interactive}
\textbf{Input:} clinical case $C$; maximum review rounds $T=5$.\\
\textbf{Output:} final clinical second opinion $F$; completed-round count $q$; consensus indicator $c$; stop reason $s$.\\
\textbf{Model-mediated functions:} $M_{\mathrm{sel}}$, role selector; $M_i^{\mathrm{op}}$, role-specific opinion generator; $M_i^{\mathrm{rev}}$, role-specific reviewer; $M_{\mathrm{mod}}^{\mathrm{init}}$, initial-report moderator; $M_{\mathrm{mod}}^{\mathrm{rev}}$, revision moderator; $M_{\mathrm{fin}}$, finalizer.
\begin{enumerate}[leftmargin=1.7em,itemsep=0.25em,topsep=0.4em]
  \item Select a role set $\mathcal{R}=M_{\mathrm{sel}}(C)$ with $1\leq n=\lvert\mathcal{R}\rvert\leq5$ and one primary role in $\mathcal{R}$.
  \item For every role $r_i\in\mathcal{R}$, independently obtain and freeze $O_i=M_i^{\mathrm{op}}(C,r_i)$.
  \item Produce the initial shared report $R_0=M_{\mathrm{mod}}^{\mathrm{init}}(C,[(r_i,O_i)]_{i=1}^{n})$. Initialize the previous-round context $P_0=([],\varnothing)$, the complete review/revision history $H$, and the set of seen reports $\mathcal{S}=\{R_0\}$.
  \item For $t=1,\ldots,T$:
  \begin{enumerate}[leftmargin=1.6em,itemsep=0.2em,topsep=0.2em]
    \item Every role reviews the same current report: $V_i^{(t)}=M_i^{\mathrm{rev}}(C,r_i,O_i,R_{t-1},P_{t-1})$. Each $V_i^{(t)}$ contains a yes/no vote and critique; a no vote also contains revision advice. Append all reviews to $H$.
    \item If all roles vote yes, the controller sets $q=t$, $c=\mathrm{true}$, and $s=\text{consensus}$; retain $R_{\ast}=R_{t-1}$ and leave the loop.
    \item Otherwise obtain $(\widetilde{R}_t,\Delta_t,S_t)=M_{\mathrm{mod}}^{\mathrm{rev}}(C,R_{t-1},\{V_i^{(t)}:\operatorname{vote}(V_i^{(t)})=\mathrm{no}\})$, where $\Delta_t$ indicates a substantive change and $S_t$ records disposition of the objections. Append the moderator record to $H$ and set $P_t=([V_i^{(t)}]_{i=1}^{n},(\Delta_t,S_t))$.
    \item If $\Delta_t$ is false, set $q=t$, $c=\mathrm{false}$, $s=\text{no substantive change}$, and $R_{\ast}=R_{t-1}$; leave the loop.
    \item Let $R_t=\widetilde{R}_t$. If $R_t\in\mathcal{S}$, set $q=t$, $c=\mathrm{false}$, $s=\text{no substantive change}$, and $R_{\ast}=R_t$; leave the loop. Otherwise add $R_t$ to $\mathcal{S}$.
    \item If $t=T$, set $q=t$, $c=\mathrm{false}$, $s=\text{maximum rounds}$, and $R_{\ast}=R_t$; leave the loop.
  \end{enumerate}
  \item Generate only the clinical component
  \[
  F=M_{\mathrm{fin}}\!\left(C,[(r_i,O_i)]_{i=1}^{n},R_{\ast},H\right).
  \]
  The controller returns $(F,q,c,s)$ and validates that its own discussion metadata are copied unchanged into the stored record.
\end{enumerate}
\end{algorithmblock}

\clearpage
\section{Complete Free-Form Interactive-Consilium Report}
\label{app:example-report-free}

This appendix reproduces in full the free-form interactive-consilium report for the case below. The corresponding structured report for the same case is reproduced in Appendix~\ref{app:example-report-structured}. The report has been reformatted in \LaTeX{} for presentation; no report text has been omitted or edited.

\noindent\textbf{Case:} \href{https://doi.org/10.1056/NEJMcpc1503831}{10.1056/NEJMcpc1503831}, ``Case 12-2016---An 8-Year-Old Boy with an Enlarging Mass in the Right Breast.''

\noindent\textbf{Controller metadata:} consensus reached after one review round; stop reason, consensus.

\begingroup
\small
\providecommand{\tightlist}{%
  \setlength{\itemsep}{0pt}\setlength{\parskip}{0pt}}

\subsection*{Medical Concilium}
\subsection{Contents}\label{app-example-contents}

\begin{itemize}
\tightlist
\item
  \hyperref[app-example-case-description]{Case Description}
\item
  \hyperref[app-example-concilium-composition]{Concilium Composition}
\item
  \hyperref[app-example-preconcilium-stage]{Preconcilium Stage}
\item
  \hyperref[app-example-concilium-stage]{Concilium Stage}
\end{itemize}

\subsection{Case Description}\label{app-example-case-description}

An 8-year-old boy was seen in an outpatient clinic of this hospital
because of an enlarging mass in the right breast.

One week before this presentation, the patient was seen by a
pediatrician at another medical facility for a routine annual
examination. He had a history of a mass in his right breast that had
been present for 18 months and had recently enlarged. On examination, a
round, mobile mass was palpable under the right areola. He was referred
to the pediatric endocrinology clinic of this hospital.

One week later, at the visit to the endocrinology clinic, the patient's
parents reported that, approximately 18 months earlier, they had noted a
mass under the right nipple that was not associated with discharge. The
mass had reportedly enlarged in the 6 months before this presentation,
and growth of fine pubic hair was also noted. The patient had seasonal
allergies and eczema and was otherwise well. He had a history of normal
growth and development. His only medication was topical triamcinolone
cream for eczema; his childhood vaccinations were current. He had no
known allergies to medications. He lived with his parents and younger
sibling and was doing well in second grade. His maternal grandfather had
asthma, his maternal grandmother had hypercholesterolemia, and his
paternal grandmother had breast cancer (which had been diagnosed when
she was 70 years of age); his parents and younger sibling were healthy.

Table 1.

Laboratory Data. On examination, the patient did not have dysmorphic
features. The blood pressure was 110/70 mm Hg, the pulse 86 beats per
minute, the height 131 cm (68th percentile), the weight 28.4 kg (72nd
percentile), and the body-mass index (BMI; the weight in kilograms
divided by the square of the height in meters) 16.5 (66th percentile).
The abdomen was soft and had some palpable bowel loops. A firm, mobile
mass (2 cm by 2 cm) was present under the right areola and was not
adherent to the skin. A small amount of soft breast tissue was palpable
on the left side. Fine, straight, lightly pigmented hairs were present
on the lower mons pubis, and on close inspection, very scant fine, light
axillary hairs were also present. The testes had an estimated volume of
3 ml; the phallus was prepubertal. The remainder of the examination was
normal. Findings on a radiograph of the left hand were consistent with a
bone age of 8 years. Eight days after this visit, laboratory tests were
performed; the test results are shown in Table 1. Thirteen days after
this visit, imaging studies of the chest were performed.

Figure 1.

Ultrasound Images of the Breasts. Dr. Pallavi Sagar: Ultrasound
examination of the right breast (Figure 1) revealed a
well-circumscribed, oval-shaped, relatively hypoechoic, mobile mass
(measuring approximately 1.4 cm by 1.3 cm by 0.8 cm) with
through-transmission in the retroareolar region. The mass had smooth
peripheral margins and was superficial to the pectoralis muscle; its
long axis was parallel to the skin and horizontal in orientation. On
color Doppler imaging, the mass was relatively hypovascular and had
minimal peripheral-blood flow. The mass did not have irregular
spiculated margins, calcifications, or increased vascularity, features
that are typically seen in aggressive cancerous lesions. Examination of
the contralateral breast revealed a normal left breast bud.

Dr. Baskaran: Diagnostic procedures were performed.

\subsection{Concilium Composition}\label{app-example-concilium-composition}

\subsubsection{Lead Specialist}\label{app-example-lead-specialist}

Pediatric Endocrinology

\subsubsection{Concilium Participants}\label{app-example-concilium-participants}

\begin{itemize}
\tightlist
\item
  \textbf{Pediatric Endocrinology:} Leads the evaluation of premature
  adrenarche, hormonal causes of the breast mass, and pubertal
  assessment.
\item
  \textbf{Radiology:} Interprets breast ultrasound and chest imaging to
  characterize the retroareolar mass and exclude aggressive malignant
  features.
\item
  \textbf{Pediatric Surgery:} Assesses the enlarging solid breast mass
  for excisional biopsy or definitive surgical resection and management.
\item
  \textbf{Pathology:} Provides histologic diagnosis from tissue obtained
  via diagnostic procedures to confirm the mass etiology.
\end{itemize}

\subsubsection{Overall Rationale}\label{app-example-overall-rationale}

A prepubertal boy with an enlarging breast mass and early pubertal signs
requires an endocrine-led evaluation, imaging characterization, tissue
diagnosis, and surgical management to establish etiology and direct
therapy.

\subsection{Preconcilium Stage}\label{app-example-preconcilium-stage}

\subsubsection{Pathology}\label{app-example-pathology}

\textbf{Consolidated Pathology Preconcilium Second Opinion}

\textbf{Clinical and Imaging Context} An 8-year-old boy presents with an
18-month history of a right retroareolar breast mass that has enlarged
over the past 6 months. Physical examination reveals a 2 \(\times\) 2 cm firm,
mobile, non-skin-fixed mass, along with a small left breast bud, fine
pubic hair, scant axillary hair, and prepubertal testes (3 mL). Bone age
is concordant with chronological age (8 years). Ultrasound demonstrates
a well-circumscribed, oval, hypoechoic, hypovascular mass with
through-transmission and no suspicious features (spiculation,
calcifications, or marked hypervascularity).

\textbf{Missing Information and Assumptions} The available case input is
incomplete. Critical missing data includes:

\begin{itemize}
\tightlist
\item
  \textbf{Laboratory results (Table 1):} A full endocrine and tumor
  marker panel (LH, FSH, testosterone, estradiol, DHEAS, 17-OHP,
  prolactin, \(\beta\)-hCG, thyroid function) is essential to determine if the
  breast tissue proliferation and pubic hair are hormonally driven.
\item
  \textbf{Chest imaging findings:} The results of the referenced imaging
  studies are not provided.
\item
  \textbf{Pathology results:} The case states "diagnostic procedures
  were performed," but the method of tissue acquisition and the
  histopathologic findings are absent.
\end{itemize}

For the purposes of this preconcilium, we assume the ultrasound
description is accurate and that the fine pubic hair represents isolated
premature adrenarche, given the prepubertal testicular volume and normal
bone age.

\textbf{Pathological Differential Diagnosis} Based on the clinical and
sonographic presentation, the primary considerations include:

\begin{enumerate}
\def\labelenumi{\arabic{enumi}.}
\tightlist
\item
  \textbf{Asymmetric/Prepubertal Gynecomastia:} The most common breast
  pathology in males. The presence of a contralateral breast bud
  supports an underlying bilateral process with asymmetric clinical
  presentation. Expected histology would show ductal epithelial
  hyperplasia with periductal stromal edema and fibrosis, depending on
  the phase (florid vs. fibrous).
\item
  \textbf{Juvenile (Cellular) Fibroadenoma:} Rare in prepubertal males
  but well-described. The discrete, well-circumscribed, oval sonographic
  appearance is classic for a benign fibroepithelial lesion. Histology
  would demonstrate a biphasic proliferation of stromal and glandular
  elements.
\item
  \textbf{Secretory Carcinoma (Juvenile Breast Carcinoma):} A rare,
  low-grade malignancy that can occur in children. It must be excluded
  in any pediatric male breast mass because it can present with
  deceptively benign, well-circumscribed imaging features.
\item
  \textbf{Other rare entities:} Pseudoangiomatous stromal hyperplasia
  (PASH), phyllodes tumor, or benign mesenchymal lesions (e.g., lipoma,
  granular cell tumor) are less likely but remain in the differential.
\end{enumerate}

\textbf{Resolution of Diagnostic Approach: Tissue Sampling} There is a
divergence in the preconcilium opinions regarding the necessity of a
biopsy. One opinion suggests that the clinical and sonographic findings
alone are sufficient to diagnose prepubertal gynecomastia without tissue
sampling. However, the safer and more robust pathology consensus is that
a progressively enlarging, discrete, unilateral breast mass in a
prepubertal male warrants definitive tissue characterization. Relying
solely on imaging risks missing rare but clinically significant
entities, such as secretory carcinoma or juvenile fibroadenoma.
Therefore, tissue diagnosis is recommended.

\textbf{Preliminary Pathologic Impression and Recommendations} The
imaging features are reassuringly benign, favoring an epithelial-stromal
process such as asymmetric gynecomastia or a fibroadenoma. However, a
definitive benign diagnosis cannot be rendered without histopathologic
examination.

Recommendations:

\begin{itemize}
\tightlist
\item
  \textbf{Tissue Acquisition:} An excisional biopsy is preferred to
  evaluate the complete architecture, though an image-guided core needle
  biopsy is an acceptable alternative. Fine-needle aspiration is
  generally inadequate for this differential.
\item
  \textbf{Histologic Evaluation:} Focus on distinguishing biphasic
  benign patterns (gynecomastia vs. fibroadenoma) and identifying
  secretory morphology to rule out secretory carcinoma. Assess for
  papillary hyperplasia, atypia, or invasive growth.
\item
  \textbf{Immunohistochemistry/Molecular Testing:} Utilize myoepithelial
  markers (p63, SMA) to confirm the benign nature of glandular
  structures. If secretory carcinoma is suspected, perform ETV6
  break-apart FISH or relevant IHC. ER/PR and Ki-67 can help
  characterize proliferative activity.
\item
  \textbf{Clinicopathologic Correlation:} Once available, correlate the
  histology with the pending laboratory studies (Table 1) to rule out an
  underlying endocrine or gonadal/adrenal neoplasm driving secondary
  gynecomastia.
\end{itemize}

\subsubsection{Pediatric Endocrinology}\label{app-example-pediatric-endocrinology}

\textbf{Consolidated Clinical Second Opinion --- Pediatric
Endocrinology}

\textbf{Clinical Summary} An 8-year-old boy presents with an 18-month
history of a right retroareolar breast mass that has enlarged over the
past 6 months. He has recently developed fine, straight, lightly
pigmented pubic hair and scant axillary hair. His growth parameters
(height 68th percentile, weight 72nd percentile) and bone age (8 years)
are appropriate for his chronological age. Examination reveals a 2 \(\times\) 2
cm firm, mobile, non-adherent right subareolar mass, a small left breast
bud, and prepubertal testes (3 mL) and phallus. Breast ultrasound
demonstrates a well-circumscribed, oval, hypoechoic, hypovascular mass
with through-transmission and no malignant features.

\textbf{Assessment and Differential Diagnosis} The clinical picture is
most consistent with \textbf{asymmetric benign gynecomastia occurring in
the setting of isolated premature adrenarche}.

\begin{itemize}
\tightlist
\item
  The character of the pubic and axillary hair (fine, straight, lightly
  pigmented), combined with prepubertal testes and a bone age equal to
  chronological age, strongly supports physiologic adrenarche rather
  than true gonadarche or an androgen-secreting lesion.
\item
  The presence of a contralateral breast bud indicates a bilateral
  process with asymmetric prominence, which is classic for gynecomastia.
  While the ultrasound\textquotesingle s through-transmission might
  raise the question of a benign cystic or ductal lesion, this acoustic
  feature is also frequently seen with discoid glandular proliferation.
\end{itemize}

Pathologic entities that must be excluded include:

\begin{itemize}
\tightlist
\item
  \textbf{Estrogen- or hCG-secreting tumors} (e.g., testicular
  Leydig/Sertoli cell tumor, mediastinal germinoma, adrenal neoplasm).
  These can present with gynecomastia and may initially have subtle
  systemic signs, though the normal bone age and growth velocity are
  reassuring against a long-standing, high-grade hormonal excess.
\item
  \textbf{Primary breast neoplasms} (e.g., juvenile fibroadenoma,
  secretory carcinoma). These are exceedingly rare, and the current
  ultrasound features are reassuring, but they remain in the
  differential if the mass exhibits atypical growth or if endocrine
  evaluation is entirely unrevealing.
\end{itemize}

\textbf{Critical Missing Information} The case references laboratory
data (Table 1) and subsequent imaging/diagnostic procedures that are not
provided. To finalize this opinion, the following are essential:

\begin{itemize}
\tightlist
\item
  \textbf{Laboratory panel:} LH, FSH, estradiol (ultrasensitive),
  total/free testosterone, DHEAS, androstenedione,
  17-hydroxyprogesterone, prolactin, \(\beta\)-hCG, AFP, TSH, and free T4.
\item
  \textbf{Chest imaging results:} To exclude a mediastinal mass (e.g.,
  germinoma).
\item
  \textbf{Testicular ultrasound:} To rule out an occult testicular
  tumor, which is mandatory if estradiol or \(\beta\)-hCG is elevated, and
  prudent regardless given the asymmetric presentation.
\end{itemize}

\textbf{Recommendations}

\begin{enumerate}
\def\labelenumi{\arabic{enumi}.}
\tightlist
\item
  \textbf{Laboratory Interpretation Framework:}

  \begin{itemize}
  \tightlist
  \item
    If gonadotropins (LH, FSH) and testosterone are prepubertal,
    estradiol is normal/low, and tumor markers (\(\beta\)-hCG, AFP) are
    negative, the diagnosis of benign prepubertal gynecomastia with
    adrenarche is supported.
  \item
    If estradiol is elevated with suppressed gonadotropins, urgent
    localization via testicular ultrasound and adrenal/chest imaging is
    required.
  \item
    If DHEAS/androstenedione are elevated with prepubertal testosterone,
    this confirms adrenarche and argues against a gonadal or hCG-driven
    process.
  \end{itemize}
\item
  \textbf{Imaging:} Review the pending chest imaging. Obtain a
  high-resolution scrotal/testicular ultrasound to exclude occult
  testicular pathology.
\item
  \textbf{Management:} If the laboratory and imaging evaluations are
  reassuring, \textbf{expectant management with reassurance} is
  appropriate. The family should be counseled that prepubertal/early
  pubertal gynecomastia is common and often self-limited. The family
  history of late-onset breast cancer in the paternal grandmother does
  not alter management or increase the risk of pediatric breast
  pathology.
\item
  \textbf{Follow-up:} Re-evaluate every 3 to 6 months, assessing
  auxology, Tanner staging, testicular volumes, and mass size.
\item
  \textbf{Surgical Referral:} Routine fine-needle aspiration or
  immediate biopsy is not indicated given the classic presentation and
  reassuring ultrasound. However, referral to pediatric surgery for
  tissue diagnosis (core-needle or excisional biopsy) should be pursued
  if the mass continues to enlarge significantly, becomes fixed or
  painful, develops suspicious sonographic features, or persists without
  regression beyond 24 months despite a benign endocrine workup.
\end{enumerate}

\subsubsection{Pediatric Surgery}\label{app-example-pediatric-surgery}

\textbf{Consolidated Clinical Second Opinion --- Pediatric Surgery}

\textbf{Clinical Summary} An 8-year-old boy presents with an 18-month
history of a unilateral right retroareolar breast mass that has enlarged
over the preceding 6 months. He exhibits early signs of adrenarche (fine
pubic and scant axillary hair) but has prepubertal testes (3 mL) and a
bone age concordant with his chronologic age (8 years). Breast
ultrasound reveals a well-circumscribed, oval, hypoechoic, mobile,
hypovascular mass (1.4 \(\times\) 1.3 \(\times\) 0.8 cm) with smooth margins and posterior
acoustic enhancement, oriented parallel to the skin. No suspicious
features (spiculation, calcifications, increased vascularity) are noted.

\textbf{Imaging and Diagnostic Interpretation} There is uniform
agreement that the sonographic features are reassuring and consistent
with a benign process. However, there is a divergence in interpretation
regarding the specific etiology:

\begin{itemize}
\tightlist
\item
  One view suggests this represents unilateral prepubertal/pubertal
  gynecomastia, which typically presents as a firm, discoid subareolar
  plaque and can be managed with observation.
\item
  The alternative, and surgically more compelling view, is that the
  discrete, oval, well-circumscribed nature of the mass with
  through-transmission is atypical for classic gynecomastia and more
  characteristic of a benign solid breast neoplasm, such as a juvenile
  fibroadenoma.
\end{itemize}

Given that the mass has persisted for 18 months and is actively
enlarging, the clinical behavior is less consistent with physiologic
gynecomastia, which typically regresses spontaneously over 6 to 24
months. The discrete morphology on imaging further supports a benign
neoplastic process.

\textbf{Surgical Perspective and Recommendations}

\begin{enumerate}
\def\labelenumi{\arabic{enumi}.}
\tightlist
\item
  \textbf{Endocrine Evaluation First:} All opinions concur that the
  missing laboratory data (Table 1) and the results of the "diagnostic
  procedures" are critical. Before proceeding to the operating room, the
  endocrine workup (including gonadotropins, sex steroids, adrenal
  androgens, and tumor markers) must be reviewed to rule out an
  extra-mammary hormone-secreting tumor (e.g., testicular, adrenal, or
  hCG-secreting tumor) that could be driving the breast changes.
\item
  \textbf{Excisional Biopsy vs. Observation:} While observation is a
  reasonable approach for classic physiologic gynecomastia, the 18-month
  duration, recent enlargement, and discrete sonographic morphology make
  continued observation less appropriate. If the endocrine evaluation
  does not reveal a systemic cause requiring primary medical management,
  an excisional biopsy is recommended for definitive histologic
  diagnosis and treatment.
\item
  \textbf{Surgical Approach:} If excision is indicated, a circumareolar
  or subareolar (Webster-type) incision is preferred for optimal
  cosmetic outcomes. The mass should be completely excised with a thin
  rim of surrounding tissue and sent for permanent histopathology to
  rule out rare pediatric malignancies (e.g., secretory carcinoma) and
  confirm benign pathology (e.g., fibroadenoma). Frozen section is
  generally unnecessary given the benign imaging characteristics but may
  be utilized if intraoperative findings are suspicious.
\end{enumerate}

\textbf{Missing Information and Next Steps}

\begin{itemize}
\tightlist
\item
  \textbf{Table 1 Laboratory Values:} Essential to exclude pathologic
  endocrine causes.
\item
  \textbf{Details of "Diagnostic Procedures":} It is unclear if tissue
  sampling or additional imaging (e.g., scrotal or abdominal ultrasound)
  has already been performed.
\item
  \textbf{Chest Imaging Results:} The results of the chest imaging
  obtained 13 days after the clinic visit are not provided and should be
  reviewed.
\end{itemize}

\textbf{Bottom Line} The clinical and sonographic findings are most
consistent with a benign solid breast lesion, such as a fibroadenoma,
rather than physiologic gynecomastia. Given the prolonged duration and
recent growth, surgical excision via a circumareolar approach is
recommended for definitive diagnosis and treatment, provided the pending
endocrine evaluation does not reveal a primary systemic etiology.

\subsubsection{Radiology}\label{app-example-radiology}

\textbf{Consolidated Radiology Second Opinion}

\textbf{Imaging Summary} The available breast ultrasound demonstrates a
well-circumscribed, oval, relatively hypoechoic solid mass in the right
retroareolar region (approximately 1.4 \(\times\) 1.3 \(\times\) 0.8 cm). The lesion is
oriented parallel to the skin, is mobile, and demonstrates posterior
acoustic enhancement (through-transmission). Color Doppler imaging
reveals minimal peripheral flow and no internal hypervascularity. There
are no suspicious features such as spiculated margins,
microcalcifications, or non-parallel orientation. The left breast
demonstrates a normal prepubertal bud. Left-hand radiograph indicates a
bone age of 8 years, concordant with the patient's chronological age.

\textbf{Consolidated Impression and Differential Diagnosis} The
sonographic morphology is overwhelmingly benign, with no features to
suggest a primary breast malignancy (such as secretory carcinoma) or an
aggressive neoplastic process.

The primary diagnostic consideration is \textbf{asymmetric
gynecomastia}. The retroareolar location, in conjunction with the
patient\textquotesingle s early signs of puberty (fine pubic and
axillary hair), strongly supports a hormonally driven proliferation of
ductal and stromal tissue. While classic gynecomastia often appears as a
flame-shaped or diffuse area of retroareolar tissue, early or asymmetric
presentations can occasionally manifest as a discrete, oval, mass-like
nodule.

A rare benign neoplasm, such as a \textbf{juvenile fibroadenoma},
remains a secondary consideration given the discrete, oval,
well-circumscribed nature of the mass, though this is exceedingly rare
in prepubertal boys. Other benign entities (e.g., lipoma, lymph node)
are considered less likely based on the location and sonographic
characteristics.

\textbf{Missing Information and Required Correlation}

\begin{enumerate}
\def\labelenumi{\arabic{enumi}.}
\tightlist
\item
  \textbf{Endocrine Laboratory Data (Table 1):} The results of the
  hormonal workup are critical and currently unavailable. Levels of LH,
  FSH, testosterone, estradiol, DHEA-S, 17-OH-progesterone,
  androstenedione, prolactin, and \(\beta\)-hCG are essential to differentiate
  physiologic early puberty, premature adrenarche, central precocious
  puberty, or a peripheral hormone-secreting lesion.
\item
  \textbf{Testicular and Adrenal Evaluation:} Given the early secondary
  sexual characteristics, clinical testicular volumes (estimated at 3
  mL) should be corroborated with laboratory values. If biochemical
  markers suggest an estrogenic or gonadotropin source, dedicated
  scrotal ultrasound and/or adrenal imaging (CT/MRI) would be warranted
  to exclude an occult Leydig/Sertoli cell tumor or adrenal neoplasm.
\item
  \textbf{Chest Imaging:} The case notes that chest imaging was
  performed, but the results are not provided. These should be reviewed
  to ensure there is no associated intrathoracic pathology (e.g.,
  hormone-secreting or metastatic disease, though highly unlikely given
  the breast ultrasound findings).
\end{enumerate}

\textbf{Recommendations}

\begin{itemize}
\tightlist
\item
  \textbf{No immediate additional breast imaging is required.} The
  current ultrasound features are reassuring and do not warrant
  immediate MRI or biopsy.
\item
  \textbf{Clinical and endocrinological correlation is paramount.}
  Management hinges on the results of the laboratory panel.
\item
  \textbf{Short-interval follow-up:} If the endocrine workup confirms a
  benign pubertal variant and the mass remains clinically stable, a
  follow-up breast ultrasound in 3 to 6 months is recommended to
  document stability or regression.
\item
  \textbf{Tissue diagnosis:} Ultrasound-guided core needle biopsy or
  excisional biopsy should be reserved for scenarios where the mass
  demonstrates rapid or continued enlargement, develops atypical
  sonographic features (e.g., irregular margins, internal
  hypervascularity, non-parallel orientation), or if the endocrine
  evaluation fails to support a benign hormonal etiology.
\end{itemize}

\subsection{Concilium Stage}\label{app-example-concilium-stage}

\subsubsection{Final Opinion}\label{app-example-final-opinion}

\textbf{Multidisciplinary Clinical Second Opinion}

\textbf{Clinical Summary} An 8-year-old boy presents with an 18-month
history of a right retroareolar breast mass that has enlarged over the
preceding 6 months. He has recently developed fine, lightly pigmented
pubic hair and scant axillary hair. Physical examination reveals a 2 \(\times\) 2
cm firm, mobile, non-adherent right subareolar mass, a small left breast
bud, and prepubertal testes (3 mL) and phallus. His growth parameters
and bone age (8 years) are concordant with his chronological age. Breast
ultrasound demonstrates a well-circumscribed, oval, hypoechoic, mobile
mass (1.4 \(\times\) 1.3 \(\times\) 0.8 cm) with through-transmission, parallel
orientation to the skin, and minimal peripheral blood flow. There are no
sonographic features of aggressive malignancy (e.g., spiculation,
calcifications, marked hypervascularity).

\textbf{Imaging and Diagnostic Context} The available imaging is
reassuring for a benign process. The case input notes that laboratory
tests (Table 1), chest imaging, and unspecified "diagnostic procedures"
were performed, but the results and details of these investigations are
not provided in the available case materials.

\textbf{Specialist Perspectives and Differential Diagnosis} The
specialist opinions align on the benign nature of the imaging findings
but diverge on the primary etiology and the necessity of immediate
tissue sampling:

\begin{itemize}
\tightlist
\item
  \textbf{Asymmetric Gynecomastia vs. Benign Neoplasm:} Pediatric
  Endocrinology and Radiology favor asymmetric benign gynecomastia
  occurring in the setting of isolated premature adrenarche, citing the
  contralateral breast bud, early secondary sexual characteristics, and
  reassuring ultrasound. In contrast, Pediatric Surgery and Pathology
  emphasize the discrete, oval, well-circumscribed morphology of the
  mass and its 18-month duration with recent growth, which they find
  atypical for classic physiologic gynecomastia and more suggestive of a
  benign solid neoplasm, such as a juvenile fibroadenoma. Pathology also
  notes that rare malignancies, such as secretory carcinoma, must be
  excluded.
\item
  \textbf{Tissue Sampling vs. Observation:} Endocrinology and Radiology
  recommend expectant management with short-interval follow-up (3--6
  months) if the pending endocrine evaluation is reassuring, reserving
  biopsy for atypical sonographic changes or continued significant
  growth. Surgery and Pathology recommend proceeding to tissue diagnosis
  (excisional or core needle biopsy) to definitively characterize the
  lesion and rule out rare pediatric breast neoplasms, provided no
  systemic endocrine cause is identified.
\end{itemize}

\textbf{Missing Information and Required Next Steps} All specialists
note that critical data is absent from the provided case input. To
resolve the diagnostic uncertainty, the following results must be
reviewed:

\begin{itemize}
\tightlist
\item
  \textbf{Laboratory results (Table 1):} A full endocrine and tumor
  marker panel is essential to determine if the breast tissue
  proliferation and pubic hair are hormonally driven (e.g., evaluating
  for central precocious puberty, premature adrenarche, or
  estrogen/hCG-secreting tumors).
\item
  \textbf{Chest imaging results:} To exclude intrathoracic pathology,
  such as a hormone-secreting mediastinal mass.
\item
  \textbf{Details of "diagnostic procedures":} It is unclear what
  procedures were performed and whether tissue has already been
  acquired.
\item
  \textbf{Proposed additional testing:} If biochemical markers suggest
  an estrogenic or gonadotropin source, dedicated scrotal/testicular
  ultrasound and/or adrenal imaging have been proposed to exclude occult
  gonadal or adrenal neoplasms.
\end{itemize}

\textbf{Unresolved Disagreements} The primary etiology of the right
breast mass (physiologic gynecomastia vs. benign fibroepithelial
neoplasm) and the immediate need for tissue sampling versus observation
remain unresolved. All specialists agree that reviewing the pending
endocrine laboratory results and chest imaging is the necessary first
step before finalizing the management plan.

\endgroup

\clearpage
\section{Complete Structured Interactive-Consilium Report}
\label{app:example-report-structured}

This appendix reproduces in full the structured counterpart to the free-form report in Appendix~\ref{app:example-report-free}. Both reports concern the same case, allowing direct comparison of the two response formats. The report has been reformatted in \LaTeX{} for presentation; no report text has been omitted or edited.

\noindent\textbf{Case:} \href{https://doi.org/10.1056/NEJMcpc1503831}{10.1056/NEJMcpc1503831}, ``Case 12-2016---An 8-Year-Old Boy with an Enlarging Mass in the Right Breast.''

\noindent\textbf{Controller metadata:} consensus reached after one review round; stop reason, consensus.

\begingroup
\small
\providecommand{\tightlist}{%
  \setlength{\itemsep}{0pt}\setlength{\parskip}{0pt}}
\newcommand{\exampleCaseDOI}{\mbox{10.1056/NEJMcpc1503831}}

\subsection*{Medical Concilium}
\subsection{Contents}\label{app-structured-example-contents}

\begin{itemize}
\tightlist
\item
  \hyperref[app-structured-example-case-description]{Case Description}
\item
  \hyperref[app-structured-example-concilium-composition]{Concilium Composition}
\item
  \hyperref[app-structured-example-preconcilium-stage]{Preconcilium Stage}
\item
  \hyperref[app-structured-example-concilium-stage]{Concilium Stage}
\end{itemize}

\subsection{Case Description}\label{app-structured-example-case-description}

An 8-year-old boy was seen in an outpatient clinic of this hospital
because of an enlarging mass in the right breast.

One week before this presentation, the patient was seen by a
pediatrician at another medical facility for a routine annual
examination. He had a history of a mass in his right breast that had
been present for 18 months and had recently enlarged. On examination, a
round, mobile mass was palpable under the right areola. He was referred
to the pediatric endocrinology clinic of this hospital.

One week later, at the visit to the endocrinology clinic, the patient's
parents reported that, approximately 18 months earlier, they had noted a
mass under the right nipple that was not associated with discharge. The
mass had reportedly enlarged in the 6 months before this presentation,
and growth of fine pubic hair was also noted. The patient had seasonal
allergies and eczema and was otherwise well. He had a history of normal
growth and development. His only medication was topical triamcinolone
cream for eczema; his childhood vaccinations were current. He had no
known allergies to medications. He lived with his parents and younger
sibling and was doing well in second grade. His maternal grandfather had
asthma, his maternal grandmother had hypercholesterolemia, and his
paternal grandmother had breast cancer (which had been diagnosed when
she was 70 years of age); his parents and younger sibling were healthy.

Table 1.

Laboratory Data. On examination, the patient did not have dysmorphic
features. The blood pressure was 110/70 mm Hg, the pulse 86 beats per
minute, the height 131 cm (68th percentile), the weight 28.4 kg (72nd
percentile), and the body-mass index (BMI; the weight in kilograms
divided by the square of the height in meters) 16.5 (66th percentile).
The abdomen was soft and had some palpable bowel loops. A firm, mobile
mass (2 cm by 2 cm) was present under the right areola and was not
adherent to the skin. A small amount of soft breast tissue was palpable
on the left side. Fine, straight, lightly pigmented hairs were present
on the lower mons pubis, and on close inspection, very scant fine, light
axillary hairs were also present. The testes had an estimated volume of
3 ml; the phallus was prepubertal. The remainder of the examination was
normal. Findings on a radiograph of the left hand were consistent with a
bone age of 8 years. Eight days after this visit, laboratory tests were
performed; the test results are shown in Table 1. Thirteen days after
this visit, imaging studies of the chest were performed.

Figure 1.

Ultrasound Images of the Breasts. Dr. Pallavi Sagar: Ultrasound
examination of the right breast (Figure 1) revealed a
well-circumscribed, oval-shaped, relatively hypoechoic, mobile mass
(measuring approximately 1.4 cm by 1.3 cm by 0.8 cm) with
through-transmission in the retroareolar region. The mass had smooth
peripheral margins and was superficial to the pectoralis muscle; its
long axis was parallel to the skin and horizontal in orientation. On
color Doppler imaging, the mass was relatively hypovascular and had
minimal peripheral-blood flow. The mass did not have irregular
spiculated margins, calcifications, or increased vascularity, features
that are typically seen in aggressive cancerous lesions. Examination of
the contralateral breast revealed a normal left breast bud.

Dr. Baskaran: Diagnostic procedures were performed.

\subsection{Concilium Composition}\label{app-structured-example-concilium-composition}

\subsubsection{Lead Specialist}\label{app-structured-example-lead-specialist}

Pediatric Endocrinology

\subsubsection{Concilium Participants}\label{app-structured-example-concilium-participants}

\begin{itemize}
\tightlist
\item
  \textbf{Pediatric Endocrinology:} Leads the evaluation of premature
  adrenarche, hormonal causes of the breast mass, and pubertal
  assessment.
\item
  \textbf{Radiology:} Interprets breast ultrasound and chest imaging to
  characterize the retroareolar mass and exclude aggressive malignant
  features.
\item
  \textbf{Pediatric Surgery:} Assesses the enlarging solid breast mass
  for excisional biopsy or definitive surgical resection and management.
\item
  \textbf{Pathology:} Provides histologic diagnosis from tissue obtained
  via diagnostic procedures to confirm the mass etiology.
\end{itemize}

\subsubsection{Overall Rationale}\label{app-structured-example-overall-rationale}

A prepubertal boy with an enlarging breast mass and early pubertal signs
requires an endocrine-led evaluation, imaging characterization, tissue
diagnosis, and surgical management to establish etiology and direct
therapy.

\subsection{Preconcilium Stage}\label{app-structured-example-preconcilium-stage}

\subsubsection{Pathology}\label{app-structured-example-pathology}

\paragraph{Subjective and Objective
Information}\label{app-structured-example-subjective-and-objective-information}

Approximately 18 months before presentation (2016), the parents noted a
mass under the right nipple with no associated discharge (DOI:
\exampleCaseDOI{}). Significance: establishes chronicity and
absence of inflammatory or malignant discharge signs. Approximately 6
months before presentation, the mass enlarged and growth of fine pubic
hair was noted (DOI: \exampleCaseDOI{}). Significance: suggests
recent hormonal influence and possible adrenarche. One week before
presentation, a pediatrician palpated a round, mobile right retroareolar
mass on routine examination and referred the patient to pediatric
endocrinology (DOI: \exampleCaseDOI{}). Significance: confirms a
discrete mass requiring specialist evaluation. At the endocrinology
clinic visit (2016), physical examination revealed a firm, mobile 2 cm
by 2 cm mass under the right areola, not adherent to skin; a small
amount of soft breast tissue on the left; fine, straight, lightly
pigmented pubic hairs on the lower mons pubis and scant fine axillary
hairs; testes 3 ml with prepubertal phallus; height 131 cm (68th
percentile), weight 28.4 kg (72nd percentile), BMI 16.5 (66th
percentile); blood pressure 110/70 mm Hg, pulse 86 bpm; no dysmorphic
features (DOI: \exampleCaseDOI{}). Significance: unilateral, firm
mass with early pubic/axillary hair but prepubertal testes indicates
isolated premature adrenarche, not central precocious puberty.
Radiograph of the left hand: bone age 8 years, consistent with
chronological age (DOI: \exampleCaseDOI{}). Significance: no
advanced skeletal maturation, arguing against sustained estrogenic or
gonadal exposure. Eight days after the visit, laboratory tests were
performed; results are in Table 1 but not provided in the supplied text
(DOI: \exampleCaseDOI{}). Significance: hormonal profile unknown, a
critical data gap. Thirteen days after the visit, breast ultrasound
revealed a well-circumscribed, oval, hypoechoic, mobile retroareolar
mass measuring 1.4 x 1.3 x 0.8 cm, with smooth margins,
through-transmission, horizontal orientation, superficial to pectoralis,
and minimal peripheral color Doppler flow; no spiculated margins,
calcifications, or hypervascularity; left breast showed a normal bud
(DOI: \exampleCaseDOI{}). Significance: benign imaging features
make malignancy unlikely but do not replace histology. Diagnostic
procedures were performed (likely biopsy/excision), but results are not
yet available (DOI: \exampleCaseDOI{}). Significance: tissue
diagnosis pending, which is the definitive step. Past medical history
includes seasonal allergies and eczema treated with topical
triamcinolone; no other medications or endocrine disorders (DOI:
\exampleCaseDOI{}). Significance: no exogenous hormone exposure or
chronic disease. Family history includes paternal grandmother with
breast cancer at age 70; no known familial endocrine tumor syndromes
(DOI: \exampleCaseDOI{}). Significance: limited relevance given
late-onset breast cancer in a female relative.

\paragraph{Clinical Syndromes and Involved
Mechanisms}\label{app-structured-example-clinical-syndromes-and-involved-mechanisms}

The primary syndrome is a localized retroareolar proliferative lesion in
a prepubertal male. It is confirmed by an 18-month history of a firm,
mobile, 2 cm right retroareolar mass (DOI: \exampleCaseDOI{}) and a
corresponding well-circumscribed, hypoechoic, hypovascular sonographic
nodule measuring 1.4 x 1.3 x 0.8 cm with smooth margins and
through-transmission (DOI: \exampleCaseDOI{}). The pathogenetic
mechanism involves localized proliferation of ductal and stromal
elements within the breast bud, either as a hormonally influenced benign
process (prepubertal gynecomastia) or as a benign biphasic neoplasm
(fibroadenoma). The presence of fine pubic and axillary hair with
prepubertal testes (3 ml) and normal bone age indicates isolated
premature adrenarche (DOI: \exampleCaseDOI{}). This peripheral
androgenic activity may contribute to breast tissue stimulation via
aromatization, but the absence of central puberty suggests the hormonal
influence is limited. A malignant infiltrative syndrome is rejected on
imaging grounds because the mass lacks spiculated margins,
calcifications, and increased vascularity, and there is no skin fixation
(DOI: \exampleCaseDOI{}). However, imaging alone cannot exclude
rare low-grade malignancies such as secretory carcinoma, which can
present as a well-circumscribed mass in children.

\paragraph{Differential Diagnosis with
Reasoning}\label{app-structured-example-differential-diagnosis-with-reasoning}

High probability: Prepubertal gynecomastia. Arguments for: most common
cause of male pediatric breast masses; retroareolar location; small
breast tissue also noted on the left; sonographic features consistent
with benign glandular tissue (DOI: \exampleCaseDOI{}). Arguments
against: the mass is distinctly nodular and firm rather than the typical
diffuse discoid subareolar plaque, and there is no reported tenderness.
Confirmation: histology showing benign ductal proliferation with
periductal stromal proliferation without atypia, correlated with normal
or mildly altered estrogen-to-androgen ratio. Refutation: histology
showing a biphasic neoplasm or malignancy. High probability:
Fibroadenoma. Arguments for: sonographic
appearance---well-circumscribed, oval, smooth-margined, horizontally
oriented, mobile---is classic for fibroadenoma (DOI:
\exampleCaseDOI{}). Arguments against: rare in prepubertal males.
Confirmation: histology demonstrating a benign biphasic tumor with
intracanalicular or pericanalicular architecture. Medium probability:
Lipoma or other benign mesenchymal lesion. Arguments for:
well-circumscribed, mobile, hypoechoic mass. Arguments against:
retroareolar location is less typical, and the mass is hypoechoic rather
than classically hyperechoic. Confirmation: histology showing mature
adipose or benign mesenchymal tissue. Low probability but must be
excluded: Secretory carcinoma (juvenile breast carcinoma). Arguments
for: rare malignant breast tumor of childhood presenting as a painless
mass. Arguments against: no sonographic malignant features (no
spiculation, no increased vascularity, no calcifications), no skin
fixation (DOI: \exampleCaseDOI{}). Confirmation/exclusion:
histopathology and, if morphology is suspicious, molecular testing for
the characteristic ETV6-NTRK3 translocation. Low probability: Estrogen-
or hCG-secreting tumor (testicular, adrenal, hepatic). Arguments for:
prepubertal male with breast mass. Arguments against: symmetric
prepubertal testes, no abdominal mass, no dysmorphic features (DOI:
\exampleCaseDOI{}). Confirmation: elevated beta-hCG, estradiol, or
imaging lesion. Refutation: prepubertal gonadotropins, normal estradiol,
beta-hCG, and testicular ultrasound.

\paragraph{Diagnostic and Treatment
Tactics}\label{app-structured-example-diagnostic-and-treatment-tactics}

Diagnostics: Tissue diagnosis is mandatory to resolve the differential
and exclude malignancy. An ultrasound-guided core needle biopsy is the
initial pathologic procedure, with the goal of obtaining adequate tissue
for histology and immunohistochemistry while minimizing cosmetic
morbidity. Expected result: benign breast tissue consistent with
gynecomastia or fibroadenoma. If core needle biopsy is nondiagnostic or
shows atypia, an excisional biopsy should follow. If histology confirms
benign gynecomastia or fibroadenoma, observation or limited surgical
excision may follow based on symptoms. If secretory carcinoma or sarcoma
is identified, oncologic surgical planning with margin control and
staging is required. Histologic evaluation must assess for ductal versus
stromal predominance, cytologic atypia, infiltrative growth,
lymphovascular invasion, and mitotic activity. Immunohistochemistry
(e.g., S100, cytokeratins, ER, PR, myoepithelial markers) should be
performed if morphology is atypical. Radiologic-pathologic concordance
is mandatory: if imaging suggests benignity but pathology is discordant,
repeat biopsy or excision is required. Additionally, obtain the full
numeric results of Table 1 (LH, FSH, estradiol, testosterone, DHEA-S,
17-hydroxyprogesterone, beta-hCG, TSH, free T4, prolactin) to determine
whether the gonadal axis is prepubertal and to screen for an hCG- or
estrogen-secreting tumor. If basal LH is borderline or clinical
progression occurs, a GnRH stimulation test may be considered to exclude
central precocious puberty. Testicular ultrasound should be performed to
exclude a subclinical gonadal tumor if hormonal results are abnormal.
Treatment: If histology confirms prepubertal gynecomastia, no further
pathologic intervention is needed; clinical observation and endocrine
follow-up are managed by pediatrics and endocrinology. If fibroadenoma
is confirmed, simple excision is curative if symptomatic or growing. If
secretory carcinoma or sarcoma is diagnosed, ensure adequate margin
assessment on the excision specimen and coordinate sentinel lymph node
evaluation per pediatric oncology protocols. The surgical goal in any
excision is complete removal with preservation of the nipple-areolar
complex and developing breast tissue. Adjacent specialties: pediatric
endocrinology to provide the complete hormone panel and manage
adrenarche; pediatric radiology for image-guided biopsy and concordance
review; pediatric surgery for excisional biopsy or definitive resection;
pediatric oncology if malignancy is identified.

\paragraph{Final Judgment}\label{app-structured-example-final-judgment}

This 8-year-old boy presents with an 18-month history of a right
retroareolar mass that recently enlarged, now measuring 2 cm clinically
and 1.4 x 1.3 x 0.8 cm sonographically, in the setting of peripheral
adrenarche (fine pubic and axillary hair) but prepubertal testicular
volume (3 ml) and normal bone age (DOI: \exampleCaseDOI{}). The
leading pathologic considerations are benign localized
proliferations---most likely prepubertal gynecomastia or a
fibroadenoma---given the retroareolar location, patient age, smooth
sonographic margins, and absence of malignant imaging features (DOI:
\exampleCaseDOI{}). However, rare pediatric breast malignancies,
particularly secretory carcinoma, must be excluded because clinical and
imaging benignity cannot guarantee benign histology in this age group.
The priority action is to obtain a definitive tissue diagnosis via
ultrasound-guided core needle biopsy or excisional biopsy, with
permanent sections and appropriate immunohistochemical or molecular
studies if indicated. Concurrently, the missing endocrine laboratory
values from Table 1 must be reviewed to correlate any hormonal influence
with the histopathologic findings. The most impactful data gaps are the
unavailable endocrine laboratory values and the pending histopathologic
results, both of which are required to distinguish a hormonally driven
benign process from a neoplasm and to guide whether simple observation,
local excision, or oncologic staging is appropriate (DOI:
\exampleCaseDOI{}).

\subsubsection{Pediatric Endocrinology}\label{app-structured-example-pediatric-endocrinology}

\paragraph{Subjective and Objective
Information}\label{app-structured-example-subjective-and-objective-information-1}

Approximately 18 months before presentation (date not specified), the
parents noted a mass under the right nipple without discharge (DOI:
\exampleCaseDOI{}). This establishes chronicity and absence of
inflammatory or galactorrheic features. Six months before presentation,
the mass began to enlarge, and growth of fine pubic hair was noted,
indicating recent acceleration and possible hormonal influence (DOI:
\exampleCaseDOI{}). One week prior to the endocrinology evaluation,
a pediatrician palpated a round, mobile mass under the right areola and
referred the patient (DOI: \exampleCaseDOI{}). At the endocrinology
clinic visit in 2016, the patient had a history of normal growth and
development, seasonal allergies, and eczema treated with topical
triamcinolone cream; family history included paternal grandmother with
breast cancer at age 70 (DOI: \exampleCaseDOI{}). This largely
excludes exogenous systemic hormone exposure and suggests low-risk
family history. On examination, blood pressure was 110/70 mm Hg, pulse
86 beats per minute, height 131 cm (68th percentile), weight 28.4 kg
(72nd percentile), and BMI 16.5 (66th percentile), indicating normal
growth without obesity (DOI: \exampleCaseDOI{}). A firm, mobile 2
cm by 2 cm mass was present under the right areola and was not adherent
to the skin, while a small amount of soft breast tissue was palpable on
the left, signifying bilateral but asymmetric gynecomastia (DOI:
\exampleCaseDOI{}). Fine, straight, lightly pigmented hairs were
present on the lower mons pubis and very scant fine axillary hairs were
noted, but the testes were 3 mL and the phallus was prepubertal,
signifying adrenarche without gonadarche (DOI: \exampleCaseDOI{}).
Bone age by left hand radiograph was 8 years, consistent with
chronological age and arguing against significant systemic estrogen
effect (DOI: \exampleCaseDOI{}). Eight days after the visit,
laboratory tests were drawn (Table 1), but the specific values are not
provided in the supplied text, representing a critical data gap (DOI:
\exampleCaseDOI{}). Thirteen days after the visit, chest imaging
studies were performed, but results are not provided, leaving evaluation
for ectopic hCG-secreting tumors incomplete (DOI:
\exampleCaseDOI{}). Ultrasound of the right breast revealed a
well-circumscribed, oval, hypoechoic, mobile retroareolar mass measuring
approximately 1.4 cm by 1.3 cm by 0.8 cm with smooth margins, minimal
peripheral vascularity, and no calcifications, spiculation, or increased
vascularity; the left breast showed a normal bud (DOI:
\exampleCaseDOI{}). These sonographic features are significant
because they are classic for benign glandular tissue and argue against
aggressive malignancy. Diagnostic procedures were performed
(unspecified), representing another critical data gap (DOI:
\exampleCaseDOI{}). Important negative findings include: no nipple
discharge, no skin adherence or retraction, no dysmorphic features, no
testicular enlargement, no advanced bone age, and no malignant
ultrasound characteristics.

\paragraph{Clinical Syndromes and Involved
Mechanisms}\label{app-structured-example-clinical-syndromes-and-involved-mechanisms-1}

The main syndrome is unilateral prepubertal gynecomastia with a
contralateral breast bud. It is confirmed by an 18-month history of a
right subareolar mass that enlarged over 6 months, a firm mobile 2x2 cm
right retroareolar mass on exam, and a sonographically
well-circumscribed hypoechoic retroareolar mass with benign features
(DOI: \exampleCaseDOI{}). The mechanism is proliferation of ductal
and periductal fibroglandular tissue driven by an imbalance between
estrogen and androgen action at the breast bud. In this prepubertal boy
with normal bone age and prepubertal testes, any estrogenic drive is
likely peripheral (adrenal, exogenous, or increased aromatization)
rather than centrally mediated. A concurrent syndrome is premature
adrenarche, confirmed by fine pubic and scant axillary hair with
prepubertal 3 ml testes and a prepubertal phallus (DOI:
\exampleCaseDOI{}). This is realized through adrenal zona
reticularis secretion of weak androgens (DHEA/DHEA-S) independent of
gonadotropin stimulation. Central precocious puberty is rejected because
testicular volume is 3 ml (prepubertal), the phallus is prepubertal, and
bone age is not advanced, which are incompatible with central HPG axis
activation.

\paragraph{Differential Diagnosis with
Reasoning}\label{app-structured-example-differential-diagnosis-with-reasoning-1}

High probability: Idiopathic or early physiologic prepubertal
gynecomastia associated with premature adrenarche. For: classic clinical
and sonographic features, prepubertal gonadarche, normal growth and bone
age. Against: persistence for 18 months with recent enlargement is
unusual for typical physiologic prepubertal gynecomastia. Confirm:
normal basal labs and stability/regression on follow-up. Refute:
elevated sex steroids or tumor localization. Medium probability:
Peripheral estrogen excess due to occult testicular or adrenal neoplasm.
For: unilateral enlarging gynecomastia suggests active hormonal
stimulation. Against: no palpable testicular or abdominal mass, normal
bone age. Confirm: elevated estradiol with suppressed LH/FSH,
testicular/adrenal imaging. Refute: normal imaging and estradiol. Low
probability (must exclude): hCG-secreting germ cell tumor (testicular,
mediastinal, CNS). For: can cause gynecomastia via Leydig cell
stimulation. Against: no testicular mass, no neurologic/respiratory
symptoms. Confirm: elevated hCG, imaging. Refute: undetectable hCG,
normal imaging. Low probability: Late-onset congenital adrenal
hyperplasia (CAH). For: premature pubic hair. Against: normal bone age,
no virilization, gynecomastia not typical. Confirm: elevated 17-OHP,
ACTH stim test. Refute: normal 17-OHP. Low probability: Exogenous
estrogen exposure. For: can cause gynecomastia. Against: no history of
exposure, triamcinolone is not estrogenic. Confirm: exposure history,
elevated estradiol. Refute: no exposure, normal labs. Low probability:
Malignant breast lesion. For: enlarging mass. Against: benign ultrasound
features, very rare in this age. Confirm: biopsy. Refute: benign
histology.

\paragraph{Diagnostic and Treatment
Tactics}\label{app-structured-example-diagnostic-and-treatment-tactics-1}

Diagnostics: The immediate priority is to review the pending laboratory
panel (Table 1) and the results of the unspecified diagnostic procedure.
Required lab results include LH, FSH, estradiol, total testosterone,
DHEA-S, androstenedione, 17-hydroxyprogesterone, serum hCG, AFP, and
prolactin. If estradiol is elevated with suppressed LH/FSH, obtain
testicular ultrasound and adrenal CT/MRI to localize an
estrogen-secreting source. If hCG is elevated, perform testicular
ultrasound and chest/abdominal/CNS imaging to exclude ectopic
hCG-secreting tumors. If all hormones are prepubertal/normal, consider a
GnRH stimulation test to detect subclinical gonadarche or proceed to
tissue diagnosis if the mass persists or enlarges. A scrotal/testicular
ultrasound is mandatory to exclude an occult testicular neoplasm not
palpable on examination. Treatment: If benign physiologic gynecomastia
is confirmed by normal labs and imaging, reassurance and observation
with 3-6 month follow-up is the primary approach. Avoid empirical
anti-estrogen or aromatase inhibitor therapy until a hormonally active
tumor is excluded. If a specific cause is identified (e.g., tumor, CAH),
treat accordingly. If the mass persists, enlarges beyond 3-6 months, or
causes significant psychological distress despite a negative workup,
consider excisional biopsy or surgical referral for definitive
histology.

\paragraph{Final Judgment}\label{app-structured-example-final-judgment-1}

An 8-year-old boy presents with an 18-month history of a right breast
mass that enlarged over 6 months, accompanied by fine pubic and axillary
hair (DOI: \exampleCaseDOI{}). Examination reveals a 2-cm firm,
mobile retroareolar mass, a small left breast bud, Tanner 2 pubic hair,
prepubertal testes (3 ml), and normal bone age. Ultrasound shows a
benign-appearing solid mass consistent with gynecomastia. The leading
syndromes are unilateral prepubertal gynecomastia and premature
adrenarche, while central precocious puberty is excluded by prepubertal
testes and normal bone age. The top hypothesis is idiopathic or early
physiologic gynecomastia, but peripheral estrogen excess
(testicular/adrenal tumor) and hCG-secreting tumors must be excluded.
Priority actions are to review the pending laboratory results (Table 1)
and the unspecified diagnostic procedure, and to perform a scrotal
ultrasound. If labs and imaging are normal, management is expectant with
reassurance. The greatest data gaps are the unavailable lab values and
diagnostic procedure results, which are essential to differentiate a
benign self-limited condition from an occult hormonally active tumor.

\subsubsection{Pediatric Surgery}\label{app-structured-example-pediatric-surgery}

\paragraph{Subjective and Objective
Information}\label{app-structured-example-subjective-and-objective-information-2}

In March 2016 (DOI: \exampleCaseDOI{}), an 8-year-old boy presented
to the outpatient clinic with an 18-month history of a right breast mass
that had enlarged over the preceding 6 months. One week earlier, a
pediatrician at another facility had palpated a round, mobile subareolar
mass under the right areola, prompting referral. This chronology
establishes a subacute-to-chronic process with recent progression,
significant for pediatric surgery because a persistent and enlarging
discrete retroareolar mass in a prepubertal boy raises concern beyond
simple prepubertal gynecomastia and warrants surgical-pathologic
correlation. At the endocrinology clinic visit, parents denied nipple
discharge. History included seasonal allergies, eczema treated with
topical triamcinolone, and normal growth and development; no systemic
hormonal or off-label exposure is reported, which is relevant because
exogenous systemic steroids and certain medications can produce
gynecomastia, and absence of such exposure supports an endogenous or
idiopathic mechanism. Family history includes paternal grandmother with
breast cancer diagnosed at age 70, maternal grandfather with asthma, and
maternal grandmother with hypercholesterolemia; parents and sibling are
healthy. An elderly-onset female relative with breast cancer does not
significantly raise the index of suspicion for an inherited male breast
cancer syndrome in this 8-year-old, but it should be noted. On
examination, the patient appeared well, without dysmorphic features;
blood pressure 110/70 mm Hg, pulse 86 bpm, height 131 cm (68th
percentile), weight 28.4 kg (72nd percentile), BMI 16.5 (66th
percentile). Normal stature and BMI argue against an underlying
endocrinopathy such as hypothyroidism or hypercortisolism. A firm,
mobile 2 \(\times\) 2 cm mass was palpable under the right areola, not adherent
to the skin; a small amount of soft breast tissue was palpable on the
left; fine, straight, lightly pigmented pubic hairs were noted on the
lower mons pubis, and very scant fine, light axillary hair was present
on close inspection. For pediatric surgery, the key findings are the
size (\textgreater2 cm, clearly outside the usual
\textquotesingle breast bud\textquotesingle{} range of pubertal
gynecomastia), unilateral predominance, firmness, mobility, and
non-adherence to skin or deep structures---features that mandate imaging
and likely excisional biopsy. Testicular volume was 3 ml with a
prepubertal phallus. Prepubertal testicular volume essentially excludes
a testosterone-secreting testicular tumor as the cause of the breast
mass, although it does not exclude an estrogen-secreting gonadal or
extragonadal source. A left-hand radiograph showed a bone age of 8
years, consistent with chronologic age, arguing against a significant
systemic hormonal drive and supporting an isolated, probably benign,
prepubertal process. Eight days after the endocrinology visit,
laboratory tests (Table 1) were obtained, and 13 days after the visit,
chest imaging was performed. The specific numeric laboratory values are
not reproduced in the supplied text and constitute a data gap for the
preconcilium analysis. Breast ultrasound (Dr. Pallavi Sagar, Figure 1)
demonstrated a well-circumscribed, oval, relatively hypoechoic, mobile
mass approximately 1.4 \(\times\) 1.3 \(\times\) 0.8 cm in the retroareolar region, with
smooth peripheral margins, superficial to the pectoralis muscle, long
axis parallel to the skin, minimal peripheral blood flow on color
Doppler, and no irregular margins, calcifications, or hypervascularity;
the contralateral left breast showed a normal breast bud.
Sonographically, the lesion is solid-appearing, well-defined,
superficial, and oriented parallel to the skin---features classically
described for a fibroadenoma-like or fibro-glandular proliferation, but
in a prepubertal boy the differential remains restricted and tissue
diagnosis is required. Important negative findings include absence of
skin retraction or fixation, nipple discharge, palpable lymphadenopathy,
dysmorphic features, and testicular masses, reducing the likelihood of
advanced or metastatic disease and hormonally active testicular tumors.
Data gap: the numeric results of the hormonal and tumor-marker panel in
Table 1 and the findings of the chest imaging study are not reproduced
in the supplied case excerpt; their values are necessary before final
surgical decision-making.

\paragraph{Clinical Syndromes and Involved
Mechanisms}\label{app-structured-example-clinical-syndromes-and-involved-mechanisms-2}

The principal syndrome is a unilateral, persistent, enlarging breast
mass in a prepubertal male, confirmed by the 18-month history with
recent enlargement, the palpable 2 cm by 2 cm firm mobile right
retroareolar mass in March 2016, and the sonographic finding of a
well-circumscribed, hypoechoic, hypovascular nodule without malignant
features (DOI: \exampleCaseDOI{}). The mass is realized through
localized proliferation or hormonal stimulation of mammary tissue in the
retroareolar region, superficial to the pectoralis muscle. The
pathogenetic mechanism that most plausibly underlies this syndrome is
localized, asymmetric proliferation of ductal and periductal
fibro-glandular tissue in the right breast, consistent with idiopathic
prepubertal or early-pubertal gynecomastia; in this child, the mass has
progressed beyond the typical small, transient \textquotesingle breast
bud,\textquotesingle{} which is why surgical sampling is being
considered. A lower-probability mechanism is a true neoplasm of the
prepubertal male breast---most commonly a fibroepithelial lesion
analogous to a juvenile fibroadenoma, and very rarely a ductal carcinoma
or other malignancy---in which the same fibro-glandular proliferation is
clonal/expansile rather than hormonally driven. This is considered
because the lesion is unilateral, \textgreater2 cm, persistent for 18
months, and continues to enlarge, features that exceed the usual natural
history of simple pubertal gynecomastia. A concurrent syndrome of
premature adrenarche is suggested by fine pubic and axillary hair in an
8-year-old with prepubertal 3 mL testes, prepubertal phallus, and bone
age of 8 years; this is realized through early adrenal androgen
secretion independent of gonadal activation. This adrenarche is
significant because it does not explain the unilateral breast
mass---systemic estrogen excess would typically produce bilateral breast
enlargement and testicular or phallic changes, which are
absent---thereby supporting a localized rather than systemic endocrine
mechanism. The absence of central puberty is confirmed by prepubertal
genitalia and normal bone age, rejecting a
hypothalamic-pituitary-gonadal axis-driven process as the cause of the
mass. A hormonally active endocrine tumor (for example, an hCG- or
estrogen-secreting gonadal or extragonadal germ-cell tumor) producing
secondary gynecomastia is explicitly considered and rejected on the
basis of prepubertal testicular volume (3 ml) and prepubertal phallus,
normal stature and BMI, bone age equal to chronologic age, and absence
of any reported features of precocious puberty beyond scant
pubic/axillary hair. If Table 1 hormonal and tumor-marker results show
elevated hCG, estradiol, or suppressed LH/FSH, this rejection must be
revisited. A chest-wall or pectoral soft-tissue neoplasm (for example,
rhabdomyosarcoma, desmoid tumor, or vascular malformation) presenting as
a \textquotesingle breast\textquotesingle{} mass is rejected because the
lesion is retroareolar within the breast disc, mobile, non-adherent, and
superficial to the pectoralis muscle on ultrasound.

\paragraph{Differential Diagnosis with
Reasoning}\label{app-structured-example-differential-diagnosis-with-reasoning-2}

High probability: persistent idiopathic (prepubertal/early-pubertal)
unilateral gynecomastia or benign fibroepithelial lesion (e.g.,
fibroadenoma). Arguments for: typical age, unilateral subareolar
location, mobility, lack of skin fixation, ultrasound features of a
well-circumscribed, parallel-to-skin, hypovascular, hypoechoic
retroareolar mass with no malignant features, prepubertal testicular
exam, and bone age equal to chronologic age (DOI:
\exampleCaseDOI{}). Arguments against: the lesion has persisted for
18 months, is enlarging, measures \textgreater2 cm clinically, and is
firm and discrete---features more pronounced than the typical
self-limited pubertal breast bud. What would confirm: complete
resolution or regression after a defined observation interval, or
histology showing ductal hyperplasia with periductal stromal fibrosis
and no neoplastic features after excisional biopsy. What would refute:
histology demonstrating a neoplastic fibroepithelial or other tumor.
Medium probability: benign fibroepithelial neoplasm of the prepubertal
male breast---juvenile fibroadenoma-like lesion. Arguments for:
persistent, unilateral, well-circumscribed, solid, mobile,
sonographically \textquotesingle benign\textquotesingle{} mass that
continues to grow in a prepubertal boy; fibroepithelial tumors are the
most common breast neoplasm in adolescents of both sexes. Arguments
against: this entity is exceedingly rare in 8-year-old boys and the
imaging is not specific. What would confirm: histopathology
demonstrating a pericanalicular or intracanalicular pattern with a
monoclonal-appearing stromal component. What would refute: histology
showing only usual ductal hyperplasia without a clonal stromal
proliferation. Low probability but must be excluded: malignant breast
neoplasm (for example, secretory carcinoma, ductal carcinoma, or other
rare pediatric breast malignancy). Arguments for: persistence and
progressive enlargement of a unilateral, firm, discrete mass. Arguments
against: classic sonographic benign features (smooth margins, parallel
orientation, no calcifications, no hypervascularity, no spiculation) and
the absence of any suspicious clinical findings such as skin retraction,
nipple discharge, fixation, or lymphadenopathy. Because this category,
even at low probability, can be life-threatening, it must be actively
excluded by histologic diagnosis before the mass is accepted as benign.
Low probability: estrogen- or hCG-secreting tumor (testicular, adrenal,
or germ cell). Arguments for include gynecomastia and adrenarche.
Arguments against include prepubertal testes (3 mL) without masses,
prepubertal phallus, normal bone age, and absence of other feminizing
signs. Confirmation would be elevated estradiol or hCG with
imaging-identified tumor; refutation would be normal hormone levels and
negative imaging. Low probability: drug-induced gynecomastia from
topical triamcinolone. Arguments for include documented exposure.
Arguments against include low systemic absorption with topical use and
absence of cushingoid features. Confirmation would be regression after
discontinuation; refutation is persistence after cessation. The
potentially dangerous conditions of malignancy and hormonally active
tumors must be excluded first through histopathology and hormone
studies.

\paragraph{Diagnostic and Treatment
Tactics}\label{app-structured-example-diagnostic-and-treatment-tactics-2}

Diagnostics: The first surgical action is review of the Table 1 hormonal
and tumor-marker results (hCG, LH, FSH, estradiol, testosterone,
prolactin, thyroid function, DHEA-S, 17-OHP, androstenedione as
available) and of the chest imaging performed 13 days after the
endocrinology visit. The expected result is normal age-appropriate
gonadotropin/steroid profile and a chest image without intrathoracic or
osseous lesion; this would reinforce the idiopathic/localized diagnosis
and permit a targeted surgical plan. If any hormonal value is abnormal
(for example elevated hCG, elevated estradiol, or suppressed LH/FSH),
the surgical plan must be deferred in favor of endocrine and oncologic
work-up, because the breast mass may then be a manifestation of a
gonadal or extragonadal tumor and not an isolated surgical problem. The
second action is a focused surgical examination under anesthesia at the
time of planned excision to confirm that the mass is indeed discrete,
mobile, retroareolar, and not fixed to pectoralis fascia or skin, and to
inspect for any clinically occult contralateral lesion. The third action
is complete excisional biopsy of the right retroareolar mass through a
periareolar or circumareolar incision, with the specimen sent for
histopathology in fresh, oriented, and inked form. The expected result
is definitive histologic diagnosis distinguishing usual ductal
hyperplasia from a fibroepithelial or other neoplasm, and this result
will determine all subsequent management. If histology returns as usual
ductal hyperplasia/stromal fibrosis consistent with gynecomastia, no
further surgical treatment is needed and the family is counseled that
recurrence on the contralateral side is possible during puberty. If
histology returns as a fibroepithelial neoplasm (for example juvenile
fibroadenoma), complete excision is considered adequate therapy provided
margins are uninvolved; the contralateral left \textquotesingle breast
bud\textquotesingle{} is observed. If histology is malignant, the
patient is referred immediately to pediatric oncology for staging (chest
imaging, breast MRI if indicated, axillary ultrasound, and systemic
work-up) and for multidisciplinary surgical-oncologic planning.
Treatment: The principal treatment proposed at this preconcilium stage
is right periareolar excisional biopsy of the retroareolar mass as a
combined diagnostic and therapeutic maneuver, performed under general
anesthesia, with the patient supine, the arm on the side of the lesion
abducted, and a small inferior or lateral periareolar incision planned
to preserve nipple-areolar viability and to allow access to the
retroareolar disc. The rationale is that the lesion is unilateral,
persistent for 18 months, enlarging, \textgreater2 cm clinically, and
sonographically solid, and therefore exceeds the threshold for simple
observation of prepubertal gynecomastia. The goal of the operation is
complete excision of the mass with a small margin of surrounding breast
tissue, hemostasis, and preservation of the nipple-areolar complex;
criteria for surgical readiness are normal coagulation, normal
preoperative endocrine panel (or endocrinology clearance if any value is
abnormal), and signed informed consent including discussion of risks of
hematoma, seroma, infection, nipple-areolar compromise, scarring,
recurrence, and the possibility of finding an unexpected neoplasm.
Pharmacologic treatment is not indicated preoperatively; the topical
triamcinolone for eczema should be continued and is not a
contraindication. Postoperative care includes a soft dressing,
analgesia, activity restriction for 1--2 weeks, and wound review at 1--2
weeks with pathology results; no hormonal therapy is initiated.
Effectiveness and safety monitoring is by histopathology of the excised
specimen (definitive diagnostic endpoint) and by clinical wound review
at 1--2 weeks and again at 3 months to detect recurrence, contour
deformity, or contralateral changes. Expectations from adjacent
specialties: endocrinology is asked to provide the numeric Table 1
results and interpretation, and to confirm that no treatable endocrine
cause is present before the surgical date; radiology is asked to make
the breast ultrasound images available intraoperatively and to comment
on the chest imaging already performed; pathology is asked to process
the specimen for routine histology and to retain tissue for potential
additional stains if a fibroepithelial or unusual lesion is identified;
oncology is on standby in the unlikely event of malignancy.

\paragraph{Final Judgment}\label{app-structured-example-final-judgment-2}

This 8-year-old, phenotypically normal boy with a unilateral, firm,
mobile, 2 \(\times\) 2 cm, non-adherent, enlarging right retroareolar mass
present for 18 months, accompanied by only minimal contralateral breast
tissue and very early pubic/axillary hair, prepubertal testicular volume
of 3 ml, prepubertal phallus, bone age equal to chronologic age (8
years), and a sonographically well-circumscribed, oval, hypoechoic,
hypovascular retroareolar lesion without features of malignancy,
presents as a surgical case at the preconcilium stage with a working
diagnosis of persistent idiopathic prepubertal unilateral gynecomastia
or a benign fibroepithelial lesion (high probability), with
juvenile-type fibroepithelial neoplasm as an intermediate alternative
and prepubertal male breast malignancy as a low-probability but
mandatory-to-exclude category (DOI: \exampleCaseDOI{}). The first
surgical priority is to obtain and review the Table 1 hormonal and
tumor-marker results and the chest imaging already performed before the
operative date, because any abnormal endocrine value would redirect
management away from a straightforward excisional biopsy and toward an
endocrine or oncologic work-up. Once the endocrine panel is confirmed
normal, the priority surgical action is right periareolar excisional
biopsy of the retroareolar mass for definitive histologic diagnosis,
which simultaneously serves as treatment if the lesion proves to be a
benign fibroepithelial tumor, while malignancy---though unlikely on the
available imaging---must be definitively excluded by tissue. The data
gaps with the greatest impact on the surgical decision are the missing
numeric values from Table 1 (hCG, LH, FSH, estradiol, testosterone,
prolactin, adrenal steroids) and the formal report of the chest imaging
study, both of which must be reviewed before the patient is taken to the
operating room. With those data available and reviewed, the proposed
preconcilium plan is a single, definitive excisional biopsy via a
periareolar approach, with endocrinology providing preoperative
clearance, radiology providing intraoperative ultrasound support,
pathology providing definitive histology, and pediatric oncology on
standby only if malignancy is unexpectedly identified (DOI:
\exampleCaseDOI{}).

\subsubsection{Radiology}\label{app-structured-example-radiology}

\paragraph{Subjective and Objective
Information}\label{app-structured-example-subjective-and-objective-information-3}

Approximately 18 months before presentation (2014-2015), the parents
noted a mass under the right nipple without discharge (DOI:
\exampleCaseDOI{}). This establishes chronicity and benign initial
behavior. Six months before presentation, the mass enlarged and fine
pubic hair was noted (DOI: \exampleCaseDOI{}), suggesting hormonal
stimulation coinciding with adrenarche. At presentation (2016), physical
exam revealed a firm, mobile 2x2 cm mass under the right areola not
adherent to skin, a small amount of soft left breast tissue, fine pubic
and scant axillary hair, and prepubertal testes (3 ml) with a
prepubertal phallus (DOI: \exampleCaseDOI{}). This indicates a
unilateral discrete mass with early sexual hair but prepubertal gonads.
A radiograph of the left hand showed a bone age of 8 years, consistent
with chronological age (DOI: \exampleCaseDOI{}), excluding advanced
skeletal maturation. Eight days after the visit, laboratory tests were
performed (Table 1), but the values are not provided in the available
text (DOI: \exampleCaseDOI{}), creating a critical data gap
regarding hormone levels. Thirteen days after the visit, imaging studies
of the chest were performed, but the modality and findings are not
provided (DOI: \exampleCaseDOI{}), creating another explicit data
gap. Breast ultrasound (Figure 1) revealed a well-circumscribed, oval,
hypoechoic, mobile mass (1.4x1.3x0.8 cm) in the retroareolar region with
smooth margins, through-transmission, parallel orientation, and minimal
peripheral flow, without spiculation, calcifications, or increased
vascularity; the left breast showed a normal bud (DOI:
\exampleCaseDOI{}). These features are characteristic of a benign
solid lesion.

\paragraph{Clinical Syndromes and Involved
Mechanisms}\label{app-structured-example-clinical-syndromes-and-involved-mechanisms-3}

The dominant imaging-supported syndrome is a unilateral,
benign-appearing retroareolar breast mass. This is confirmed by the
sonographic findings of a well-circumscribed, oval, hypoechoic, mobile
mass with smooth margins, through-transmission, parallel orientation,
and minimal peripheral flow (DOI: \exampleCaseDOI{}, Figure 1). The
mechanism involves hormone-sensitive proliferation of breast tissue;
however, the discrete, nodular morphology suggests a localized
fibroepithelial proliferation (such as a fibroadenoma) rather than the
diffuse ductal and stromal proliferation typical of gynecomastia. A
second relevant syndrome is early adrenarche, supported by the presence
of fine, straight, lightly pigmented pubic hairs and scant axillary
hairs at age 8, alongside prepubertal testes (3 mL) and a prepubertal
phallus (DOI: \exampleCaseDOI{}). This is realized through early
adrenal androgen production, which can be peripherally aromatized to
estrogen, potentially stimulating localized breast tissue growth. The
concordance of bone age with chronologic age (8 years) argues against a
long-standing, advanced systemic sex-steroid effect. Considered but
rejected is true central precocious puberty, as the testes remain
prepubertal in volume and the bone age is not advanced.

\paragraph{Differential Diagnosis with
Reasoning}\label{app-structured-example-differential-diagnosis-with-reasoning-3}

High probability: Juvenile fibroadenoma. Arguments for: the mass is
discrete, oval, well-circumscribed, and hypoechoic with
through-transmission and smooth margins---classic features of a
fibroadenoma; fibroadenomas can occur in males during periods of
hormonal change. Arguments against: rare in prepubertal males. What
would confirm: histology showing biphasic proliferation of stroma and
epithelium. What would refute: regression or evolution into fan-shaped
tissue on follow-up. Medium probability: Unilateral gynecomastia.
Arguments for: retroareolar location, recent enlargement, and background
of early adrenarche with possible estrogen effect. Arguments against:
gynecomastia typically appears as fan-shaped, hypoechoic tissue blending
with surrounding fat, not a discrete, mobile mass with a capsule-like
margin. What would confirm: resolution or flattening on follow-up
without intervention. What would refute: persistence as a discrete
nodule. Medium probability: Gynecomastia secondary to an estrogen- or
hCG-secreting testicular or adrenal tumor. Arguments for: unilateral
breast enlargement and early pubic hair in a prepubertal boy. Arguments
against: no palpable testicular mass, bone age is not advanced. What
would confirm: scrotal or adrenal imaging demonstrating a mass, or
elevated estradiol/hCG on laboratory testing. What would refute: normal
testicular and adrenal imaging. Low probability but must be excluded:
Male breast carcinoma. Arguments for: any solid breast mass warrants
consideration of malignancy; remote family history of breast cancer.
Arguments against: benign sonographic features lacking irregular
margins, calcifications, and hypervascularity; exceedingly rare in
prepubertal children. What would confirm: aggressive sonographic
features or biopsy. What would refute: benign histology or long-term
stability.

\paragraph{Diagnostic and Treatment
Tactics}\label{app-structured-example-diagnostic-and-treatment-tactics-3}

Diagnostics: The primary goal is to confirm the benign nature of the
mass and exclude rare pathologic hormone sources. Scrotal ultrasound
with color Doppler is the next essential imaging step to exclude a
small, non-palpable estrogen-secreting testicular tumor (e.g., Leydig
cell tumor). If scrotal ultrasound is normal, adrenal imaging (MRI or
contrast-enhanced CT) should be considered if pending laboratory values
(Table 1) suggest an adrenal source of estrogen excess. Follow-up breast
ultrasound in 3 to 6 months is recommended to document interval
stability; development of irregular margins, increased vascularity, or
calcifications would prompt ultrasound-guided core biopsy. Treatment:
Conservative observation is the appropriate initial approach given the
benign imaging phenotype. Pharmacologic therapy or surgical excision is
not indicated unless the mass grows disproportionately, becomes
symptomatic, or develops suspicious imaging features. Expectations from
adjacent specialties: Pediatric endocrinology must provide the results
of the pending laboratory tests (Table 1) to guide further
cross-sectional imaging decisions.

\paragraph{Final Judgment}\label{app-structured-example-final-judgment-3}

An 8-year-old boy presented with an 18-month history of a right breast
mass that enlarged over the preceding 6 months, accompanied by fine
pubic and scant axillary hair but with prepubertal testes (3 mL) and a
prepubertal phallus (DOI: \exampleCaseDOI{}). Breast ultrasound
demonstrated a 1.4 \(\times\) 1.3 \(\times\) 0.8 cm well-circumscribed, oval, hypoechoic,
hypovascular retroareolar mass with through-transmission, smooth
margins, and parallel orientation, lacking malignant features; the left
breast showed a normal bud (DOI: \exampleCaseDOI{}, Figure 1). The
leading hypothesis is a juvenile fibroadenoma, given the discrete,
nodular morphology with classic benign solid-lesion features, though
unilateral gynecomastia driven by early adrenarche remains a
consideration. Bone age is concordant with chronological age, arguing
against chronic systemic sex-steroid excess. Priority actions are to
perform scrotal ultrasound to exclude a non-palpable testicular neoplasm
and to review the pending laboratory values (Table 1) to determine if
adrenal imaging is warranted. If targeted imaging and laboratory results
are negative, follow-up breast ultrasound in 3 to 6 months is
appropriate. The most impactful data gaps are the missing laboratory
values (Table 1) and the unreported results of the chest imaging
performed 13 days after presentation, both of which are essential to
narrow the differential and direct further workup.

\subsection{Concilium Stage}\label{app-structured-example-concilium-stage}

\subsubsection{Final Opinion}\label{app-structured-example-final-opinion}

\paragraph{Concilium Summary}\label{app-structured-example-concilium-summary}

An 8-year-old boy presents with an 18-month history of a right
retroareolar breast mass that has enlarged over the last 6 months,
accompanied by the recent onset of fine pubic and axillary hair.
Physical examination reveals a firm, mobile 2x2 cm mass under the right
areola, a small left breast bud, prepubertal testes (3 ml), and a
prepubertal phallus. Bone age is concordant with chronological age (8
years). Breast ultrasound shows a well-circumscribed, oval, hypoechoic,
hypovascular mass with benign features. The specialists agree that the
presentation is most consistent with a benign localized process (either
prepubertal gynecomastia or a fibroepithelial lesion such as a
fibroadenoma) in the setting of premature adrenarche, while central
precocious puberty is excluded. Critical data gaps include the
unprovided results of Table 1 laboratory tests and chest imaging. There
is a disagreement regarding the next diagnostic step: radiology favors
conservative observation with follow-up ultrasound, while surgery and
pathology favor tissue diagnosis via excisional or core needle biopsy
due to the mass\textquotesingle s persistence, size, and enlargement.

\paragraph{Final Differential
Diagnosis}\label{app-structured-example-final-differential-diagnosis}

\begin{enumerate}
\def\labelenumi{\arabic{enumi}.}
\tightlist
\item
  Prepubertal gynecomastia (idiopathic or associated with premature
  adrenarche)
\item
  Benign fibroepithelial lesion (e.g., juvenile fibroadenoma)
\item
  Hormone-secreting tumor (testicular, adrenal, or extragonadal germ
  cell tumor)
\item
  Malignant breast neoplasm (e.g., secretory carcinoma)
\end{enumerate}

\paragraph{Final Diagnostic Plan}\label{app-structured-example-final-diagnostic-plan}

\begin{enumerate}
\def\labelenumi{\arabic{enumi}.}
\tightlist
\item
  Review the pending Table 1 laboratory results (including LH, FSH,
  estradiol, testosterone, DHEA-S, 17-hydroxyprogesterone, beta-hCG, and
  prolactin) to evaluate for a systemic hormonal or tumor-mediated
  process.
\item
  Review the results of the chest imaging performed 13 days after the
  endocrinology visit.
\item
  Perform a scrotal/testicular ultrasound to exclude an occult estrogen-
  or hCG-secreting testicular neoplasm.
\item
  Depending on the resolution of the disagreement between observation
  and biopsy, either perform a follow-up breast ultrasound in 3-6 months
  or obtain a definitive tissue diagnosis via excisional biopsy or
  ultrasound-guided core needle biopsy.
\end{enumerate}

\paragraph{Final Management Plan}\label{app-structured-example-final-management-plan}

\begin{enumerate}
\def\labelenumi{\arabic{enumi}.}
\tightlist
\item
  Await and review the pending laboratory and chest imaging results.
\item
  Perform a scrotal ultrasound.
\item
  If laboratory and imaging results are normal, management will either
  be conservative observation with follow-up ultrasound or surgical
  excisional biopsy, pending resolution of the specialist disagreement.
\item
  If a hormone-secreting tumor is identified, refer to endocrinology and
  oncology for targeted treatment.
\item
  If histology confirms a benign lesion, no further intervention is
  needed. If malignancy is identified, refer to pediatric oncology for
  staging and multidisciplinary surgical-oncologic planning.
\end{enumerate}

\paragraph{Safety Warnings}\label{app-structured-example-safety-warnings}

\begin{enumerate}
\def\labelenumi{\arabic{enumi}.}
\tightlist
\item
  The benign appearance of the mass on ultrasound does not definitively
  exclude rare pediatric breast malignancies or hormonally active
  tumors; tissue diagnosis may be required given the 18-month
  persistence and recent enlargement.
\item
  Occult testicular neoplasms must be excluded via scrotal ultrasound
  and laboratory evaluation before attributing the breast mass solely to
  idiopathic or adrenarche-related gynecomastia.
\end{enumerate}

\endgroup

\clearpage
\section{Instructions for Physician Evaluation of Medical Consensus Reports}

\subsection*{Purpose of the assessment}

The purpose of the assessment is to evaluate the quality of an already
generated medical consensus report, rather than to reassess the patient or the
source article. The physician rater should assign numeric scores in the
spreadsheet and, when needed, provide a concise explanation of the main
deficiencies.

\begin{itemize}
  \item Evaluate only what is explicitly stated in the consensus report and
  supported by the source clinical case.
  \item Do not infer missing information on behalf of the report author. If a
  relevant fact, test result, diagnosis, or explanation is absent, treat it as
  absent.
  \item Do not assign a high score solely because the text is fluent or
  well-structured if it contains fabricated data, weak clinical logic, or an
  unsafe management strategy.
  \item All numeric scores in the spreadsheet must be 0, 1, 2, or 3.
\end{itemize}

\subsection*{Files to be reviewed}

\begin{enumerate}
  \item Open the PDF or text file of the consensus report from the folder
  corresponding to the relevant specialty.
  \item Open the spreadsheet file for the same specialty. The specialty is
  indicated at the top of the worksheet.
  \item Identify the row corresponding to the case number. The row identifier
  should match the beginning of the consensus report filename.
  \item When verification of a fact is required, refer to the clinical case
  description, tables, investigation results, and other source data included in
  the case file.
\end{enumerate}

\subsection*{Structure of the spreadsheet assessment}

Each case contains three evaluated report components. The same eight criteria
are repeated for each component. Scores assigned to one component should not be
transferred automatically to another component: a strong final conclusion does
not compensate for weak reasoning, and strong reasoning does not compensate for
an inadequate final conclusion.

\begin{longtable}{|p{0.13\textwidth}|p{0.10\textwidth}|p{0.42\textwidth}|p{0.17\textwidth}|}
\hline
\textbf{Spreadsheet component} & \textbf{Columns} & \textbf{Material to review} & \textbf{Scores to assign} \\
\hline
\endfirsthead
\hline
\textbf{Spreadsheet component} & \textbf{Columns} & \textbf{Material to review} & \textbf{Scores to assign} \\
\hline
\endhead
Reasoning & B--I & In the pre-consensus stage, identify the block corresponding to the rater's specialty. Evaluate the sections describing clinical data, syndromes, differential diagnosis, diagnostic strategy, and management strategy. Do not include the section titled ``Final conclusion.'' & Eight scores, one for each criterion. \\
\hline
First-stage final judgment & J--Q & In the same specialty-specific block, evaluate only the section titled ``Final conclusion.'' This represents the specialist's final position before the overall consensus discussion. & Eight scores, one for each criterion. \\
\hline
Consensus conclusion & R--Y & Evaluate the overall consensus stage, including the final opinion, summary, final differential diagnosis, diagnostic plan, management plan, and safety warnings. & Eight scores, one for each criterion. \\
\hline
Whole-case assessment & Z--AB & Evaluate the consensus report as a whole, including the adequacy of the specialist composition, clinical complexity, and overall quality of all stages. & Three scores: specialist composition, complexity, and overall quality. \\
\hline
Comment & AC & Briefly explain, if possible, why low scores were assigned or identify the most important errors. & One to four sentences. \\
\hline
\end{longtable}

\subsection*{Eight repeated criteria}

The following eight criteria are applied separately to the reasoning component,
the first-stage final judgment, and the final consensus conclusion.

\begin{longtable}{|p{0.13\textwidth}|p{0.30\textwidth}|p{0.43\textwidth}|}
\hline
\textbf{Columns} & \textbf{Criterion} & \textbf{Operational meaning} \\
\hline
\endfirsthead
\hline
\textbf{Columns} & \textbf{Criterion} & \textbf{Operational meaning} \\
\hline
\endhead
B / J / R & Factual accuracy and sources & Facts, numerical values, dates, investigations, and treatments are not fabricated or distorted. \\
\hline
C / K / S & Identification and prioritization of clinical problems & The main clinical problems are identified and prioritized appropriately. \\
\hline
D / L / T & Completeness and relevance of the data used & Relevant data are used, while irrelevant or incidental details do not drive the assessment. \\
\hline
E / M / U & Coherence of clinical reasoning & The report contains a coherent chain linking data, explanation, diagnoses, conclusions, and management. \\
\hline
F / N / V & Quality of differential diagnosis & Plausible alternatives are considered, with arguments for and against them and methods for verification. \\
\hline
G / O / W & Quality of the final clinical conclusion & The conclusion is clear, clinically appropriate, and answers the main clinical question. \\
\hline
H / P / X & Justification and specificity of management strategy & The action plan is concrete, justified, and clinically usable. \\
\hline
I / Q / Y & Clinical safety & Dangerous conditions, contraindications, monitoring requirements, and actions in case of deterioration are not missed. \\
\hline
\end{longtable}

\subsection*{General 0--3 scoring scale}

\begin{longtable}{|p{0.10\textwidth}|p{0.58\textwidth}|p{0.22\textwidth}|}
\hline
\textbf{Score} & \textbf{Assign this score when} & \textbf{Interpretation} \\
\hline
\endfirsthead
\hline
\textbf{Score} & \textbf{Assign this score when} & \textbf{Interpretation} \\
\hline
\endhead
0 & The criterion is not met. There is a major error, an unsafe recommendation, or fabricated information that invalidates the conclusion. & The response cannot be trusted. \\
\hline
1 & Some useful content is present, but there is a substantial deficiency: important information is missing, the logic is weak, the conclusion is incomplete, or factual reliability is poor. & The response requires substantial revision. \\
\hline
2 & The criterion is generally met, but there is a noticeable deficiency, such as missing data, incomplete arguments, absent timeframes, unclear conditions, or insufficient specification. & The response may be acceptable after revision. \\
\hline
3 & The criterion is fully met and clinically convincing. Important facts are accurate, the logic is clear, and the conclusion and management plan are specific. & The response is strong. \\
\hline
\end{longtable}

If the rater is uncertain between two scores, the lower score should be assigned
when the deficiency could affect the diagnosis, patient risk, or management
strategy.

\subsection*{Rule for fabricated facts}

A fabricated fact is a clinically significant test result, numerical value,
date, symptom, diagnosis, complication, procedure, or treatment that is absent
from the source case but is presented by the consensus report as an established
fact.

\begin{itemize}
  \item If a fabricated fact affects the diagnosis, differential diagnosis,
  final conclusion, or management strategy, the criterion ``Factual accuracy
  and sources'' should receive 0 points.
  \item If the reasoning, differential diagnosis, final conclusion, or
  management strategy is based on the fabricated fact, the corresponding
  criteria should receive no more than 1 point.
  \item If the fabricated fact creates a risk of unsafe treatment or omission
  of an important threat, the criterion ``Clinical safety'' should also receive
  no more than 1 point.
  \item The comment should specify which fact was fabricated and how it was
  used in the report.
\end{itemize}

\begin{longtable}{|p{0.44\textwidth}|p{0.44\textwidth}|}
\hline
\textbf{Example} & \textbf{Scoring implication} \\
\hline
\endfirsthead
\hline
\textbf{Example} & \textbf{Scoring implication} \\
\hline
\endhead
The source case contains no leukocytosis, C-reactive protein value, procalcitonin level, creatine kinase value, troponin value, or acute kidney injury, but the consensus report presents these findings as facts and uses them to support a diagnosis. & This is not a minor inaccuracy. Factual accuracy should receive 0 points. Reasoning, differential diagnosis, the final conclusion, and management strategy should receive no more than 1 point if they rely on these data. \\
\hline
The consensus report states that C-reactive protein may be elevated and should be checked, although no C-reactive protein result is available. & This is a hypothesis rather than a fact. It is permissible if it is clearly framed as a hypothesis and not presented as an already available test result. \\
\hline
\end{longtable}

\subsection*{Procedure for scoring one case row}

\begin{enumerate}
  \item Identify the case number in the spreadsheet.
  \item Read the clinical case description, including the patient
  characteristics, main complaints, important laboratory results, imaging,
  clinical course, treatment, and safety risks.
  \item In the pre-consensus stage, identify the block corresponding to the
  rater's specialty.
  \item Assign eight scores for the ``Reasoning'' component in columns B--I.
  \item In the same specialty-specific block, identify the ``Final conclusion''
  section and assign eight scores for the ``First-stage final judgment''
  component in columns J--Q.
  \item Proceed to the overall consensus stage and assign eight scores for the
  ``Consensus conclusion'' component in columns R--Y.
  \item Assign the whole-case scores: column Z for adequacy of the specialist
  composition, column AA for clinical complexity, and column AB for the overall
  quality of all stages. The complexity score is the rater's expert judgment
  from the perspective of the rater's specialty.
  \item Complete the general comment in column AC. Scores of 0 or 1 should be
  explained whenever possible.
\end{enumerate}

\subsection*{Criterion-specific scoring rules}

\subsubsection*{Factual accuracy and sources}

Assess whether the facts in the consensus report match the source case,
including symptoms, dates, laboratory values, test results, diagnoses,
procedures, and treatments.

\begin{longtable}{|p{0.10\textwidth}|p{0.78\textwidth}|}
\hline
\textbf{Score} & \textbf{Assign this score when} \\
\hline
\endfirsthead
\hline
\textbf{Score} & \textbf{Assign this score when} \\
\hline
\endhead
0 & A fabricated or distorted clinically significant fact is present and could alter the diagnosis, risk assessment, or management strategy; or the conclusion directly contradicts the source data. \\
\hline
1 & There are several factual errors or one important unsupported interpretation; the sources are unclear; however, the entire conclusion does not depend exclusively on fabricated data. \\
\hline
2 & The main facts are reported correctly and no clinically significant fabricated facts are present, but there are minor inaccuracies or incomplete source references. \\
\hline
3 & All important facts, numerical values, dates, and investigations are accurate; their sources are clear; and hypotheses are distinguished from established facts. \\
\hline
\end{longtable}

\subsubsection*{Identification and prioritization of clinical problems}

Assess whether the report identifies what is clinically central, what is
secondary, and what represents an immediate threat.

\begin{longtable}{|p{0.10\textwidth}|p{0.78\textwidth}|}
\hline
\textbf{Score} & \textbf{Assign this score when} \\
\hline
\endfirsthead
\hline
\textbf{Score} & \textbf{Assign this score when} \\
\hline
\endhead
0 & The main clinical problem is missed or incorrectly stated. \\
\hline
1 & The main problem is only partly recognized, but priorities are incorrect or an important risk is omitted. \\
\hline
2 & The main problem and most important associated problems are identified correctly, but there is a limited omission or weak prioritization. \\
\hline
3 & All key clinical problems and risks are identified and arranged in an appropriate clinical order. \\
\hline
\end{longtable}

\subsubsection*{Completeness and relevance of the data used}

Assess whether the report uses the data required to make a reliable diagnosis
or select an appropriate management strategy.

\begin{longtable}{|p{0.10\textwidth}|p{0.78\textwidth}|}
\hline
\textbf{Score} & \textbf{Assign this score when} \\
\hline
\endfirsthead
\hline
\textbf{Score} & \textbf{Assign this score when} \\
\hline
\endhead
0 & Key data are omitted, making the conclusion unreliable or incorrect. \\
\hline
1 & Only part of the available data is used; several important domains are omitted, such as history, physical examination, clinical course, laboratory results, imaging, treatment, or risks. \\
\hline
2 & Most important data are considered, but one or two meaningful omissions remain, or excessive irrelevant detail reduces focus. \\
\hline
3 & All data affecting diagnosis, risk assessment, and management are considered; irrelevant information does not drive the conclusion. \\
\hline
\end{longtable}

\subsubsection*{Coherence of clinical reasoning}

Assess whether the report contains a valid causal and inferential chain rather
than a set of disconnected statements.

\begin{longtable}{|p{0.10\textwidth}|p{0.78\textwidth}|}
\hline
\textbf{Score} & \textbf{Assign this score when} \\
\hline
\endfirsthead
\hline
\textbf{Score} & \textbf{Assign this score when} \\
\hline
\endhead
0 & Conclusions do not follow from the data, contradict the data, or are based on fabricated facts. \\
\hline
1 & Logic is present only in fragments; important conclusions are not explained, and causal links are superficial. \\
\hline
2 & Reasoning is generally coherent, but there are limited logical gaps, weak mechanisms, or incomplete chronology. \\
\hline
3 & There is a clear chain from data to syndromes or mechanisms, differential diagnosis, conclusion, and management. Chronology and clinical dynamics are considered. \\
\hline
\end{longtable}

\subsubsection*{Quality of differential diagnosis}

Assess whether plausible alternatives are considered rather than a single
preferred diagnosis.

\begin{longtable}{|p{0.10\textwidth}|p{0.78\textwidth}|}
\hline
\textbf{Score} & \textbf{Assign this score when} \\
\hline
\endfirsthead
\hline
\textbf{Score} & \textbf{Assign this score when} \\
\hline
\endhead
0 & Differential diagnosis is absent, irrelevant, omits an obvious dangerous alternative, or is based on fabricated data. \\
\hline
1 & Too few alternatives are considered, probabilities are not ordered, and arguments for and against the alternatives are largely absent. \\
\hline
2 & The main alternatives are identified, but arguments, prioritization, or verification methods are incomplete. \\
\hline
3 & Diagnoses are relevant and prioritized; for each important alternative, there are arguments for and against it and a way to confirm or exclude it. \\
\hline
\end{longtable}

\subsubsection*{Quality of the final clinical conclusion}

Assess whether the final conclusion answers the main clinical question: what is
most likely occurring, how well this is supported, and what remains uncertain.

\begin{longtable}{|p{0.10\textwidth}|p{0.78\textwidth}|}
\hline
\textbf{Score} & \textbf{Assign this score when} \\
\hline
\endfirsthead
\hline
\textbf{Score} & \textbf{Assign this score when} \\
\hline
\endhead
0 & The conclusion is absent, clinically incorrect, contradictory, or based on fabricated clinically significant facts. \\
\hline
1 & The conclusion is overly general, incomplete, does not answer the main question, or merely repeats a list of possible diagnoses without a clear position. \\
\hline
2 & The main conclusion is generally correct and understandable, but requires clarification regarding probability, conditions, limitations, or links to key data. \\
\hline
3 & The conclusion is clear, specific, and justified. It names the main diagnosis, syndrome, or working hypothesis, states the level of confidence, and identifies data that could change the decision. \\
\hline
\end{longtable}

\subsubsection*{Justification and specificity of management strategy}

Assess whether the plan can be acted upon clinically: what should be done,
when, why, and what the next step should be.

\begin{longtable}{|p{0.10\textwidth}|p{0.78\textwidth}|}
\hline
\textbf{Score} & \textbf{Assign this score when} \\
\hline
\endfirsthead
\hline
\textbf{Score} & \textbf{Assign this score when} \\
\hline
\endhead
0 & Management strategy is absent, clearly inappropriate, unsafe, or based on fabricated clinically significant facts. \\
\hline
1 & Some appropriate actions are present, but the plan is generic: priorities, timeframes, reassessment criteria, monitoring, or links to the diagnosis are missing. \\
\hline
2 & The main actions are appropriate, but specific details are missing, such as timeframes, monitoring parameters, responsible specialties, or escalation criteria. \\
\hline
3 & The plan is specific and prioritized. Diagnostic actions, treatment, monitoring, timeframes, reassessment criteria, and next steps are linked to the clinical reasoning. \\
\hline
\end{longtable}

\subsubsection*{Clinical safety}

Assess whether the consensus report creates risk for the patient or omits
threats that require immediate monitoring or action.

\begin{longtable}{|p{0.10\textwidth}|p{0.78\textwidth}|}
\hline
\textbf{Score} & \textbf{Assign this score when} \\
\hline
\endfirsthead
\hline
\textbf{Score} & \textbf{Assign this score when} \\
\hline
\endhead
0 & There is a potentially unsafe recommendation, an omitted life-threatening condition, or an ignored contraindication. \\
\hline
1 & Some risks are recognized, but there is a substantial gap in monitoring, contraindications, drug interactions, or actions in case of deterioration. \\
\hline
2 & The management strategy is generally safe, but one important precaution or one deterioration scenario is not addressed. \\
\hline
3 & Urgent risks, contraindications, interactions, complications, monitoring requirements, and an action plan for deterioration are addressed. \\
\hline
\end{longtable}

\subsection*{Whole-case criteria}

These three scores are assigned once per case row and are not repeated for each
component. The rater should evaluate the consensus report as a whole.

\subsubsection*{Adequacy of specialist composition, column Z}

Assess whether the selected specialists were clinically necessary for resolving
the case.

\begin{longtable}{|p{0.10\textwidth}|p{0.78\textwidth}|}
\hline
\textbf{Score} & \textbf{Assign this score when} \\
\hline
\endfirsthead
\hline
\textbf{Score} & \textbf{Assign this score when} \\
\hline
\endhead
0 & The specialist composition does not cover the main clinical problem; a key specialty is absent, and the consensus cannot reliably address the case. \\
\hline
1 & Part of the composition is appropriate, but one important specialty is absent or the roles of the participants do not match the main clinical tasks. \\
\hline
2 & Most necessary specialties are present, but the composition does not fully cover diagnosis, treatment, or safety. \\
\hline
3 & The composition is appropriate: key specialists are present, and their roles correspond to the main problems and patient risks. \\
\hline
\end{longtable}

\subsubsection*{Clinical complexity of the case, column AA}

This score reflects the difficulty of the case itself, not the quality of the
generated consensus report. A poorly written consensus report does not make the
case more complex. The score represents the rater's specialty-specific expert
judgment: 0 indicates a case that is simple for the rater's specialty and for
which the need for a consensus is questionable; 3 indicates a very complex case
for which convening a consensus is highly appropriate.

\begin{longtable}{|p{0.10\textwidth}|p{0.78\textwidth}|}
\hline
\textbf{Score} & \textbf{Assign this score when} \\
\hline
\endfirsthead
\hline
\textbf{Score} & \textbf{Assign this score when} \\
\hline
\endhead
0 & The case is simple for the rater's specialty, with one obvious problem, a standard diagnostic pathway, limited need for additional analysis, and low risk of error. \\
\hline
1 & The case is moderately complex, with several relevant data points and simple alternatives, but the expected decision is largely standard. \\
\hline
2 & The case is complex, with involvement of several systems, ambiguous data, important clinical dynamics, or several plausible differential diagnoses. \\
\hline
3 & The case is very complex, with a rare or atypical presentation, high risk, incomplete or conflicting data, and a clear need for multidisciplinary input. \\
\hline
\end{longtable}

\subsubsection*{Overall quality of all consensus stages, column AB}

This is a global expert judgment of whether the consensus report is usable as a
coherent clinical product.

\begin{longtable}{|p{0.10\textwidth}|p{0.78\textwidth}|}
\hline
\textbf{Score} & \textbf{Assign this score when} \\
\hline
\endfirsthead
\hline
\textbf{Score} & \textbf{Assign this score when} \\
\hline
\endhead
0 & The consensus report as a whole is unusable: there are many major errors, and fabricated data lead to an incorrect conclusion or unsafe management. \\
\hline
1 & Some useful fragments are present, but overall quality is low: substantial errors occur across several stages, and the result requires major revision. \\
\hline
2 & The consensus report is generally acceptable, but there are noticeable deficiencies in facts, reasoning, conclusion, management, or consistency between stages. \\
\hline
3 & All stages are consistent, factually accurate, logically coherent, and clinically useful; the result can be accepted without major revision. \\
\hline
\end{longtable}

\subsection*{Writing comments}

The comment should not restate the entire case. Its purpose is to make clear
why low scores were assigned and where the main deficiency is located.

\begin{itemize}
  \item A comment should be provided when any score of 0 or 1 is assigned.
  \item For fabricated facts, specify the exact fact and explain why it is not
  present in the source data.
  \item If the assessment is generally favorable, one concise sentence about
  the main limitation is sufficient. If there are no major problems, the field
  may be left blank.
  \item Avoid non-specific comments such as ``many errors,'' ``acceptable but
  could be better,'' or ``poor conclusion'' unless the problem is specified.
\end{itemize}

\begin{longtable}{|p{0.33\textwidth}|p{0.55\textwidth}|}
\hline
\textbf{Non-specific comment} & \textbf{Acceptable comment} \\
\hline
\endfirsthead
\hline
\textbf{Non-specific comment} & \textbf{Acceptable comment} \\
\hline
\endhead
Many errors. & The reasoning cites a non-existent C-reactive protein value and uses it to support an infectious diagnosis; factual accuracy is 0, and reasoning and differential diagnosis should not exceed 1. \\
\hline
Acceptable, but could be better. & The main diagnosis is selected appropriately, but the final plan lacks timeframes for reassessment, escalation criteria, and safety monitoring. \\
\hline
Poor conclusion. & The final conclusion lists three diagnoses without prioritization and does not identify the leading hypothesis at the current stage. \\
\hline
\end{longtable}

\subsection*{Final check before submission}

Before submitting the completed spreadsheet, verify that:

\begin{itemize}
  \item all numeric cells in columns B--AB are completed for the case row;
  \item each numeric cell contains only 0, 1, 2, or 3;
  \item fabricated analyses, symptoms, diagnoses, or treatments, if present,
  are reflected in the factual accuracy score and in the comment;
  \item the ``Clinical complexity of the case'' score refers to the case
  itself, not to the quality of the generated consensus report;
  \item no facts absent from the source case or consensus report have been
  added by the rater during scoring.
\end{itemize}

\subsection*{Brief checklist}

\begin{longtable}{|p{0.44\textwidth}|p{0.44\textwidth}|}
\hline
\textbf{Question} & \textbf{If the answer is no} \\
\hline
\endfirsthead
\hline
\textbf{Question} & \textbf{If the answer is no} \\
\hline
\endhead
Do the facts match the source case? & Reduce the factual accuracy score. If a fabricated fact affects the conclusion, assign 0 for factual accuracy. \\
\hline
Is the main clinical problem identified? & Reduce the scores for problem prioritization and the final clinical conclusion. \\
\hline
Are key laboratory tests, imaging, history, and clinical dynamics used? & Reduce the completeness score. \\
\hline
Do conclusions follow logically from the facts? & Reduce the coherence of reasoning score. \\
\hline
Is the differential diagnosis adequate? & Reduce the differential diagnosis score. \\
\hline
Does the final conclusion answer the main clinical question? & Reduce the final clinical conclusion score. \\
\hline
Is the plan specific regarding what should be done, when, why, and how it should be monitored? & Reduce the management strategy score. \\
\hline
Would it be safe for the patient to follow this plan? & Reduce the clinical safety score. \\
\hline
\end{longtable}

\end{document}